\documentclass{article} 
\usepackage{iclr2023_conference,times}

\usepackage{amsmath,amsfonts,bm}

\def\eqref#1{equation~\ref{#1}}
\def\Eqref#1{Equation~\ref{#1}}

\def\1{\bm{1}}

\DeclareMathAlphabet{\mathsfit}{\encodingdefault}{\sfdefault}{m}{sl}
\SetMathAlphabet{\mathsfit}{bold}{\encodingdefault}{\sfdefault}{bx}{n}

\DeclareMathOperator*{\argmax}{arg\,max}

\usepackage{hyperref}
\usepackage{url}

\usepackage{graphicx}
\usepackage{wrapfig}
\usepackage{amsmath}
\usepackage{amssymb}
\usepackage{bbm}
\usepackage{tabularx}
\usepackage{booktabs}
\usepackage{multirow}
\usepackage{array}
\usepackage{mathtools}
\usepackage{mwe}
\usepackage{subcaption}
\usepackage{xcolor,colortbl}
\usepackage{makecell}
\usepackage{tabularx}
\usepackage{enumitem}
\usepackage{tikz}
\usepackage{collcell}
\usepackage{etoolbox}
\usepackage{pgf}
\usepackage{caption}
\definecolor{forestgreen}{HTML}{228b22}
\definecolor{bittersweet}{rgb}{1.0, 0.44, 0.37}

\title{$k$NN Prompting: Beyond-Context Learning with Calibration-Free Nearest Neighbor Inference}

\author{Benfeng Xu$^1$, Quan Wang$^2$, Zhendong Mao$^{14}$\thanks{Corresponding author.}, Yajuan Lyu$^3$, Qiaoqiao She$^3$, Yongdong Zhang$^{14}$\\
$^1$University of Science and Technology of China, Hefei, China\\
$^2$Beijing University of Posts and Telecommunications, Beijing, China\quad$^3$Baidu Inc., Beijing, China\\
$^4$Institute of Artificial Intelligence, Hefei Comprehensive National Science Center, China\\
\texttt{benfeng@mail.ustc.edu.cn},\quad\texttt{zdmao@ustc.edu.cn}\\
}

\iclrfinalcopy 
\begin{document}

\maketitle

\begin{abstract}
In-Context Learning (ICL), which formulates target tasks as prompt completion conditioned on in-context demonstrations, has become the prevailing utilization of LLMs.
In this paper, we first disclose an actual predicament for this typical usage that it can not scale up with training data due to context length restriction.
Besides, existing works have shown that ICL also suffers from various biases and requires delicate calibration treatment.
To address both challenges, we advocate a simple and effective solution, $k$NN Prompting, which first queries LLM with training data for distributed representations, then predicts test instances by simply referring to nearest neighbors.
We conduct comprehensive experiments to demonstrate its two-fold superiority: 1)
\textbf{Calibration-Free}: $k$NN Prompting does not directly align LLM output distribution with task-specific label space, instead leverages such distribution to align test and training instances. It significantly outperforms state-of-the-art calibration-based methods under comparable few-shot scenario. 2) \textbf{Beyond-Context}: $k$NN Prompting can further scale up effectively with as many training data as are available, continually bringing substantial improvements. The scaling trend holds across 10 orders of magnitude ranging from 2 shots to 1024 shots as well as different LLMs scales ranging from 0.8B to 30B.
It successfully bridges data scaling into model scaling, and brings new potentials for the gradient-free paradigm of LLM deployment. Code is publicly available\footnote{\url{https://github.com/BenfengXu/KNNPrompting}}.
\end{abstract}

\section{Introduction}

\begin{wrapfigure}[15]{r}{0.48\textwidth}
\vspace{-43.5pt}
  \begin{center}
    \includegraphics[width=0.45\textwidth]{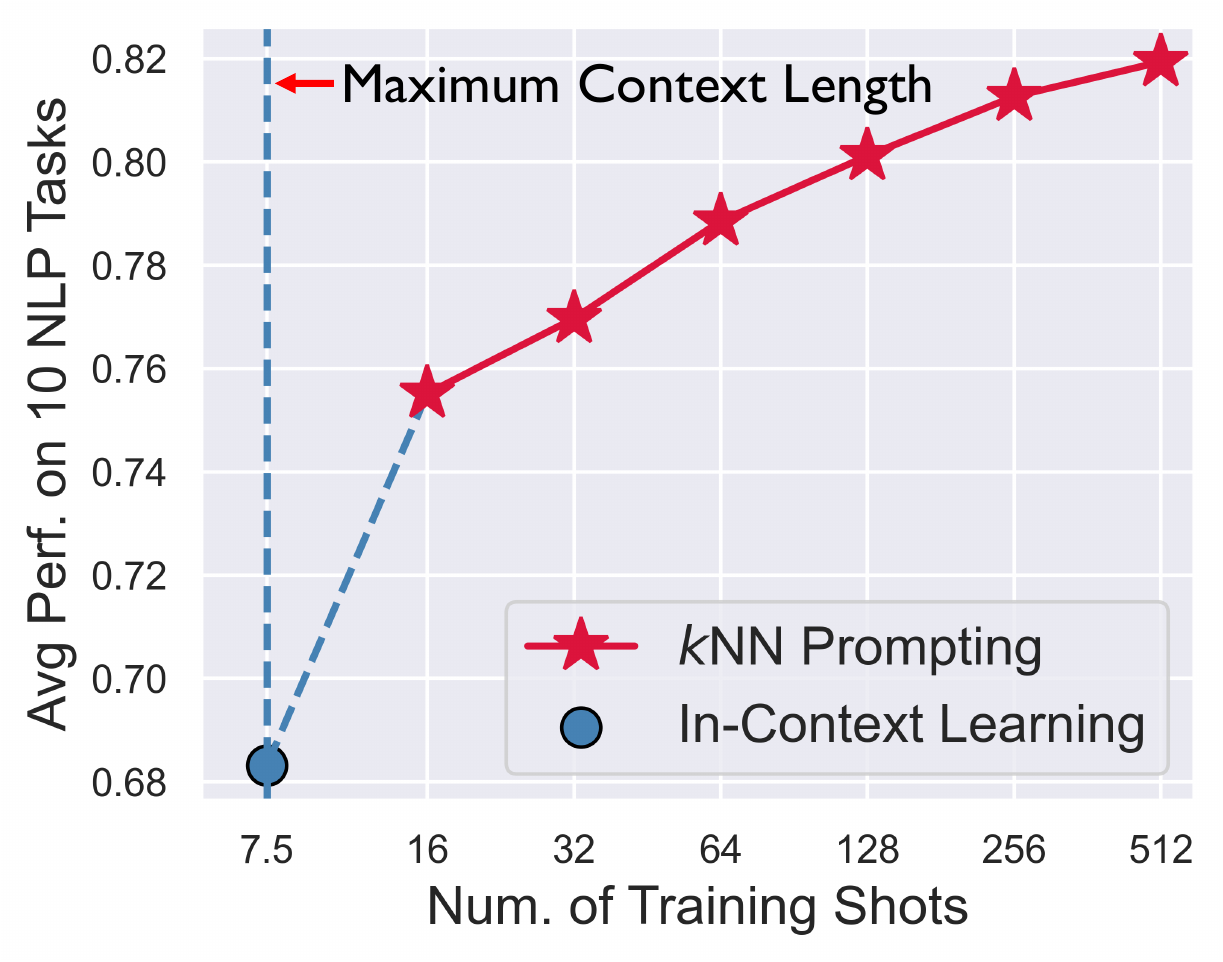}
  \end{center}
  \captionsetup{skip=0pt}
  \caption{$k$NN Prompting brings substantial improvements over standard ICL, and can continually scale up beyond the context with as many data as are available. Conducted with GPT XL.}
  \label{fig:figure1}
\end{wrapfigure}

Large language models (LLMs), when scale up to billions of parameters, have demonstrated remarkable capabilities in a wide range of NLP tasks~\citep{radford2019language,NEURIPS2020_1457c0d6}.
However, such models are prohibitively expensive to train with most of  the research- or consumer-level devices, though some of them are already publicly available~\citep{zhang2022opt}.
As a result, it is now an emerging paradigm that LLMs are hosted in a remote data center while accessed by end users or applications via simple API requests\footnote{\url{https://openai.com/api/}, \url{https://gpt3demo.com/}}.
The typical usage of LLM under such paradigm is In-Context Learning, where LLM reads and completes a prompt sequence as how it is pretrained on massive text corpora.
The prompt is constructed by concatenation of several training examples and a test instance, and the prediction is obtained by mapping the LLM word continuations back to label space.

It is widely investigated and acknowledged that modern neural networks generally perform better w.r.t. increased training data.
Specifically, there exists a power law between expected model performance and available data scale~\citep{hestness2017deep,rosenfeld2020a}.
For ICL, it is also empirically observed that the performance continually improves when more training examples are prepended into the prompt~\citep{NEURIPS2020_1457c0d6}.
However, such improvements are quickly prevented by the predicament of context length restriction, as language models are designed and trained to only process sequences within a fixed length, which is in fact 1024 or 2048 tokens.
In order to utilize more training data, several works try to select the most relevant examples to compose the prompt before querying LLM~\citep{liu-etal-2022-makes,rubin-etal-2022-learning}, but still only in-context examples can actually participate the LLM inference while most training data are discarded beforehand, thus providing marginal data scaling benefits. Besides, their reliance on external retriever also incurs further complications.
As a consequence, such a situation poses a serious challenge for many practical scenarios where more than a few training data are available.

Another vulnerability of ICL is the severe bias existed in the output distribution of LLMs, which results in considerable performance degradation~\citep{holtzman-etal-2021-surface} and instability~\citep{lu-etal-2022-fantastically} as shown in existing works.
Accordingly, many have proposed various ways to calibrate the output distribution~\citep{pmlr-v139-zhao21c,jiang-etal-2021-know,min-etal-2022-noisy}. For example,~\citet{pmlr-v139-zhao21c} measure such bias by probing LLM with a "\textit{NA}" example and record the according prior.
However, as LLMs are pretrained on general-domain natural language, its capability to complete a fabricated prompt is essentially not aligned with downstream task-specific label space.
As a consequence, such calibration-based methods can only alleviate the bias to a limited extent.

In this paper, we advocate a simple and effective solution, $k$NN Prompting, to address both challenges.
Specifically, we assign training data into a demonstration set and an anchor set.
We append each anchor example into the prompt and query LLM, then instead of aligning word continuations with labels, we collect the language modeling probability as distributed representation and cache it into a local datastore.
At inference time, for each test instance, we similarly obtain its representation and match it against the maintained datastore to make predictions.
In general, the proposed framework enables both calibration-free optimization because it avoids forced input-label alignment, and beyond-context learning because the anchor set allows utilization of unlimited training data.

We conduct comprehensive experiments using 10 established text classification tasks to demonstrate the significant superiority of $k$NN Prompting across various scenarios and against competitive opponents:
1) Under few shot scenario where training data is very limited and fits in the context, $k$NN Prompting outperforms state-of-the-art calibration-based methods by considerable margin (up to +7.07).
2) Under low resource or fully supervised scenario where training data can not fit in the context, $k$NN Prompting further exhibits its major advantage.
It can effectively scale up with as many training data as are available across 10 orders of magnitude (2 shots$\sim$1024 shots, see Figure~\ref{fig:figure1} for illustration) as well as different LLMs scales (0.8B$\sim$30B).
Specifically, with only 32 shots training data, it dramatically improves ICL by +13.58 in average score at its most, and achieves absolute improvements up to +18.84 under fully supervised setting.
We also provide formal explanation on the intrinsic mechanism of effectiveness, as well as detailed analyses regarding its robustness and choices of design.
Accompanied with these appealing aspects, $k$NN Prompting is in general a promising solution that bridges the benefits of data scaling into model scaling to take the gradient-free paradigm of LLM deployment one step further.

\section{Background: In-Context Learning}
In this section, we formulate the task and recap the ICL baseline.
Assuming a target task with training data set $\mathcal{T}=\{(x_i, y_i)\}$, and $Y$ as its categorical label space.
At inference time, the model is asked to predict $y_\mathrm{test}$ given test instance $x_\mathrm{test}$.
We then denote an LLM $\bm\theta$ that is pretrained with a standard language modeling objective.
At employment, it samely predicts a probability distribution $p(w_t|\bm{w}_{<t}, \bm\theta)$ for the next token at $t$-th position conditioned on previous context $\bm{w}_{<t}$.

In-context learning first formulates training examples $\{(x_i, y_i)\}$ in the format of input-label mapping via intuitive templates (see Appendix~\ref{appendix:template} for illustration), and concatenates them into a natural language sequence along with the test instance to construct a prompt:
\begin{equation}
P=\pi(x_1, y_1)\oplus\pi(x_2, y_2)\oplus ... \oplus\pi(x_{|\mathcal{T}|}, y_{|\mathcal{T}|})\oplus\pi(x_\mathrm{test}, *)
\label{eq_prompt}
\end{equation}
where $\pi$ denotes template-based transformation (see Appendix for illustration) and $\oplus$ denotes concatenation operation.
Note that $\pi$ implies a verbalization process that maps label space $Y$ to corresponding tokens $V$ picked from the LM vocabulary.
When queried by the prompt $P$, LLM will try to mimic the prepended training examples in the context and predict a probability distribution $p(v|P, \bm\theta)$ for the next token $v$.
We then map it back to label space $Y$ as prediction:
\begin{equation}
\hat{y}_\mathrm{test}=\argmax_{y\in Y}(v|P, \bm\theta),\quad y\xrightarrow{\pi}v
\end{equation}

\begin{table*}[!t]
\centering
\resizebox{\columnwidth}{!}{
\begin{tabular}{{ccccccccccc}}
\toprule
&\textbf{SST2}&\textbf{SUBJ}&\textbf{MPQA}&\textbf{AGNews}&\textbf{CB}&\textbf{CR}&\textbf{DBPedia}&\textbf{MR}&\textbf{RTE}&\textbf{TREC}\\
\midrule
\textbf{Num. of Shots (TP)}&20 (2\%)&12 (1\%)&39 (0\%)&3 (0\%)&2 (0\%)&14 (4\%)&1 (77\%)&14 (4\%)&4 (0\%)&8 (1\%)\\
$M_T$&16&8&32&2&2&8&1&8&4&8\\
\bottomrule
\end{tabular}}
\caption{Maximum number of training shots (per class) allowed by 1024 tokens of context. Calculated using GPT2 tokenizer. Inside the parentheses are Truncation Probability (TP, i.e., whether or not truncated, restricted below 5\%). We set $M_T$ for each task $T$ in our experiments for simplicity.}
\label{table:maxshotstatistics}
\end{table*}

\begin{wrapfigure}[17]{r}{0.50\textwidth}
\vspace{-21pt}
  \begin{center}
    \includegraphics[width=0.42\textwidth]{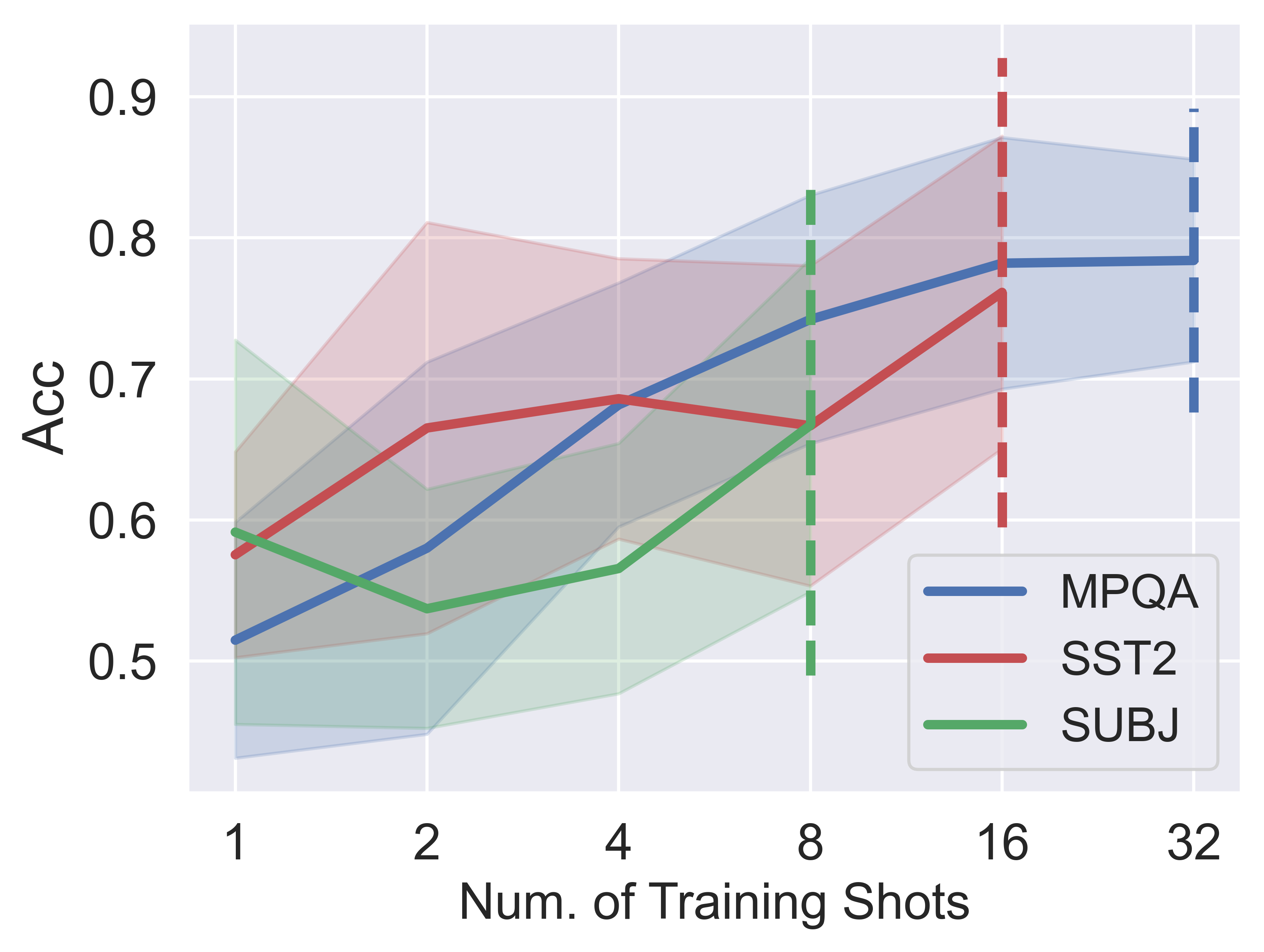}
  \end{center}
  \captionsetup{skip=0pt}
  \caption{ICL improves with num. of training examples but is limited by context length restriction.}
  \label{fig:scaleiclwithmoredata}
\end{wrapfigure}

Figure~\ref{fig:scaleiclwithmoredata} provides a pilot study showing that when prompt $P$ includes more demonstrations, the performance consistently improves, which is in accord with the power law of data scaling investigated in many existing studies~\citep{hestness2017deep,rosenfeld2020a}.
However, this trend is then prevented by the context length restriction.
We provide more comprehensive statistics in Table~\ref{table:maxshotstatistics} and Appendix~\ref{appendix:datasetstatistics}.
In conclusion, this situation poses an actual challenge in many scenarios where one would further collect training examples from few-shots to dozens and expect improved performance, but the power law fails.

\section{$k$NN Prompting}
\begin{figure*}[t!]
\centering
 \includegraphics[width=0.8\textwidth]{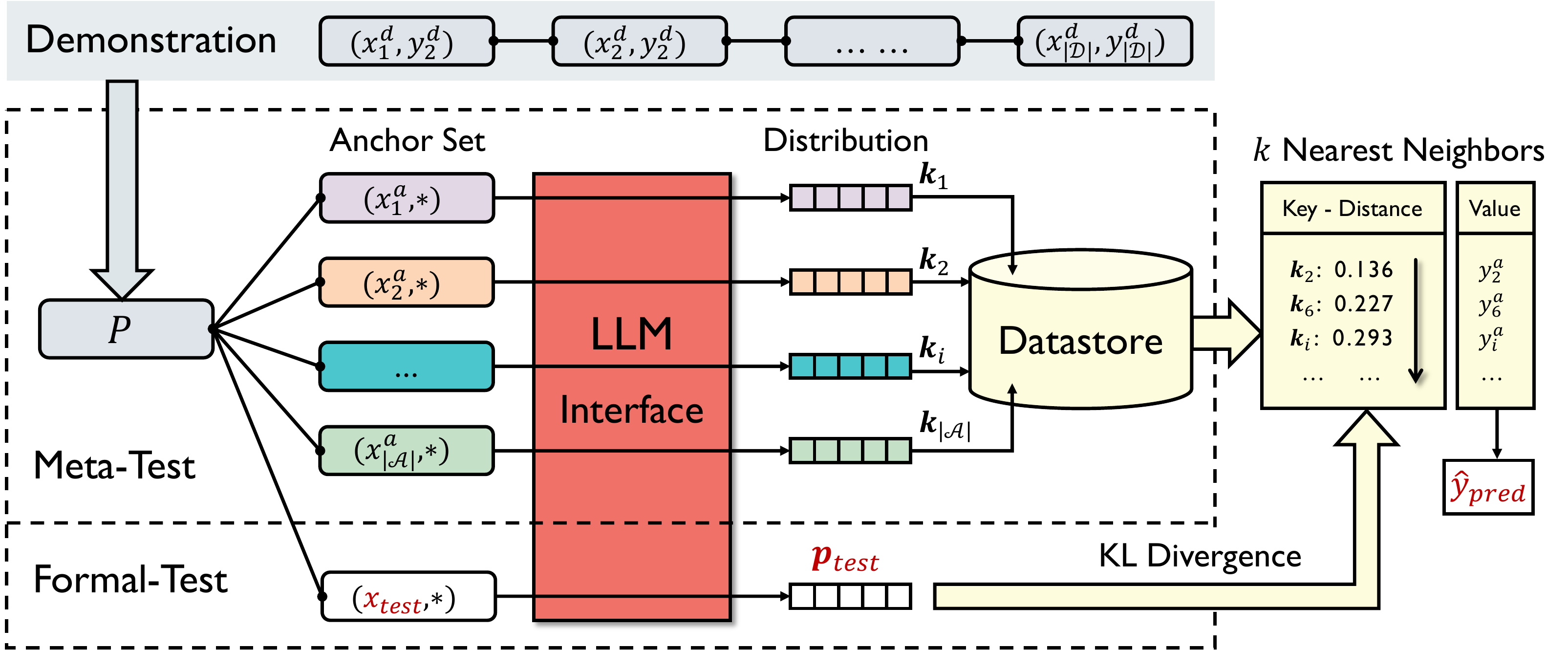}
\caption{The overall framework of $k$NN Prompting}
\label{fig:framework}
\end{figure*}

In this section, we introduce the $k$NN Prompting framework.
For a training data set $\mathcal{T}$, we split and exploit it in two respective usage: $\mathcal{T}=\mathcal{D}\cup\mathcal{A}$, i.e., a demonstration set $\mathcal{D}=\{(x_i^d, y_i^d)\}$ and an anchor set $\mathcal{A}=\{(x_i^a, y_i^a)\}$.
$k$NN Prompting consists of two stages namely meta test and formal test, the overall framework is illustrated in Figure \ref{fig:framework}.

\paragraph{Meta Test} We build a datastore that caches all anchor examples in $\mathcal{A}$ for later inference time usage.
For each $x_i^a$, we respectively concatenate it into prompt $P$, where the prompt prefix is constructed using the demonstration set the same as Equation~\ref{eq_prompt}:
\begin{equation}
P_i=\pi(x_1^d, y_1^d)\oplus\pi(x_2^d, y_2^d)\oplus ... \oplus\pi(x_{|\mathcal{D}|}^d, y^d_{|\mathcal{D}|})\oplus\pi(x_i^a, *)
\end{equation}
By querying LLM using $P_i$, we obtain the distribution $p(v|P_i, \bm\theta)$.
Instead of mapping it back to label space $V$, we cache the entire language modeling distribution as the key representation:
\begin{equation}
\bm{k}_i=p(v|P_i, \bm\theta)
\end{equation}
Accordingly, label $y_i^a$ is the value.
The entire datastore thus consists of paired $\{\bm{k}_i,y_i^a\}$, we denote the set of keys as $\mathcal{K}$.

\paragraph{Formal Test} At inference time, for each test instance $x_\mathrm{test}$, we construct the same prompt as \Eqref{eq_prompt}, and obtain distribution $\bm{p}_{\mathrm{test}}=p(v|P_{test}, \bm\theta)$.
We then match the distribution against cached $\mathcal{K}$ in the datastore, where standard $KL$ divergence is used to measure the distance:
\begin{equation}
D_{KL}(\bm{p}_{\mathrm{test}}||\bm{k}_i)=\sum_{v}p(v|P_{test}, \bm\theta)\log\frac{p(v|P_{test}, \bm\theta)}{p(v|P_i, \bm\theta)}
\label{kldistance}
\end{equation}

The predictions are calculated by aggregating its $k$ nearest neighbors:
\begin{equation}
\hat{y}_{pred}=\argmax_{y\in Y}\sum_{i\in\mathrm{NN}^k(\bm{p}_{\mathrm{test}}, \mathcal{K})}\mathbbm{1}(y_i^a=y)
\end{equation}
where $\mathrm{NN}^k(*, \mathcal{K})$ denotes the set of $k$ nearest neighbors in $\mathcal{K}$.

\section{Experiments}
\subsection{Setup}
We use 10 established text classification datasets, respectively SST2~\citep{socher-etal-2013-recursive}, SUBJ~\citep{pang-lee-2004-sentimental}, MPQA~\citep{wiebe2005annotating}, AGNews~\citep{NIPS2015_250cf8b5}, CB~\citep{de2019commitmentbank}, CR~\citep{10.1145/1014052.1014073}, DBPedia~\citep{NIPS2015_250cf8b5}, MR~\citep{pang-lee-2005-seeing}, RTE~\citep{10.1007/11736790_9} and TREC~\citep{10.1145/345508.345577}.
For each dataset, we devise intuitive prompt template (Appendix~\ref{appendix:template}), and other regarding statistics are listed in Appendix~\ref{appendix:datasetstatistics}.
We investigate a wide range of LLM scales, including GPT2 (0.8B and 1.5B)~\citep{radford2019language} and the OPT~\citep{zhang2022opt} series (3B-30B).
GPT2 XL is used for most analyses unless explicitly indicated.
We invariantly set the number of neighbors $k$ to 3.
There are no other hyper-parameters as the entire framework is training-free.
We run with 5 different random seeds for Figure~\ref{fig:fullysupervisedscaling} and Figure~\ref{fig:comparisontoretrievetocompose}, 10 seeds for all other results.
Mean and standard deviation are reported.

\subsection{Data Utility}
In this paper, we refer to data utility as whether and how performance scales up with increased training data.
It can be formally depicted by the power law~\citep{hestness2017deep}:
\begin{equation}
\varepsilon(m)\propto \alpha m^\beta
\end{equation}
where $m$ is the training data size, $\alpha$ is a constant scaling factor, $\beta$ describes the exponential steepness, and $\varepsilon(*)$ calculates the generalization error, i.e., test time performance.
We refer $m$ to training shots, i.e., number of examples per class throughout the paper.
In the following sections, we investigate various settings of $m$, respectively few shot ($m\leq M_T$), low resource ($m=128$), and the overall scaling law ($m\leq 1024$).
We refer $M_T$ to the maximum training shots allowed in context for each task $T$, the specific statistics can be found in Appendix~\ref{appendix:datasetstatistics}.

\subsubsection{Data Utility Under Few Shot Scenario}\label{sec:datautilityfewshot}
We first investigate $k$NN Prompting under few shot setting, i.e., $m\leq M_T$.
We simply set $|\mathcal{D}|=1$ and use all other examples for $\mathcal{A}$.
We leave further exploration of split strategies to Appendix~\ref{appendix:splitofdemoandanchor}.
In order to avoid context length restriction and maintain comparability to baselines as much as possible, 6 out of 10 tasks where $M_T\geq8$ are selected.
The baselines are ICL and state-of-the-art calibration-based augmentations.
ContextualCalibration~\citep{pmlr-v139-zhao21c} probs the prior bias w.r.t. each category and accordingly calibrate the outputs, and NoisyChannel~\citep{min-etal-2022-noisy} formulates ICL as computing input likelihood conditioned on labels.
Results in Table~\ref{table:mainresultsfewshot} show that $k$NN Prompting significantly outperforms competitive baselines under strictly comparable settings ($m\leq 8$), specifically, \textbf{+3.56} for 4 shot, and \textbf{+7.07} for 8 shot.

\begin{table*}[t]
\centering
\small
\resizebox{0.9\columnwidth}{!}{
\begin{tabular}{{p{0.9cm}l|cccccc|c}}
\toprule
\multicolumn{2}{c|}{\textbf{Setting\&Method}}&\textbf{SST2}&\textbf{SUBJ}&\textbf{MPQA}&\textbf{CR}&\textbf{MR}&\textbf{TREC}&\textbf{AVG}\\
\midrule
\multirow{5}*{$m=2$}&\small\textbf{ICL}&59.8$_{\pm\textrm{5.9}}$&51.4$_{\pm\textrm{7.6}}$&60.2$_{\pm\textrm{14.2}}$&57.3$_{\pm\textrm{5.5}}$&64.7$_{\pm\textrm{11.5}}$&50.0$_{\pm\textrm{3.4}}$&57.24\\
&\small\textbf{Contextual Calibration}&76.2$_{\pm\textrm{7.0}}$&69.8$_{\pm\textrm{7.5}}$&63.4$_{\pm\textrm{10.5}}$&60.1$_{\pm\textrm{5.7}}$&75.6$_{\pm\textrm{8.4}}$&45.5$_{\pm\textrm{7.1}}$&65.12\\
&\small\textbf{Noisy Channel}&82.6$_{\pm\textrm{3.1}}$&64.6$_{\pm\textrm{6.0}}$&60.2$_{\pm\textrm{9.1}}$&83.3$_{\pm\textrm{2.3}}$&79.4$_{\pm\textrm{2.1}}$&35.2$_{\pm\textrm{9.8}}$&\textbf{67.54}\\
&\small\textbf{$k$NN Prompting}&77.5$_{\pm\textrm{21.3}}$&73.4$_{\pm\textrm{9.0}}$&56.6$_{\pm\textrm{20.2}}$&69.7$_{\pm\textrm{21.3}}$&81.1$_{\pm\textrm{6.7}}$&41.3$_{\pm\textrm{12.0}}$&66.57\\
&\small\textbf{$k$NN Prompting (Partial)}&77.8$_{\pm\textrm{18.9}}$&68.9$_{\pm\textrm{10.3}}$&53.3$_{\pm\textrm{22.5}}$&66.7$_{\pm\textrm{22.7}}$&81.6$_{\pm\textrm{6.3}}$&40.2$_{\pm\textrm{8.3}}$&64.73\\
\midrule
\multirow{5}*{$m=4$}&\small\textbf{ICL}&67.2$_{\pm\textrm{12.6}}$&56.5$_{\pm\textrm{11.8}}$&70.8$_{\pm\textrm{11.4}}$&60.8$_{\pm\textrm{11.7}}$&62.8$_{\pm\textrm{10.3}}$&50.9$_{\pm\textrm{4.3}}$&61.50\\
&\small\textbf{Contextual Calibration}&70.8$_{\pm\textrm{11.2}}$&60.0$_{\pm\textrm{8.1}}$&70.5$_{\pm\textrm{9.0}}$&59.6$_{\pm\textrm{6.7}}$&70.0$_{\pm\textrm{7.0}}$&43.6$_{\pm\textrm{4.5}}$&62.43\\
&\small\textbf{Noisy Channel}&80.9$_{\pm\textrm{3.1}}$&60.5$_{\pm\textrm{8.7}}$&66.4$_{\pm\textrm{6.0}}$&83.9$_{\pm\textrm{2.3}}$&79.0$_{\pm\textrm{2.9}}$&40.9$_{\pm\textrm{9.3}}$&68.58\\
&\small\textbf{$k$NN Prompting}&87.1$_{\pm\textrm{6.2}}$&73.5$_{\pm\textrm{7.7}}$&66.4$_{\pm\textrm{11.7}}$&71.2$_{\pm\textrm{17.7}}$&82.9$_{\pm\textrm{3.0}}$&51.6$_{\pm\textrm{11.2}}$&\textbf{72.14}\\
&\small\textbf{$k$NN Prompting (Partial)}&85.9$_{\pm\textrm{6.7}}$&70.5$_{\pm\textrm{10.3}}$&67.4$_{\pm\textrm{12.0}}$&68.1$_{\pm\textrm{20.2}}$&82.0$_{\pm\textrm{3.6}}$&52.2$_{\pm\textrm{7.9}}$&71.00\\
\midrule
\multirow{5}*{$m=8$}&\small\textbf{ICL}&57.8$_{\pm\textrm{5.6}}$&66.2$_{\pm\textrm{13.0}}$&77.2$_{\pm\textrm{11.0}}$&66.0$_{\pm\textrm{11.5}}$&61.5$_{\pm\textrm{6.5}}$&50.9$_{\pm\textrm{6.1}}$&63.27\\
&\small\textbf{Contextual Calibration}&68.5$_{\pm\textrm{7.9}}$&64.5$_{\pm\textrm{9.9}}$&72.7$_{\pm\textrm{11.1}}$&64.9$_{\pm\textrm{6.8}}$&68.2$_{\pm\textrm{8.8}}$&44.0$_{\pm\textrm{5.2}}$&63.80\\
&\small\textbf{Noisy Channel}&82.0$_{\pm\textrm{2.1}}$&62.5$_{\pm\textrm{5.7}}$&70.1$_{\pm\textrm{4.2}}$&85.0$_{\pm\textrm{2.1}}$&79.2$_{\pm\textrm{2.1}}$&41.6$_{\pm\textrm{9.5}}$&70.06\\
&\small\textbf{$k$NN Prompting}&88.9$_{\pm\textrm{2.1}}$&77.7$_{\pm\textrm{5.8}}$&72.5$_{\pm\textrm{12.4}}$&75.4$_{\pm\textrm{12.1}}$&84.6$_{\pm\textrm{1.7}}$&63.7$_{\pm\textrm{5.5}}$&\textbf{77.13}\\
&\small\textbf{$k$NN Prompting (Partial)}&88.9$_{\pm\textrm{2.3}}$&69.2$_{\pm\textrm{11.1}}$&67.8$_{\pm\textrm{15.9}}$&72.8$_{\pm\textrm{12.7}}$&84.3$_{\pm\textrm{2.2}}$&54.6$_{\pm\textrm{3.7}}$&72.92\\
\midrule
\multirow{5}*{$m=16$}&\small\textbf{ICL}&67.7$_{\pm\textrm{10.8}}$&75.5${_{\pm\textrm{11.4}}^\dagger}$&77.6$_{\pm\textrm{7.6}}$&73.3${_{\pm\textrm{11.9}}^\dagger}$&61.8${_{\pm\textrm{5.6}}^\dagger}$&52.0${_{\pm\textrm{5.1}}^\dagger}$&67.97\\
&\small\textbf{Contextual Calibration}&75.7$_{\pm\textrm{7.1}}$&59.7${_{\pm\textrm{6.5}}^\dagger}$&75.2$_{\pm\textrm{8.0}}$&73.0${_{\pm\textrm{7.6}}^\dagger}$&73.1${_{\pm\textrm{7.0}}^\dagger}$&46.5${_{\pm\textrm{6.2}}^\dagger}$&67.21\\
&\small\textbf{Noisy Channel}&84.4$_{\pm\textrm{1.4}}$&62.4${_{\pm\textrm{7.2}}^\dagger}$&70.4$_{\pm\textrm{6.2}}$&83.7${_{\pm\textrm{3.3}}^\dagger}$&79.6${_{\pm\textrm{2.7}}^\dagger}$&54.2${_{\pm\textrm{7.8}}^\dagger}$&72.44\\
&\small\textbf{$k$NN Prompting}&88.8$_{\pm\textrm{1.6}}$&80.9$_{\pm\textrm{4.0}}$&68.2$_{\pm\textrm{7.6}}$&80.1$_{\pm\textrm{4.7}}$&84.8$_{\pm\textrm{2.7}}$&70.0$_{\pm\textrm{3.9}}$&\textbf{78.80}\\
&\small\textbf{$k$NN Prompting (Partial)}&89.7$_{\pm\textrm{2.5}}$&71.4$_{\pm\textrm{9.8}}$&60.5$_{\pm\textrm{12.3}}$&79.8$_{\pm\textrm{5.4}}$&84.8$_{\pm\textrm{3.0}}$&55.8$_{\pm\textrm{4.0}}$&73.67\\
\midrule
\multirow{5}*{$m=32$}&\small\textbf{ICL}&66.5${_{\pm\textrm{10.4}}^\dagger}$&70.1${_{\pm\textrm{9.6}}^\dagger}$&77.4$_{\pm\textrm{8.2}}$&67.7${_{\pm\textrm{9.6}}^\dagger}$&63.6${_{\pm\textrm{8.7}}^\dagger}$&52.0${_{\pm\textrm{2.1}}^\dagger}$&66.22\\
&\small\textbf{Contextual Calibration}&76.9${_{\pm\textrm{7.5}}^\dagger}$&58.6${_{\pm\textrm{9.2}}^\dagger}$&76.5$_{\pm\textrm{7.9}}$&78.5${_{\pm\textrm{7.8}}^\dagger}$&71.2${_{\pm\textrm{7.4}}^\dagger}$&44.3${_{\pm\textrm{3.2}}^\dagger}$&67.66\\
&\small\textbf{Noisy Channel}&84.8${_{\pm\textrm{0.9}}^\dagger}$&61.1${_{\pm\textrm{3.9}}^\dagger}$&70.8$_{\pm\textrm{5.2}}$&82.5${_{\pm\textrm{2.4}}^\dagger}$&80.0${_{\pm\textrm{1.8}}^\dagger}$&47.6${_{\pm\textrm{9.1}}^\dagger}$&71.14\\
&\small\textbf{$k$NN Prompting}&89.0$_{\pm\textrm{1.9}}$&83.2$_{\pm\textrm{3.9}}$&69.3$_{\pm\textrm{7.9}}$&77.8$_{\pm\textrm{5.6}}$&85.0$_{\pm\textrm{2.0}}$&73.5$_{\pm\textrm{3.9}}$&\textbf{79.64}\\
&\small\textbf{$k$NN Prompting (Partial)}&86.7$_{\pm\textrm{5.0}}$&65.9$_{\pm\textrm{12.9}}$&64.4$_{\pm\textrm{12.1}}$&74.2$_{\pm\textrm{7.4}}$&83.9$_{\pm\textrm{3.8}}$&59.4$_{\pm\textrm{4.0}}$&72.41\\
\bottomrule
\end{tabular}}
\caption{Results under few-shot scenario. Calibration-based baselines are reproduced using their released code\protect\footnotemark\protect\footnotemark. $\dagger$ denotes necessary truncation. The overall scaling trend is accordingly visualized in Figure~\ref{fig:fewshotscaling} and Appendix~\ref{appendix:fewshotscalingperdataset}. \textbf{Partial} means only label-words distribution is utilized.}
\label{table:mainresultsfewshot}
\end{table*}
\addtocounter{footnote}{-1}\footnotetext{\url{https://github.com/tonyzhaozh/few-shot-learning}}\addtocounter{footnote}{1}\footnotetext{\url{https://github.com/shmsw25/Channel-LM-Prompting}, as there exists slightly difference of prompt templates, we report their best template out of four.}

\paragraph{Superiority of Whole LM Distribution} Calibration-based methods~\citep{pmlr-v139-zhao21c} as well as standard ICL only access label words instead of whole LM distribution, which is inferior in two aspects: 1) loss of information. LLM always generate distribution over all words in the entire vocabulary, non-label words probabilities also reflect its understanding in certain perspectives; 2) multiple label words competing with each other. There exist various choices for label words but no oracle rules to select one, and alternative choices potentially compete with the selected label words, distorting the label space distribution. This is also referred to as surface form competition~\citep{holtzman-etal-2021-surface}.
In Table~\ref{table:mainresultsfewshot} we very this benefit by masking out non-label words (referred to as \textbf{Partial}).

\subsubsection{Data Utility Beyond the Context}\label{datautilitylowresource}
We then investigate a major advantage of $k$NN Prompting, which is scaling up to more training examples that otherwise would not fit in the context.
We increase $m$ to 128 and compare with: 1) ICL Ensemble which is an intuitive alternative to scale ICL up, and has been adopted in previous works~\citep{jiang-etal-2020-know}; 2) finetuning of standard PLM such as BERT or GPT Large, which could produce meaningful results with such amount of data.
For all methods, we append maximum $M_T$ examples into prompt $P$.
For ICL Ensemble, we split $\mathcal{T}$ into multiple non-overlap demonstration sets $\mathcal{T}=\mathcal{D}_1\cup\mathcal{D}_2\cup...\cup\mathcal{D}_N$ to construct different prompts, and ensemble their predictions.

Results in Table~\ref{table:lowresource} show that $k$NN Prompting continues to improve and outperform ICL and its ensemble respectively by \textbf{+16.96} and \textbf{+16.08} (0.8B model, average score).
Besides, the ensemble baseline is also very inefficient, assuming $M_T=8$, we need to query $128/8=16$ times for every test instance, and this keeps growing linearly if we use more training data, eventually becomes prohibitively inefficient and can not scale up either.
$k$NN Prompting also outperforms FT when the adopted LLM scales above \textbf{6B} (82.73).
By contrast, it would require LLM to scale above \textbf{30B} (82.45) using ICL Ensemble and even larger using standard ICL.

\definecolor{Gray}{gray}{0.9}
\newcommand*{\belowrulesepcolor}[1]{%
  \noalign{%
    \kern-\belowrulesep
    \begingroup
      \color{#1}%
      \hrule height\belowrulesep
    \endgroup
  }%
}
\newcommand*{\aboverulesepcolor}[1]{%
  \noalign{%
    \begingroup
      \color{#1}%
      \hrule height\aboverulesep
    \endgroup
    \kern-\aboverulesep
  }%
}
\newcolumntype{x}[1]{>{\centering\arraybackslash}p{#1}}
\begin{table*}[!tbp]
\centering
\resizebox{\columnwidth}{!}{
\begin{tabular}{{x{0.5cm}r|cccccccccc|c}}
\toprule
\multicolumn{2}{c|}{\textbf{Models \& Methods}}&\textbf{SST2}&\textbf{SUBJ}&\textbf{MPQA}&\textbf{AGNews}&\textbf{CB}&\textbf{CR}&\textbf{DBPedia}&\textbf{MR}&\textbf{RTE}&\textbf{TREC}&\textbf{AVG}\\
\midrule
\multicolumn{2}{c|}{\small\textbf{BERT Large FT}}&88.3$_{\pm\textrm{1.4}}$&90.7$_{\pm\textrm{0.6}}$&74.5$_{\pm\textrm{5.0}}$&88.0$_{\pm\textrm{1.0}}$&78.6$_{\pm\textrm{3.6}}$&88.0$_{\pm\textrm{2.6}}$&95.1$_{\pm\textrm{1.8}}$&83.0$_{\pm\textrm{3.1}}$&58.1$_{\pm\textrm{1.5}}$&78.8$_{\pm\textrm{5.3}}$&82.31\\
\multicolumn{2}{c|}{\small\textbf{GPT Large FT}}&90.7$_{\pm\textrm{1.3}}$&86.1$_{\pm\textrm{1.7}}$&87.6$_{\pm\textrm{0.9}}$&88.3$_{\pm\textrm{1.5}}$&70.0$_{\pm\textrm{2.0}}$&86.7$_{\pm\textrm{13.2}}$&96.5$_{\pm\textrm{1.2}}$&86.2$_{\pm\textrm{1.0}}$&55.4$_{\pm\textrm{3.8}}$&71.2$_{\pm\textrm{2.2}}$&81.88\\
\midrule
\multirow{3}*{\textbf{0.8B}}&\small\textbf{ICL}&63.4$_{\pm\textrm{7.3}}$&58.9$_{\pm\textrm{8.7}}$&70.5$_{\pm\textrm{5.2}}$&61.7$_{\pm\textrm{15.4}}$&45.0$_{\pm\textrm{9.1}}$&83.3$_{\pm\textrm{13.7}}$&59.9$_{\pm\textrm{11.5}}$&77.0$_{\pm\textrm{15.7}}$&53.6$_{\pm\textrm{3.1}}$&54.4$_{\pm\textrm{1.7}}$&62.77\\
&\small\textbf{ICL Ensemble}&63.0$_{\pm\textrm{6.4}}$&57.7$_{\pm\textrm{10.3}}$&69.1$_{\pm\textrm{6.2}}$&67.4$_{\pm\textrm{2.9}}$&41.1$_{\pm\textrm{3.1}}$&83.8$_{\pm\textrm{11.5}}$&67.8$_{\pm\textrm{3.7}}$&72.7$_{\pm\textrm{12.3}}$&55.1$_{\pm\textrm{3.8}}$&59.0$_{\pm\textrm{3.7}}$&63.65\\
&\small\textbf{$\bm k$NN Prompting}&84.5$_{\pm\textrm{5.3}}$&85.8$_{\pm\textrm{1.6}}$&83.1$_{\pm\textrm{0.8}}$&84.5$_{\pm\textrm{1.3}}$&62.1$_{\pm\textrm{3.4}}$&89.7$_{\pm\textrm{0.6}}$&95.8$_{\pm\textrm{0.5}}$&84.0$_{\pm\textrm{1.8}}$&53.6$_{\pm\textrm{3.2}}$&74.2$_{\pm\textrm{4.4}}$&\textbf{79.73}\\
\midrule
\multirow{3}*{\textbf{1.5B}}&\small\textbf{ICL}&81.3$_{\pm\textrm{5.4}}$&64.1$_{\pm\textrm{11.3}}$&75.2$_{\pm\textrm{8.8}}$&72.7$_{\pm\textrm{18.5}}$&60.7$_{\pm\textrm{2.8}}$&66.2$_{\pm\textrm{16.7}}$&83.5$_{\pm\textrm{3.8}}$&72.2$_{\pm\textrm{13.9}}$&53.0$_{\pm\textrm{1.7}}$&54.2$_{\pm\textrm{4.9}}$&68.31\\
&\small\textbf{ICL Ensemble}&83.4$_{\pm\textrm{1.9}}$&63.4$_{\pm\textrm{11.5}}$&75.0$_{\pm\textrm{7.2}}$&81.1$_{\pm\textrm{0.7}}$&52.9$_{\pm\textrm{8.1}}$&63.4$_{\pm\textrm{17.2}}$&83.7$_{\pm\textrm{1.3}}$&73.8$_{\pm\textrm{15.1}}$&55.9$_{\pm\textrm{1.8}}$&59.6$_{\pm\textrm{2.3}}$&69.21\\
&\small\textbf{$\bm k$NN Prompting}&86.3$_{\pm\textrm{2.9}}$&83.8$_{\pm\textrm{2.1}}$&82.3$_{\pm\textrm{1.5}}$&87.2$_{\pm\textrm{0.4}}$&64.6$_{\pm\textrm{7.9}}$&88.9$_{\pm\textrm{2.4}}$&96.5$_{\pm\textrm{0.7}}$&86.4$_{\pm\textrm{0.8}}$&51.1$_{\pm\textrm{1.7}}$&74.0$_{\pm\textrm{2.9}}$&\textbf{80.12}\\
\midrule
\multirow{3}*{\textbf{2.7B}}&\small\textbf{ICL}&89.9$_{\pm\textrm{4.5}}$&77.7$_{\pm\textrm{5.8}}$&84.5$_{\pm\textrm{2.1}}$&78.8$_{\pm\textrm{4.0}}$&51.1$_{\pm\textrm{6.1}}$&92.6$_{\pm\textrm{0.6}}$&88.8$_{\pm\textrm{1.3}}$&92.5$_{\pm\textrm{1.0}}$&52.4$_{\pm\textrm{4.2}}$&64.6$_{\pm\textrm{3.0}}$&77.29\\
&\small\textbf{ICL Ensemble}&90.2$_{\pm\textrm{3.7}}$&77.7$_{\pm\textrm{3.4}}$&86.6$_{\pm\textrm{1.9}}$&78.4$_{\pm\textrm{1.8}}$&54.6$_{\pm\textrm{1.0}}$&93.4$_{\pm\textrm{1.1}}$&89.3$_{\pm\textrm{0.9}}$&92.4$_{\pm\textrm{0.8}}$&54.3$_{\pm\textrm{2.3}}$&78.8$_{\pm\textrm{2.3}}$&79.57\\
&\small\textbf{$\bm k$NN Prompting}&93.4$_{\pm\textrm{1.3}}$&87.5$_{\pm\textrm{2.0}}$&83.3$_{\pm\textrm{3.9}}$&86.7$_{\pm\textrm{2.4}}$&57.1$_{\pm\textrm{3.8}}$&90.2$_{\pm\textrm{2.0}}$&98.6$_{\pm\textrm{0.6}}$&91.4$_{\pm\textrm{1.5}}$&53.4$_{\pm\textrm{5.6}}$&80.8$_{\pm\textrm{2.0}}$&\textbf{82.25}\\
\midrule
\multirow{3}*{\textbf{6B}}&\small\textbf{ICL}&92.7$_{\pm\textrm{1.8}}$&81.6$_{\pm\textrm{4.6}}$&86.2$_{\pm\textrm{1.9}}$&68.8$_{\pm\textrm{8.2}}$&52.5$_{\pm\textrm{9.7}}$&92.0$_{\pm\textrm{2.6}}$&89.8$_{\pm\textrm{0.9}}$&91.6$_{\pm\textrm{1.1}}$&55.0$_{\pm\textrm{1.6}}$&62.0$_{\pm\textrm{7.3}}$&77.19\\
&\small\textbf{ICL Ensemble}&93.2$_{\pm\textrm{0.9}}$&85.6$_{\pm\textrm{3.8}}$&87.7$_{\pm\textrm{2.9}}$&71.1$_{\pm\textrm{5.5}}$&45.7$_{\pm\textrm{2.0}}$&93.4$_{\pm\textrm{1.5}}$&90.9$_{\pm\textrm{1.0}}$&92.0$_{\pm\textrm{0.2}}$&54.8$_{\pm\textrm{1.3}}$&70.3$_{\pm\textrm{1.8}}$&78.47\\
&\small\textbf{$\bm k$NN Prompting}&92.2$_{\pm\textrm{1.2}}$&87.4$_{\pm\textrm{1.7}}$&85.2$_{\pm\textrm{2.5}}$&87.4$_{\pm\textrm{1.6}}$&63.6$_{\pm\textrm{4.1}}$&90.6$_{\pm\textrm{2.2}}$&98.5$_{\pm\textrm{0.4}}$&91.0$_{\pm\textrm{1.5}}$&55.6$_{\pm\textrm{2.0}}$&75.8$_{\pm\textrm{3.1}}$&\textbf{82.73}\\
\midrule
\multirow{3}*{\textbf{13B}}&\small\textbf{ICL}&89.0$_{\pm\textrm{4.3}}$&91.3$_{\pm\textrm{2.4}}$&78.4$_{\pm\textrm{7.2}}$&78.1$_{\pm\textrm{5.6}}$&53.2$_{\pm\textrm{4.4}}$&93.4$_{\pm\textrm{1.1}}$&92.2$_{\pm\textrm{2.4}}$&89.9$_{\pm\textrm{2.2}}$&55.8$_{\pm\textrm{3.0}}$&55.5$_{\pm\textrm{6.1}}$&77.68\\
&\small\textbf{ICL Ensemble}&88.2$_{\pm\textrm{4.7}}$&90.7$_{\pm\textrm{1.2}}$&78.4$_{\pm\textrm{4.8}}$&82.6$_{\pm\textrm{1.5}}$&62.1$_{\pm\textrm{3.9}}$&94.0$_{\pm\textrm{0.7}}$&94.6$_{\pm\textrm{1.0}}$&89.0$_{\pm\textrm{2.5}}$&56.2$_{\pm\textrm{2.0}}$&59.8$_{\pm\textrm{3.7}}$&79.56\\
&\small\textbf{$\bm k$NN Prompting}&94.8$_{\pm\textrm{0.6}}$&90.1$_{\pm\textrm{1.5}}$&86.2$_{\pm\textrm{3.8}}$&87.4$_{\pm\textrm{1.7}}$&78.9$_{\pm\textrm{4.8}}$&89.6$_{\pm\textrm{1.1}}$&98.9$_{\pm\textrm{0.5}}$&92.0$_{\pm\textrm{0.5}}$&58.2$_{\pm\textrm{4.4}}$&73.1$_{\pm\textrm{2.9}}$&\textbf{84.93}\\
\midrule
\multirow{3}*{\textbf{30B}}&\small\textbf{ICL}&90.8$_{\pm\textrm{4.1}}$&83.5$_{\pm\textrm{8.7}}$&80.7$_{\pm\textrm{1.9}}$&74.8$_{\pm\textrm{4.6}}$&64.6$_{\pm\textrm{8.3}}$&87.7$_{\pm\textrm{3.9}}$&93.3$_{\pm\textrm{0.4}}$&93.4$_{\pm\textrm{1.0}}$&61.6$_{\pm\textrm{2.7}}$&71.7$_{\pm\textrm{2.7}}$&80.21\\
&\small\textbf{ICL Ensemble}&92.6$_{\pm\textrm{2.3}}$&84.1$_{\pm\textrm{4.6}}$&79.9$_{\pm\textrm{1.9}}$&78.3$_{\pm\textrm{2.4}}$&67.1$_{\pm\textrm{4.8}}$&88.1$_{\pm\textrm{2.9}}$&93.8$_{\pm\textrm{1.0}}$&93.4$_{\pm\textrm{1.0}}$&65.1$_{\pm\textrm{4.6}}$&82.0$_{\pm\textrm{2.5}}$&82.45\\
&\small\textbf{$\bm k$NN Prompting}&94.3$_{\pm\textrm{0.9}}$&92.7$_{\pm\textrm{1.7}}$&84.5$_{\pm\textrm{0.9}}$&87.1$_{\pm\textrm{1.4}}$&70.7$_{\pm\textrm{8.2}}$&91.0$_{\pm\textrm{1.2}}$&98.8$_{\pm\textrm{0.5}}$&93.1$_{\pm\textrm{1.5}}$&61.7$_{\pm\textrm{4.8}}$&79.8$_{\pm\textrm{2.0}}$&\textbf{85.38}\\
\bottomrule
\end{tabular}}
\caption{Results under low resource scenario ($m=128$). Compared with FT and ICL Ensemble.}
\setlength{\belowcaptionskip}{-20pt}
\label{table:lowresource}
\end{table*}

{\setlength\intextsep{5pt}
\begin{figure}[h!]
    \centering
    \begin{minipage}{0.49\textwidth}
        \centering
        \includegraphics[width=0.95\textwidth]{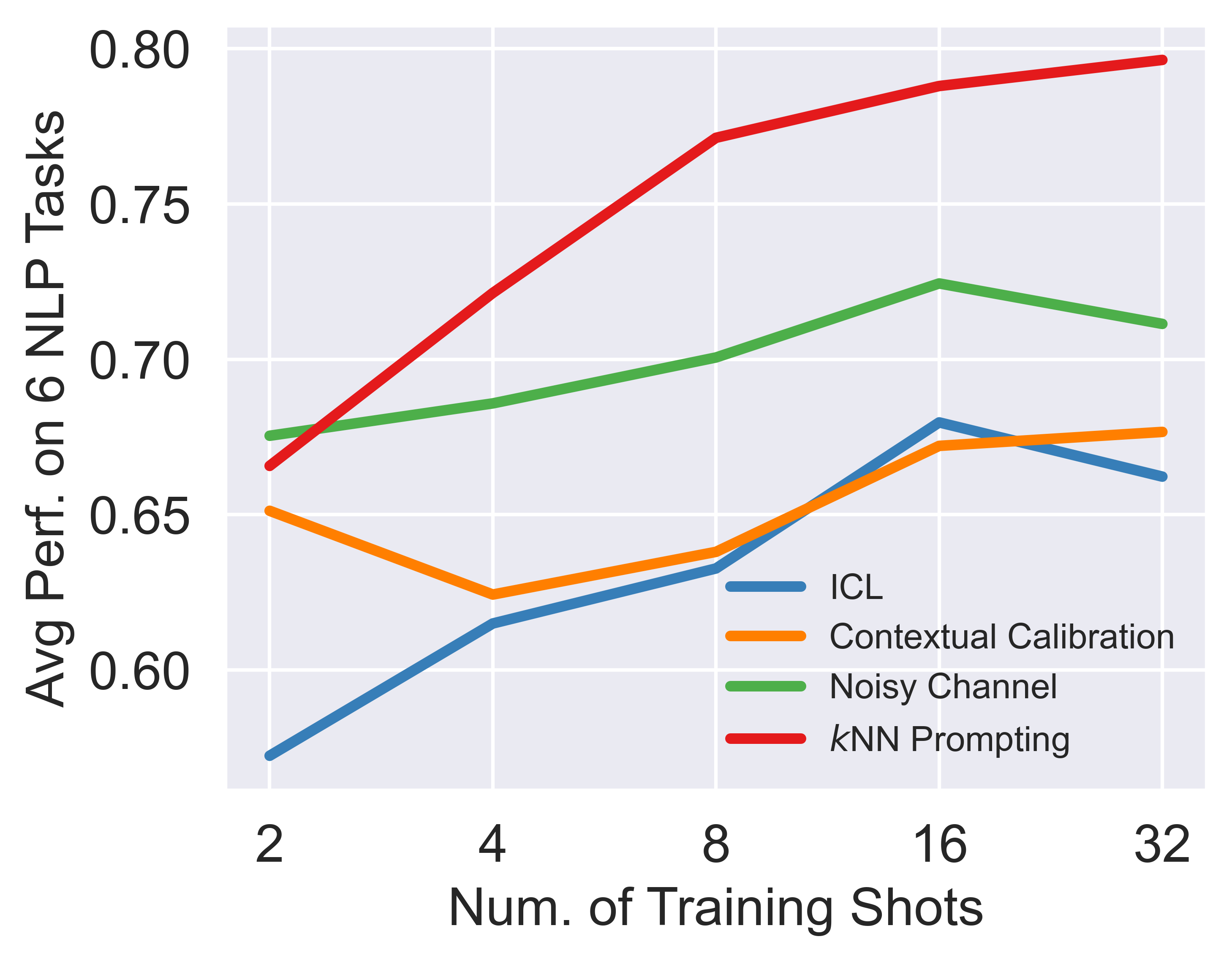}
        \caption{Data scaling under few shot scenario. Compared with calibration-based methods.}
        \label{fig:fewshotscaling}
    \end{minipage}
    \begin{minipage}{0.49\textwidth}
        \centering
        \includegraphics[width=0.95\linewidth]{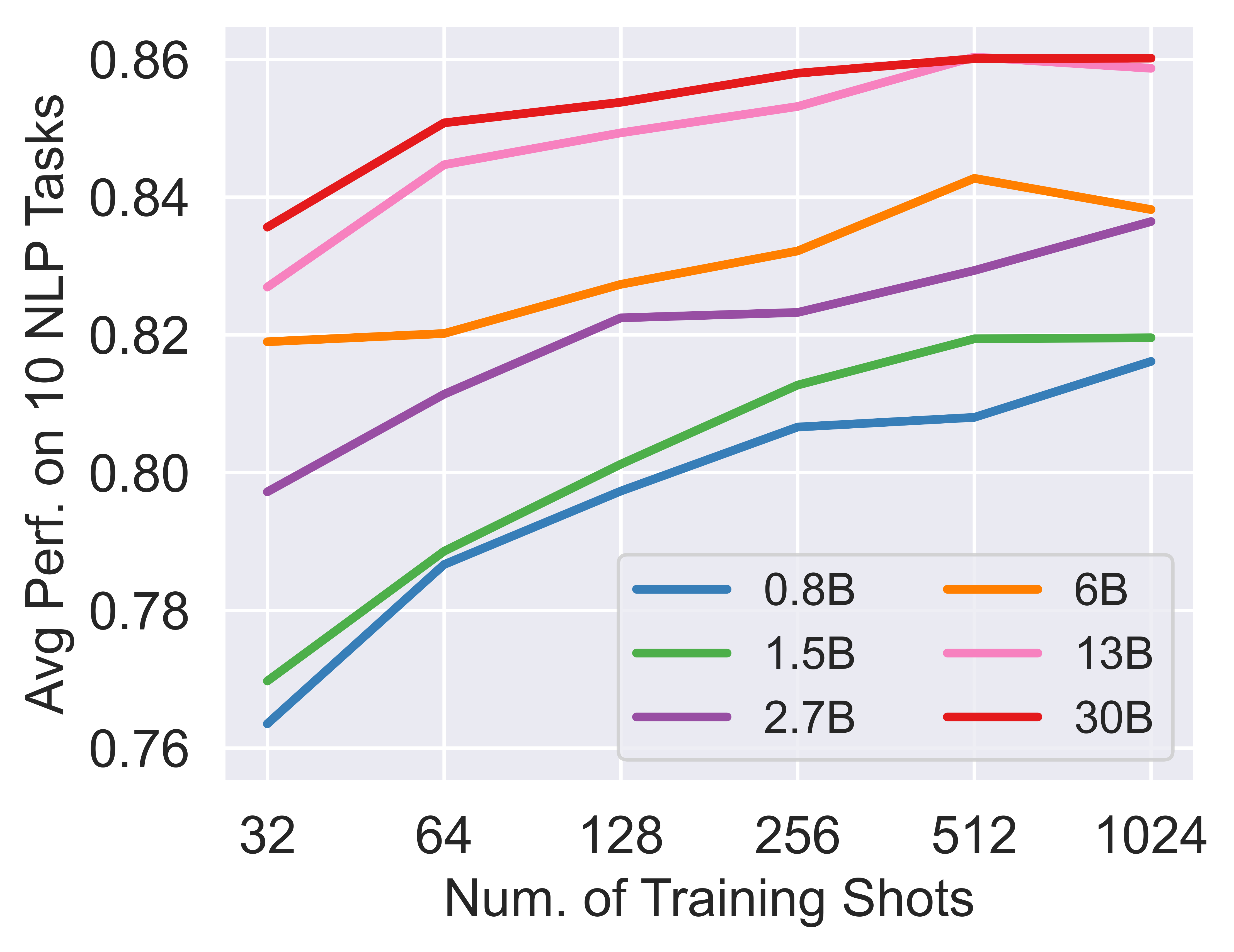}
        \caption{Data scaling under fully supervised scenario. Conducted across various LLM scales.}
        \label{fig:fullysupervisedscaling}
    \end{minipage}
\end{figure}
}

\subsubsection{Continually Scaling Up to Thousands of Training Data}
We now fully scale data up to thousands of training examples and provide extensive results across model scales to observe their overall scaling performance.
Figure~\ref{fig:fullysupervisedscaling} shows $k$NN prompting can continually generalize across the tested range to provide effective data utility, re-enabling the power law under gradient-free paradigm of LLM deployment.
The full results can be found in Appendix~\ref{appendix:fullresults}.
With only 32 shots training data, $k$NN prompting dramatically improves ICL by \textbf{+13.58} in average score at its most (0.8 B), and achieves absolute improvements up to \textbf{+18.84} under fully supervised setting.
With the largest model OPT 30B, it achieves a best performance of 86.02.

\paragraph{Comparison to Demonstration Selection} A line of related works try to utilize available training data by firstly retrieving the most relevant ones from the entire training set, then selectively composing the prompt before querying LLM~\citep{liu-etal-2022-makes,rubin-etal-2022-learning}.
We since reproduce such methods according to~\citet{liu-etal-2022-makes}.
We employ state-of-the-art general-purpose sentence encoders\footnote{Note that~\citet{liu-etal-2022-makes} also employ RoBERTa model finetuned on SST2, but this overlaps with our selected benchmark and does not generalize to other tasks beyond sentiment classification.} to represent test and training instance, and compute their cosine similarity, the most similar $M_T$ examples are selected to construct prompt $P$.
These retrieving models include BM25~\citep{10.1145/2682862.2682863}, Sentence-BERT~\citep{reimers-gurevych-2019-sentence}, SimCSE~\citep{gao-etal-2021-simcse} and Trans-Encoder~\citep{liu2022transencoder}.

Figure~\ref{fig:comparisontoretrievetocompose} shows that although such methods indeed exhibits marginal scaling benefits, they are nowhere near competitive against $k$NN Prompting.
Full results are listed in Appendix~\ref{appendix:comparetopreretrieval}.
To further solidify this conclusion, we push PromptCompose to an extreme situation trying to approximate its upper-bound.
As such methods ultimately resort to compose the prompt, it should be bounded by the best composition scheme from finite compositions.
We thus search for 1,000 prompts with different examples and report their best run.
In Table~\ref{table:comparisontopromptcomposition} we find $k$NN Prompting performs on par with or even surpasses such upper-bound approximation.
In conclusion, $k$NN Prompting essentially makes better use of training examples than PromptCompose as the latter still only refer to in-context examples during LLM inference while most training data are discarded beforehand.
 
\newcommand{\meanwithstd}[2]{#1$_{\pm\textrm{#2}}$}

\begin{figure}
\centering
\begin{minipage}{0.45\textwidth}
\centering
 \includegraphics[width=\textwidth]{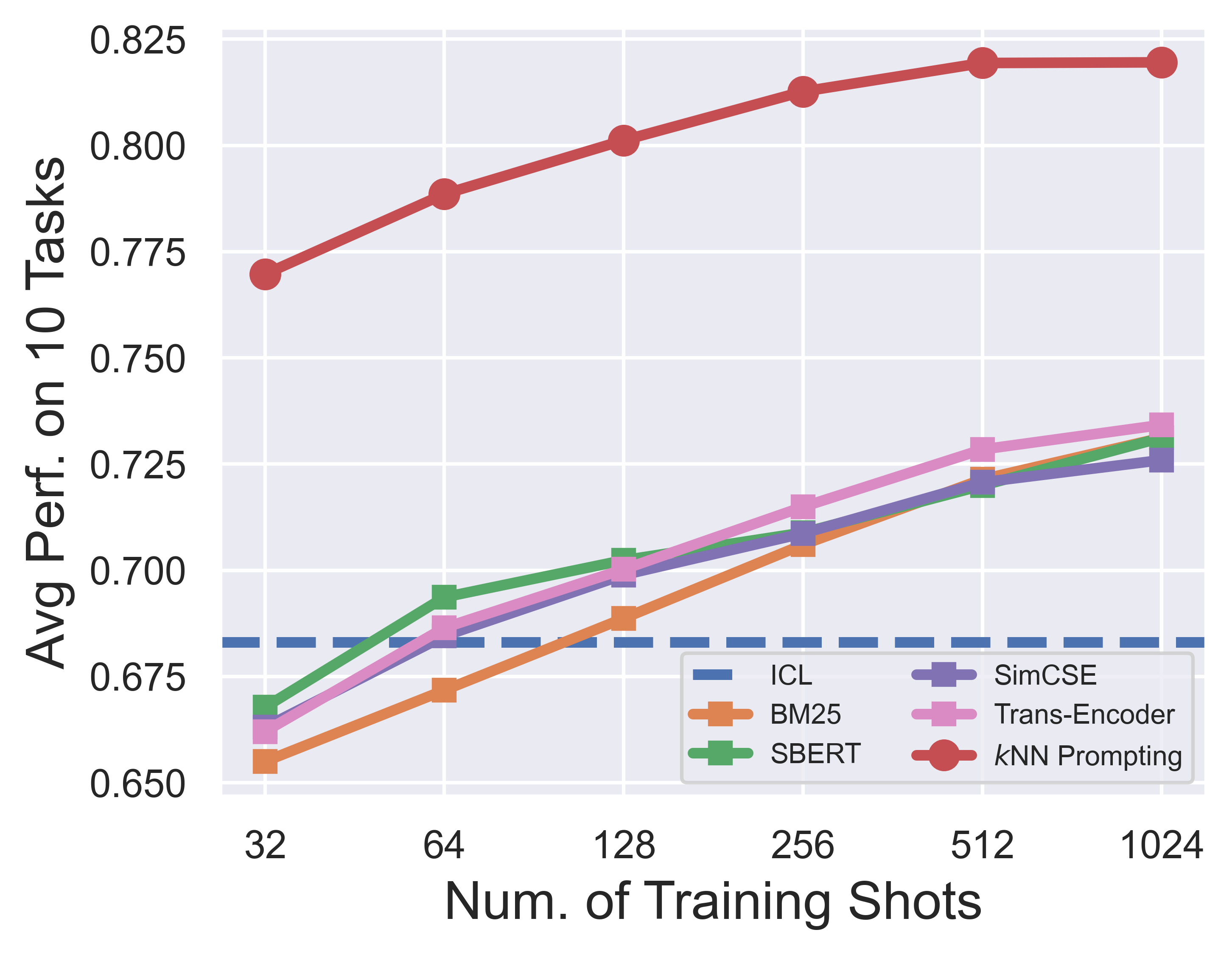}
\captionsetup{skip=0pt}
\caption{cf. Demonstration Selection.}
\label{fig:comparisontoretrievetocompose}
\end{minipage}
\hfill
\begin{minipage}{0.52\textwidth}
\centering
\resizebox{\columnwidth}{!}{
\begin{tabular}{{l|ccc}}
\toprule
\textbf{Method}&\textbf{SST2}&\textbf{SUBJ}&\textbf{MPQA}\\
\midrule
\small\textbf{ICL Baseline}&\meanwithstd{81.3}{5.4}&\meanwithstd{64.1}{11.3}&\meanwithstd{75.2}{8.8}\\
\small\textbf{DemonSelection (upper)}&\textbf{92.6} (1\textperthousand)&86.0 (3\textperthousand)&\textbf{87.5} (1\textperthousand)\\
\small\textbf{$k$NN Prompting}&\meanwithstd{88.2}{1.0}&\meanwithstd{\textbf{88.4}}{1.8}&\meanwithstd{84.1}{1.2}\\
\midrule
&\textbf{MR}&\textbf{CR}&\textbf{TREC}\\
\midrule
\small\textbf{ICL Baseline}&\meanwithstd{72.2}{13.9}&\meanwithstd{66.2}{16.7}&\meanwithstd{54.2}{4.9}\\
\small\textbf{DemonSelection (upper)}&\textbf{88.7} (1\textperthousand)&\textbf{88.7} (1\textperthousand)&73.0 (1\textperthousand)\\
\small\textbf{$k$NN Prompting}&\meanwithstd{84.4}{1.5}&\meanwithstd{86.7}{1.3}&\meanwithstd{\textbf{83.0}}{1.4}\\
\bottomrule
\end{tabular}}
\captionof{table}{Comparison to \textbf{upper-bound} of Demonstration Selection. 1\textperthousand~inside the parentheses means the result can be achieved in 1 runs out of 1,000 searches.}
\label{table:comparisontopromptcomposition}
\end{minipage}
\end{figure}

\subsection{Analyses and Explanation}

\subsubsection{Robustness w.r.t. Different Training Examples}
It is previously found that vanilla ICL suffers from severe instability~\citep{pmlr-v139-zhao21c}.
Figure~\ref{fig:robustness} is produced with 10 different seeds, which results in different choices and permutations of training data.
We show that $k$NN Prompting significantly improves the robustness.
Besides, the performance becomes more robust with increasing anchor set size.
On 10 investigated tasks, the standard deviation of $k$NN Prompting (3.83) is significantly smaller than ICL (9.14)\footnote{Calculated using Table~\ref{table:fullysupervisedscaling} statistics with 32 shot setting, 0.8B model}.

{\setlength\intextsep{5pt}
\begin{figure}[h!]
    \centering
    \begin{minipage}{0.45\textwidth}
        \centering
        \includegraphics[width=0.96\textwidth]{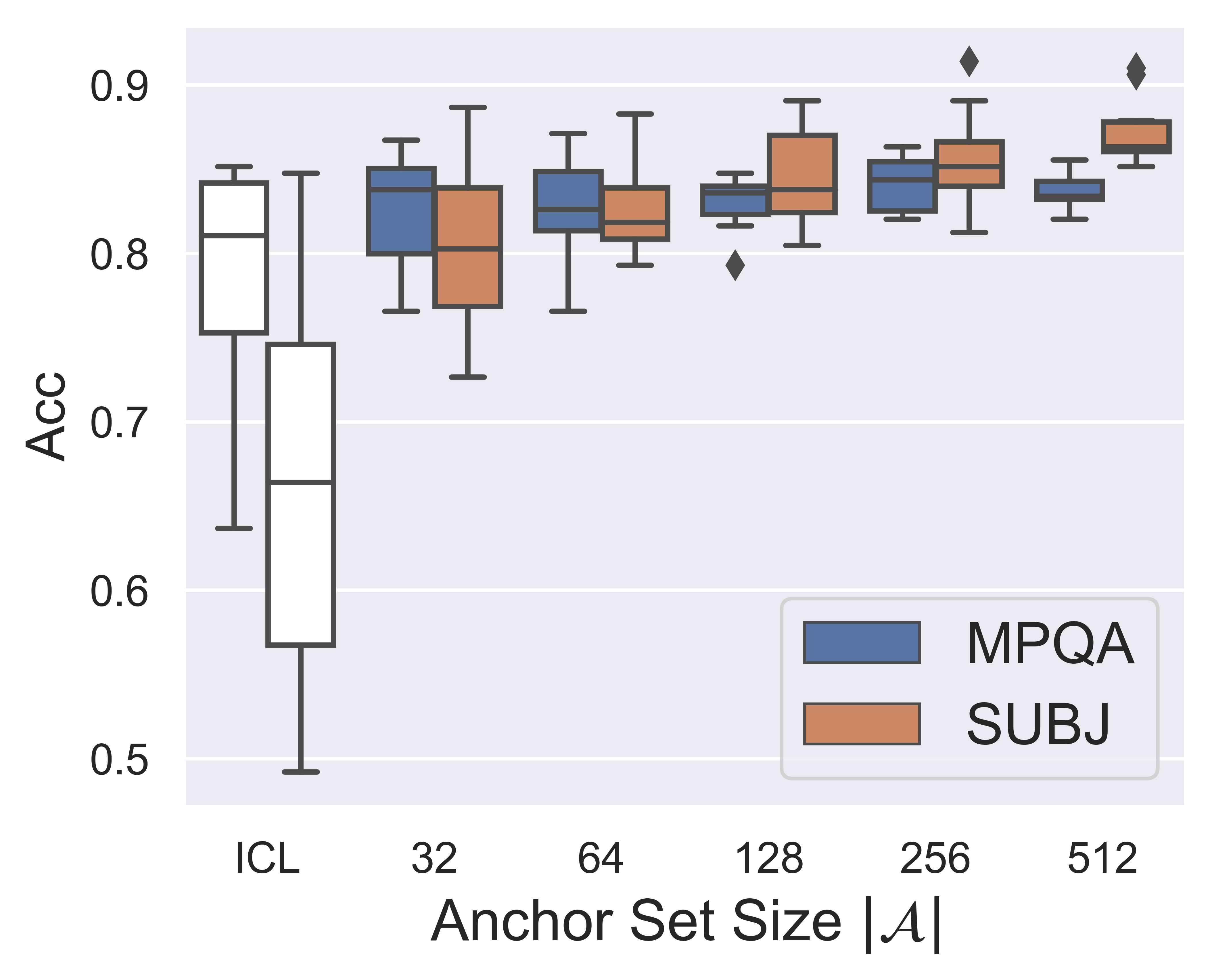}
        \captionsetup{skip=0pt}
        \caption{Robustness.}
        \label{fig:robustness}
    \end{minipage}
    \hfill
    \begin{minipage}{0.45\textwidth}
        \centering
        \includegraphics[width=0.9\linewidth]{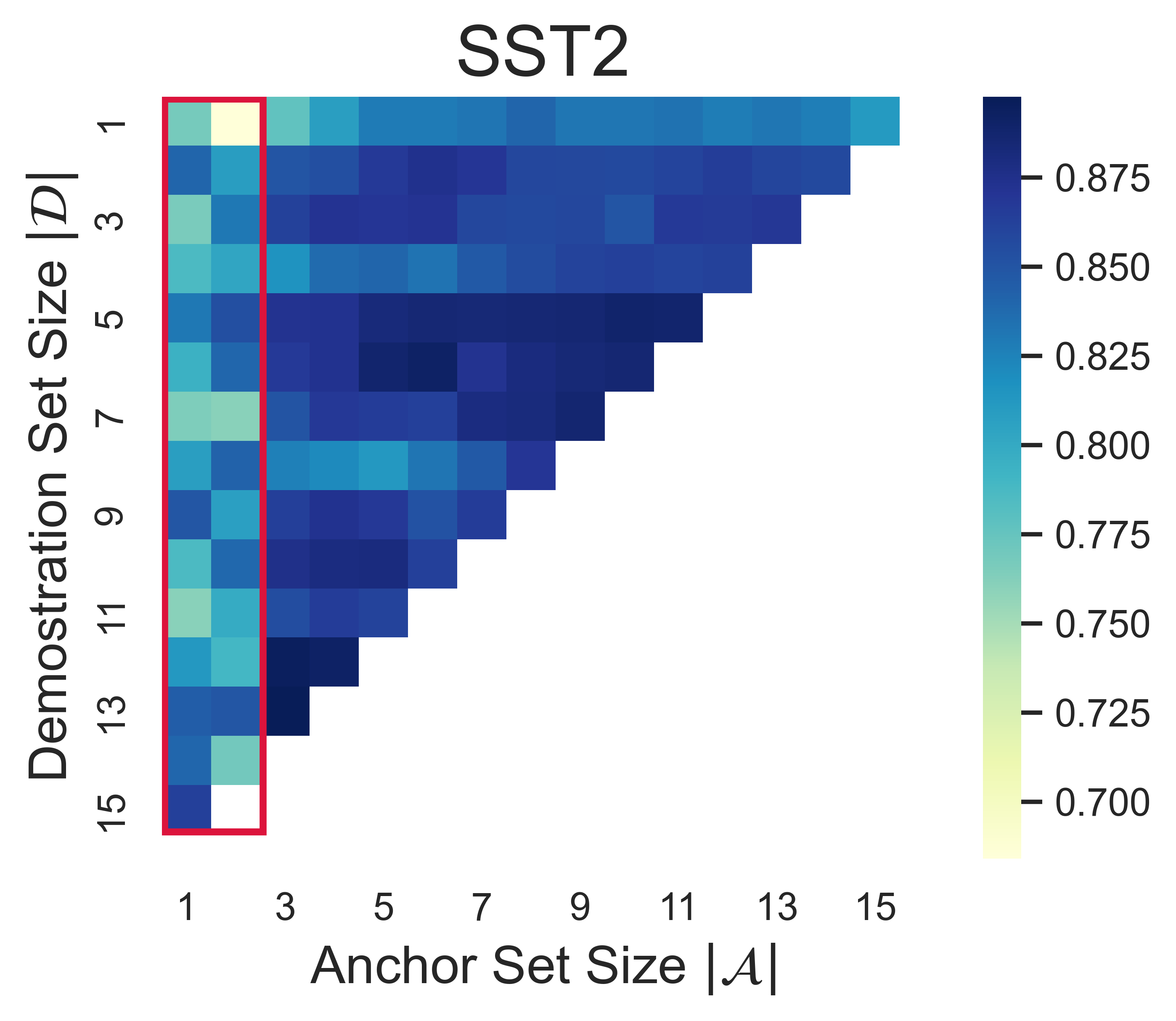}
        \captionsetup{skip=0pt}
        \caption{Split strategy.}
        \label{fig:demoanchorsplit}
    \end{minipage}
\end{figure}
}

\subsubsection{Split Strategy Between Demonstrations And Anchors}\label{sec:splitstrategy}
For fully supervised scenario where $m \gg M_T$, we can simply set $|\mathcal{D}|$ to its maximum $M_T$.
Otherwise, we might need to deliberate the trade-off between the demonstration set and anchor set.
In Figure~\ref{fig:demoanchorsplit} we search through all possible combinations given that $|\mathcal{A}|+|\mathcal{D}|\leq M_T$.
We see the left part of the heatmap generally performs inferior, i.e., $|\mathcal{A}|\in\{1,2\}$.
This means a larger $|\mathcal{A}|$ contributes more to the performance while the choice for $|\mathcal{D}|$ is relatively more robust.
The conclusion also corresponds to few shot results that anchor set yields better data utility than context concatenation.
More datasets are also provided in Appendix~\ref{appendix:splitofdemoandanchor}.

\subsubsection{Qualitative Analyses and Reasons of Effectiveness}\label{ablation:tsne}
\definecolor{Ecru}{HTML}{c49c94}
\definecolor{Red}{HTML}{d62728}
We first formally organize the explanation as follows according to Figure~\ref{fig:casestudy_dbpedia}:
\begin{itemize}[leftmargin=*]
\item The output language modeling (LM) distribution of LLM is essentially not well aligned with task-specific label space, resulting in inferior performance (\textit{83.5 test accuracy}) of default ICL.
\item If we similarly perform inference on anchor set, we would expect to get approximately \textit{83.5 anchor accuracy} by assuming i.i.d. data distribution. However, we are already aware of each of their golden labels, which actually gives \textit{100 anchor accuracy}.
\item LM distribution is inferior for making direct predictions, but superior for matching examples because it entails distributional, delicate and comprehensive representations generated by LLM. $k$NN Prompting leverages such representations (\textit{83.5 accuracy}) only for matching, and refer to their golden labels (\textit{100 accuracy}) for predicting, thus successfully transfer part of the knowledge originating from anchor labels to test instances.
\end{itemize}

In the visualization, the representations generally exhibit partially clustered pattern, we can identify proportional examples that get entangled with different categories and crowded together (Case A, C, D), these confusing cases are likely to cause erroneous predictions in ICL and corresponds to underperformed 83.5 accuracy as mentioned above.
Specifically, case $A$ is an abstract about a novelist and should belong to category \textbf{\textcolor{Ecru}{artist}}, but it is easily confused with category \textbf{\textcolor{Red}{book}} using ICL because the context did mention books.
By contrast, $k$NN Prompting can correctly predict by referring to similar anchors that are also about novelists and their books (as listed in the table).
Some of the anchors are also incorrectly predicted as \textbf{\textcolor{Red}{book}}, but it no longer matters because $k$NN Prompting only use the distribution for nearest neighbor search but refer to golden labels for prediction.
Besides, as we can clearly know how the prediction is made, i.e., which anchor examples are referred, the proposed method also exhibits explainability as a further advantage.

{\setlength\intextsep{5pt}
\begin{figure*}[t!]
\centering
 \includegraphics[width=0.78\textwidth]{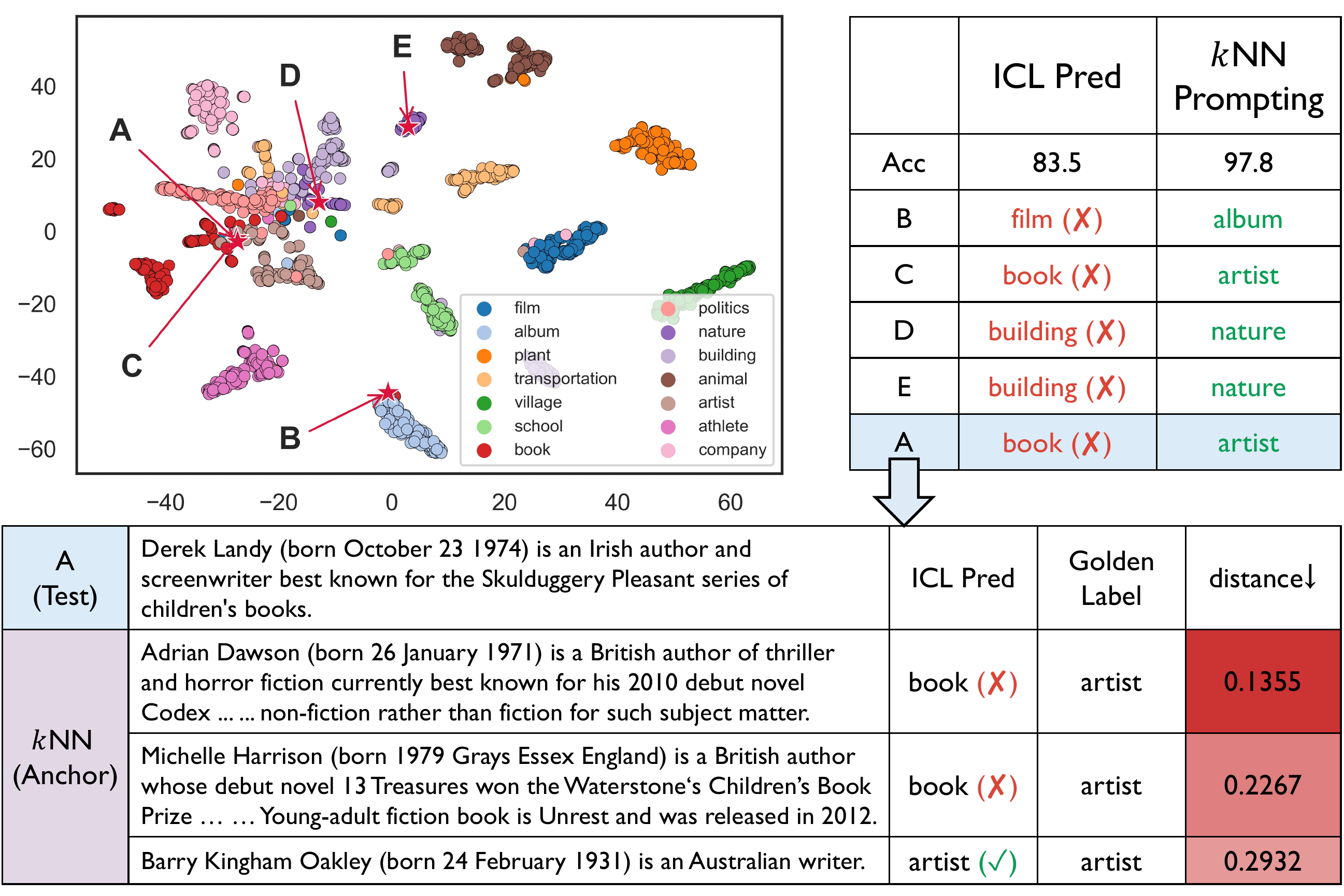}
\caption{t-SNE~\citep{JMLR:v9:vandermaaten08a} for anchors and test cases. Cases are randomly selected given that $k$NN Prompting outperforms ICL. DBPedia is an ontology classification task.}
\label{fig:casestudy_dbpedia}
\end{figure*}
}

\section{Related Works}
Large language models, since firstly scaled up to hundreds of billions parameters by~\citet{NEURIPS2020_1457c0d6} and followed by several others~\citep{rae2021scaling,zhang2022opt,chowdhery2022palm,cohen2022lamda,smith2022using}, have become the most prominent direction of NLP area.
Although these models exhibit surprisingly powerful and even emergent capabilities in a wide range of NLP tasks~\citep{wei2022emergent,hendrycks2020measuring,srivastava2022beyond}, they are prohibitively expensive for most researchers or users to train or even hold.
In-Context Learning, which suits LLM to various tasks while requires no training, therefore becomes the typical usage as is popularized by~\citet{NEURIPS2020_1457c0d6}.
Similar ideas of formulating target tasks into natural language sequences can also be found in earlier works~\citep{trinh2018simple,JMLR:v21:20-074}.

To better exploit LLM for various scenarios, it becomes a crucial problem to develop augmented methods for ICL~\citep{dong2022survey}.
\citet{xie2022an} provide theoretical explanations that formalize ICL as Bayesian inference.
\citet{dai2022can} reveal that ICL can be seen as implicit finetuning where LLM produces meta-gradients from in-context demonstrations to adapt the model behavior.
\citet{wei2022finetuned} and~\citet{sanh2022multitask} propose instruction tuning, which further pretrains LLM with a collection of downstream tasks in a shared prompting format.
\citet{min-etal-2022-metaicl} and~\citet{chen-etal-2022-meta} introduce meta-learning to better adapt LMs to ICL.
\citet{wei2022chain} and~\citet{kojima2022large} propose to augment the demonstrations with human-aided reasoning steps or hints, which surprisingly improved the performance for arithmetics and other reasoning tasks.
Closely related to this work, \citet{liu-etal-2022-makes} and~\citet{rubin-etal-2022-learning} propose to compose prompt $P$ by selecting most similar training examples.~\citet{pmlr-v139-zhao21c} and~\citet{min-etal-2022-noisy} propose to calibrate ICL prediction via either probing the bias or reversing the conditional prediction formulation.

$k$NN is a classical machine learning algorithm~\citep{fix1989discriminatory} well known for its simplicity and inspired a wide range of application~\citep{papernot2018deep,NEURIPS2018_6e091746}.
In the field of NLP, \citet{kaiser2017learning} construct a differentiable memory module for nearest neighbor searching which improves generalization to rare events.
Similar idea has also been explored for generation tasks~\citep{guu-etal-2018-generating}, such as dialog generation~\citep{weston-etal-2018-retrieve}, machine translation~\citep{khandelwal2021nearest}, etc.
\citet{wang-etal-2022-training} and~\citet{chen2022decoupling} propose to retrieve similar training examples and incorporate them into the input to jointly train the model.
While ~\citet{Khandelwal2020Generalization} and ~\citet{shi2022nearest} consider unsupervised corpus as datastore, retrieve and interpolate them with the current step language modeling probability.
The retrieved corpus can also serve as references for knowledge intensive tasks, but the retriever would need explicitly training for such purpose~\citep{NEURIPS2020_6b493230,pmlr-v162-borgeaud22a,izacard2022few}.
Different from these works, $k$NN Prompting is suitably situated in the gradient-free paradigm of LLM deployment, which avoids calibration treatment and effectively bridges data scaling into model scaling.

\section{Discussion}\label{discussion}
Under the existing ICL paradigm, it is often impossible to take advantage of both the capability of LLM and the data utility of finetuning, i.e., model scaling and data scaling.
$k$NN Prompting finds an effective solution to promise them both.
Nevertheless, we assume its data utility should still be inferior to the specialized finetuning of LLMs, if given sufficient computation resources in an ideal setting.
We believe that it is a very important and promising direction to further approach this upper-bound and expect to raise more interests in future works.

A potential concern for retrieval-based models is their efficiency, especially when corpus level datastore is utilized.
$k$NN Prompting is free of such concerns as it considers training data, which is in manageable scale.
Under few shot scenario, $k$NN Prompting even reduces computational costs at deployment time compared to standard ICL.
It works well with one shot demonstrations as experimented in Section~\ref{sec:datautilityfewshot}, while the anchor examples are queried for only once and cached locally.
By contrast, existing methods only perform better when all examples are prepended in a single prompt.
This advantage is rather important as we need to repeatedly query the prompt in practical usage of LLM service, and longer prompt results in linearly more monetary costs if charged by token numbers or super-linearly more computational costs if measured by FLOPS.

\section{Conclusion}
In this paper, we propose $k$NN Prompting as a simple and effective solution to advance gradient-free deployment of LLM inference.
Motivated as calibration-free optimization, $k$NN Prompting significantly outperforms state-of-the-art calibration-based methods under comparable few shot scenario.
While its major advantage is further revealed when training data increases and can not fit in the context.
$k$NN Prompting can effectively scale up with as many training data as are available, successfully bridging the utility of data scaling into model scaling.
The proposed framework endeavors to realize more effective, efficient and applicable utilization of large language models in realistic scenarios, and hopefully could inspire further research interests toward the same goal.

\subsubsection*{Ethical Considerations}
This work is built upon the ICL paradigm and involves querying LLM for responses.
These models might generate contents with potential ethical risks regarding fairness and bias~\citep{DBLP:journals/corr/abs-2108-07258,blodgett-etal-2020-language}, depending on specific downstream tasks.
Although the scope of this paper remains on how to better exploit LLM for task performance, it is worth further discussion to combine the proposed framework in conjunction with well-established methods that can measure~\citep{nadeem-etal-2021-stereoset} and mitigate~\citep{nadeem-etal-2021-stereoset,gupta-etal-2022-mitigating} such ethical risks.

\subsubsection*{Acknowledgments}
This work is supported in part by the National Natural Science Foundation of China under Grant 62222212, 62232006 and 61876223, Science Fund for Creative Research Groups under Grant 62121002.

\bibliography{iclr2023_conference}
\bibliographystyle{iclr2023_conference}

\appendix
\pagebreak
\section{Dataset Statistics: Max Shot In Context}\label{appendix:datasetstatistics}
We provide detailed statistics about the number of maximum shots in Table~\ref{table:moremaxshotstatistics}, i.e., $M_T$ for each task, corresponding to Table~\ref{table:maxshotstatistics} in the main manuscript.

\begin{table*}[!htbp]
\centering
\resizebox{\columnwidth}{!}{
\begin{tabular}{{cc|cccccccccc}}
\toprule
&&\textbf{SST2}&\textbf{SUBJ}&\textbf{MPQA}&\textbf{AGNews}&\textbf{CB}&\textbf{CR}&\textbf{DBPedia}&\textbf{MR}&\textbf{RTE}&\textbf{TREC}\\
\midrule
\multicolumn{2}{c|}{Num. of Classes}&2&2&2&4&3&2&14&2&2&6\\
\midrule
\multicolumn{2}{c|}{Average Length of Templates}
&19.1&34.9&10.4&59.5&90.8&29.0&71.6&32.7&79.8&17.6\\
\midrule
\multirow{2}*{1024 Toks}&Max Shots (TP)&20 (2\%)&12 (1\%)&39 (0\%)&3 (0\%)&2 (0\%)&14 (4\%)&1 (77\%)&14 (4\%)&4 (0\%)&8 (1\%)\\
&$M_T$&\cellcolor{Gray}16&\cellcolor{Gray}8&\cellcolor{Gray}32&\cellcolor{Gray}2&\cellcolor{Gray}2&\cellcolor{Gray}8&\cellcolor{Gray}1&\cellcolor{Gray}8&\cellcolor{Gray}4&\cellcolor{Gray}8\\
\midrule
\multirow{2}*{2048 Toks}&Max Shots (TP)&44 (5\%)&25 (1\%)&81 (3\%)&7 (1\%)&6 (2\%)&28 (4\%)&1 (0\%)&27 (2\%)&10 (4\%)&17 (5\%)\\
&$M_T$&\cellcolor{Gray}32&\cellcolor{Gray}16&\cellcolor{Gray}32&\cellcolor{Gray}4&\cellcolor{Gray}4&\cellcolor{Gray}16&\cellcolor{Gray}1&\cellcolor{Gray}16&\cellcolor{Gray}8&\cellcolor{Gray}16\\
\bottomrule
\end{tabular}}
\caption{Dataset statistics and the maximum shots (per class) that a context of 1024 tokens or 2048 tokens can allow. We provide maximum shots under 5\% Truncation Probability (TP) restriction as well as the actual $M_T$ taken in this paper, which is set from \{1, 2, 4, 8, 16, 32\} for simplicity.}
\label{table:moremaxshotstatistics}
\end{table*}

\section{More Analyses}

\subsection{Distance Measurement}\label{appendix:distancemeasurement}
We investigate euclidean distance as an alternative distance measurement, which has also been explored in~\citet{Khandelwal2020Generalization}.
We take the contextual representation $\bm h$ of LLM, and denotes their distance as $D_{L2}(\bm h_{\mathrm{test}}, \bm h_i)$.
Table~\ref{table:distancemeasurement} shows that both methods are effective but $D_{KL}$ (Equation~\ref{kldistance}) based on LM distribution $\bm p$ is a superior measurement.
Actually, $\bm p$ is a projection of $\bm h$ through the word embedding, we think this procedure exploits the well-structured word embeddings of LLM to provide more disentangled representations, thus can better serves as the anchor space.

\begin{table*}[h]
\centering
\resizebox{\columnwidth}{!}{
\begin{tabular}{{l|cccccccccc|c}}
\toprule
\textbf{Measurements}&\textbf{SST2}&\textbf{SUBJ}&\textbf{MPQA}&\textbf{AGNews}&\textbf{CB}&\textbf{CR}&\textbf{DBPedia}&\textbf{MR}&\textbf{RTE}&\textbf{TREC}&\textbf{AVG}\\
\midrule
\small\textbf{In-Context Learning}&81.3$_{\pm\textrm{5.4}}$&64.1$_{\pm\textrm{11.3}}$&75.2$_{\pm\textrm{8.8}}$&72.7$_{\pm\textrm{18.5}}$&\textbf{60.7}$_{\pm\textrm{2.8}}$&66.2$_{\pm\textrm{16.7}}$&83.5$_{\pm\textrm{3.8}}$&72.2$_{\pm\textrm{13.9}}$&53.0$_{\pm\textrm{1.7}}$&54.2$_{\pm\textrm{4.9}}$&68.31\\
\small\textbf{Contextual Repr + $D_{L2}$}&84.1$_{\pm\textrm{8.5}}$&73.0$_{\pm\textrm{8.4}}$&75.2$_{\pm\textrm{8.8}}$&82.0$_{\pm\textrm{0.8}}$&60.4$_{\pm\textrm{8.1}}$&78.0$_{\pm\textrm{11.7}}$&\textbf{95.1}$_{\pm\textrm{1.0}}$&82.6$_{\pm\textrm{5.0}}$&53.3$_{\pm\textrm{2.4}}$&\textbf{70.3}$_{\pm\textrm{4.4}}$&75.40 (\textcolor{red}{+7.09})\\
\small\textbf{LM Distribution + $D_{KL}$}&\textbf{87.7}$_{\pm\textrm{3.5}}$&\textbf{77.0}$_{\pm\textrm{3.5}}$&75.2$_{\pm\textrm{8.8}}$&\textbf{86.2}$_{\pm\textrm{1.8}}$&58.9$_{\pm\textrm{2.2}}$&\textbf{88.2}$_{\pm\textrm{3.5}}$&94.1$_{\pm\textrm{2.3}}$&\textbf{83.9}$_{\pm\textrm{2.4}}$&\textbf{53.6}$_{\pm\textrm{3.0}}$&64.8$_{\pm\textrm{4.2}}$&\textbf{76.97} (\textcolor{red}{+8.66})\\
\bottomrule
\end{tabular}}
\caption{Comparison for alternative distance measurement.}
\label{table:distancemeasurement}
\end{table*}

\subsection{Reliance on Prior Knowledge}\label{appendix:relianceonpriorknow}
\paragraph{Conclusion} We further explore the robustness of $k$NN Prompting regarding the reliance of prior knowledge on target distribution.
We show that among the investigated baselines, ICL and ContextualCalibration~\citep{pmlr-v139-zhao21c} are greatly impacted by prior knowledge of test distribution, while $k$NN Prompting and NoisyChannel~\citep{min-etal-2022-noisy} are much more robust.

\paragraph{Experimental Setting} We investigate various combinations of prior distribution by controlling the imbalance ratio $\lambda_{train}$ and $\lambda_{test}$. For every setting, we include 5 binary classification tasks (SST2, MPQA, SUBJ, MR, CR) and run with 10 random seeds, we report the average score of these results.
Specifically, $\lambda_{train}=0.125$ means one category (positive) accounts for 12.5\% of the entire train set, and $\lambda_{train}=1.5$ corresponds to the balanced setting.
We investigate both $\lambda_{train}<0.5$ and $\lambda_{test}<0.5$, which results in three different settings.

ContextualCalibration explicitly include a prior distribution, at such imbalanced scenario, one can either use the default assumption ($pos:neg=1:1$, as designed in the original paper) or use the observed assumption from train set ($pos:neg=\lambda_{train}/(1-\lambda_{train})$). We respectively refer to them as \textbf{Balanced Prior} and \textbf{Trainset Prior}.
Note that test distribution is unaccessible so we can not use it.
Other methods (ICL, NoisyChannel and $k$NN Prompting) do not \textit{technically} incorporate any prior knowledge. So they are not concerned with this investigation dimension.

\paragraph{Analyses} The results are reported in Table~\ref{table:relianceonpriorknow}.
For ICL, LLM naturally suffers from the bias learned in pretraining stage, thus is vulnerable to any different prior distributions. The performance greatly degrades in $Setting~B$~(-22.06) and $Setting~C$~(-22.17).

For ContextualCalibration, technically, it necessarily requires a prior distribution to rectify the LLM predicted label word probabilities. If the prior knowledge does not match the (unaccessible) test distribution, its performance will be greatly degraded.
Specifically, if Balanced Prior is consistent with test distribution, the method performs well ($Setting~A$, -1.77), otherwise, the performance degrades ($Setting~B$, -20.24 and $Setting~C$, -12.65).
Similarly, if Trainset Prior Assumption is consistent with test distribution, the method performs well ($Setting~C$, +8.65), otherwise, the performance degrades ($Setting~B$, -20.24 and $Setting~A$, -18.92).

For NoisyChannel, its performance is rather robust (-2.60/-0.63/-7.87 respectively in $Setting~A, B, C$). By re-formulating ICL into computing conditional probability of the input given the output, it is indeed an effective way to calibrate the task prediction.

For the proposed $k$NN Prompting, technically, it does \textbf{not} incorporate any prior knowledge of train or test distribution. Both the construction of datastore and the retrieving then predicting procedure do not vary w.r.t. different prior knowledge of distribution.
The proposed method can robustly adapt to all imbalanced settings, including imbalanced trainset, testset and both.
There is basically no performance degradation (-1.3/-2.6/+0.05 respectively in $Setting~A, B, C$, where -1.3/-2.6 can be considered within ordinary fluctuation).

\definecolor{flamingopink}{rgb}{0.99, 0.56, 0.67}
\definecolor{lightpink}{rgb}{1.0, 0.71, 0.76}

\begin{table*}[h!]
\centering
\resizebox{0.95\columnwidth}{!}{
\begin{tabular}{{l|cccc|c|r}}
\toprule
\textbf{Methods}&\textbf{0.125}&\textbf{0.25}&\textbf{0.375}&\textbf{0.5 (Balanced)}&\textbf{AVG}&\textbf{MaxDrop}\\
\midrule
\belowrulesepcolor{Gray}
\rowcolor{Gray}\multicolumn{7}{c}{\large{$Setting~A.\quad\lambda_{train}<0.5, \lambda_{test}=0.5$}}\\
\aboverulesepcolor{Gray}
\midrule
\textbf{ICL}&62.4&64.2&65.7&68.2&65.15&-5.85\\
\textbf{ContextualCalibration w/ Balanced Prior}&74.0&69.7&68.5&70.3&70.64&-1.77\\
\textbf{ContextualCalibration w/ Trainset Prior}&51.4&71.5&80.2&70.3&68.36&-18.92\\
\textbf{NoisyChannel}&70.0&70.9&71.9&72.6&71.33&-2.60\\
\textbf{$k$NN Prompting}&\textbf{79.0}&\textbf{80.1}&\textbf{80.5}&\textbf{80.3}&\textbf{79.98}&\cellcolor{flamingopink}\textbf{-1.29}\\
\midrule
\belowrulesepcolor{Gray}
\rowcolor{Gray}\multicolumn{7}{c}{\large{$Setting~B.\quad\lambda_{train}=0.5, \lambda_{test}<0.5$}}\\
\aboverulesepcolor{Gray}
\midrule
\textbf{ICL}&45.9&53.4&60.9&67.9&57.04&-22.06\\
\textbf{ContextualCalibration w/ Balanced Prior}&50.0&57.2&64.2&70.2&60.42&-20.24\\
\textbf{ContextualCalibration w/ Trainset Prior}&50.0&57.2&64.2&70.2&60.42&-20.24\\
\textbf{NoisyChannel}&72.1&72.2&71.8&72.4&72.14&\cellcolor{flamingopink}\textbf{-0.63}\\
\textbf{$k$NN Prompting}&\textbf{77.1}&\textbf{78.0}&\textbf{78.7}&\textbf{79.7}&\textbf{78.38}&-2.62\\
\midrule
\belowrulesepcolor{Gray}
\rowcolor{Gray}\multicolumn{7}{c}{\large{$Setting~C.\quad\lambda_{train}=\lambda_{test}<0.5$}}\\
\aboverulesepcolor{Gray}
\midrule
\textbf{ICL}&45.8&50.0&58.6&67.9&55.57&-22.17\\
\textbf{ContextualCalibration w/ Balanced Prior}&63.0&57.6&61.7&70.2&63.13&-12.65\\
\textbf{ContextualCalibration w/ Trainset Prior}&\textbf{87.8}&\textbf{84.5}&78.8&70.2&\textbf{80.33}&\cellcolor{flamingopink}\textbf{+8.56}\\
\textbf{NoisyChannel}&64.6&68.6&70.8&72.4&69.10&-7.87\\
\textbf{$k$NN Prompting}&79.8&80.7&\textbf{80.3}&\textbf{79.7}&80.12&\cellcolor{flamingopink}+0.05\\
\bottomrule
\end{tabular}}
\caption{Reliance on prior knowledge. All reported results are averaged across 5 datasets, and we further report average performance across all imbalance ratios. $\lambda_{train/test}$ denotes the subsampled ratio. MaxDrop measures the performance degradation compared to ordinary balanced setting ($\lambda_{*}=0.5$), where the best is bolded and no drop is highlighted in \textbf{\textcolor{flamingopink}{pink}}.}
\label{table:relianceonpriorknow}
\end{table*}

\subsection{Empirical Choice of $k$}\label{appendix:empiricalchoiceofk}
\begin{figure}[h]
\begin{subfigure}{.33\textwidth}
  \centering
  \includegraphics[width=\linewidth]{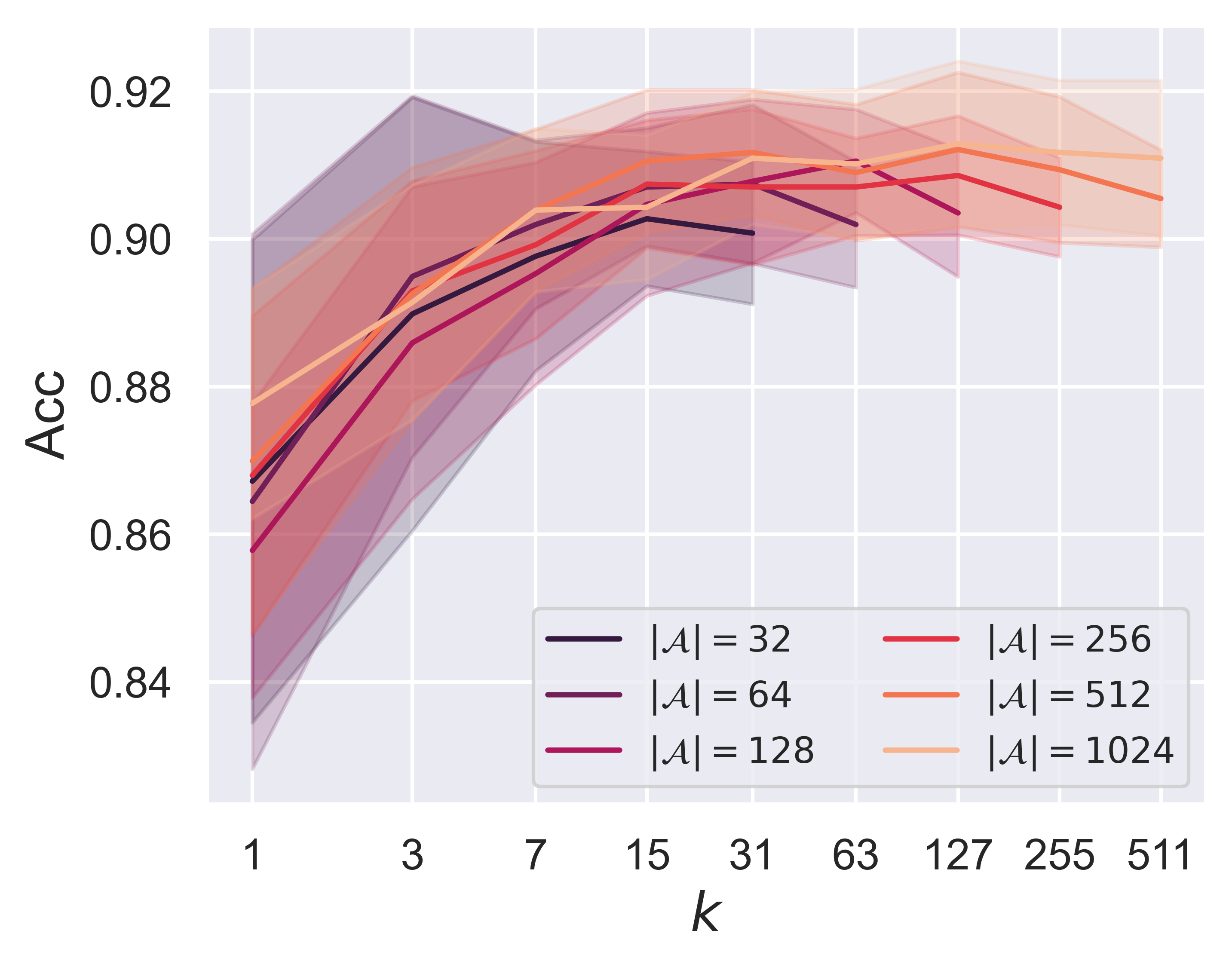}
  \label{fig:demoanchorsplitsst2}
\end{subfigure}
\begin{subfigure}{.33\textwidth}
  \centering
  \includegraphics[width=\linewidth]{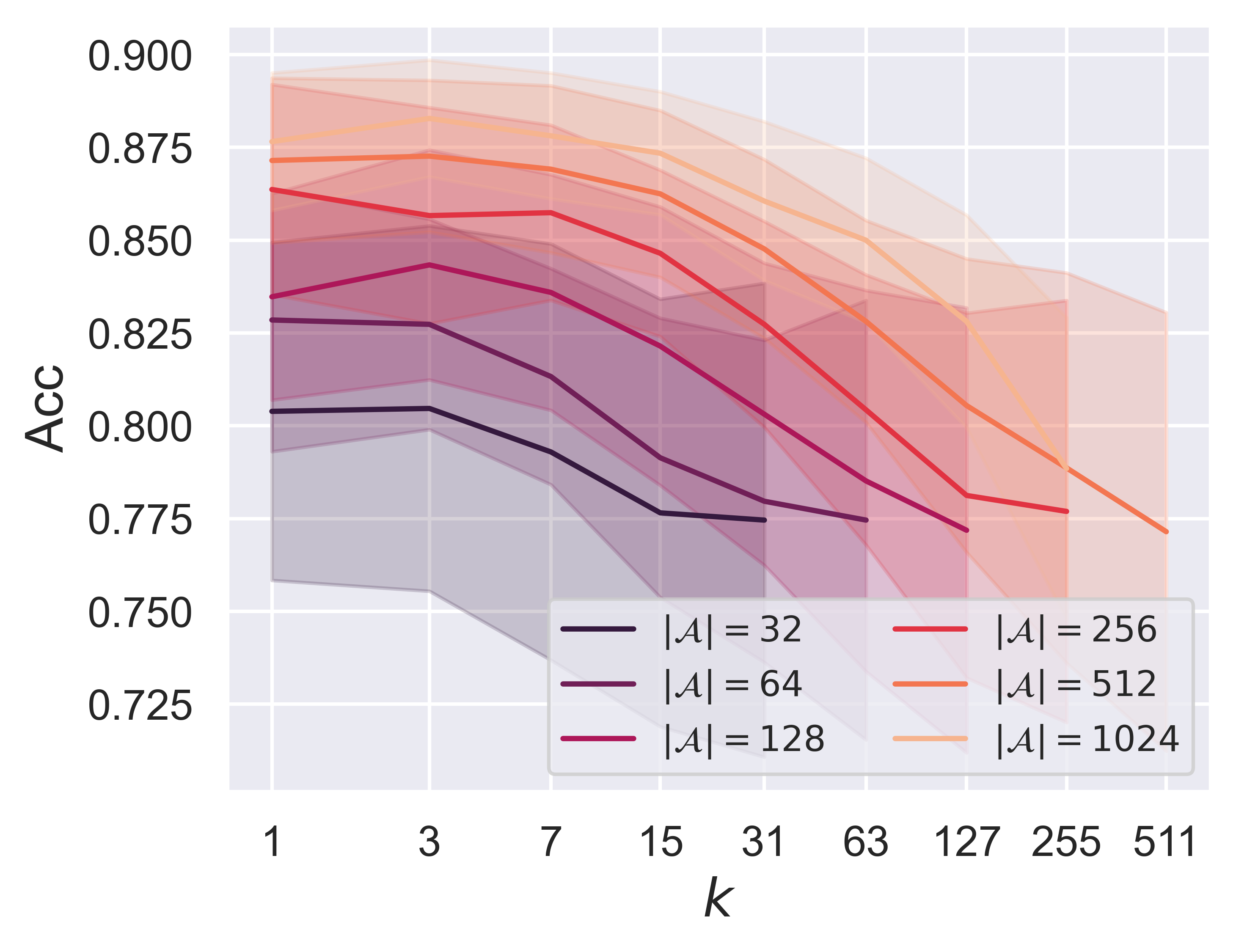}
  \label{fig:demoanchorsplitsubj}
\end{subfigure}
\begin{subfigure}{.33\textwidth}
  \centering
  \includegraphics[width=\linewidth]{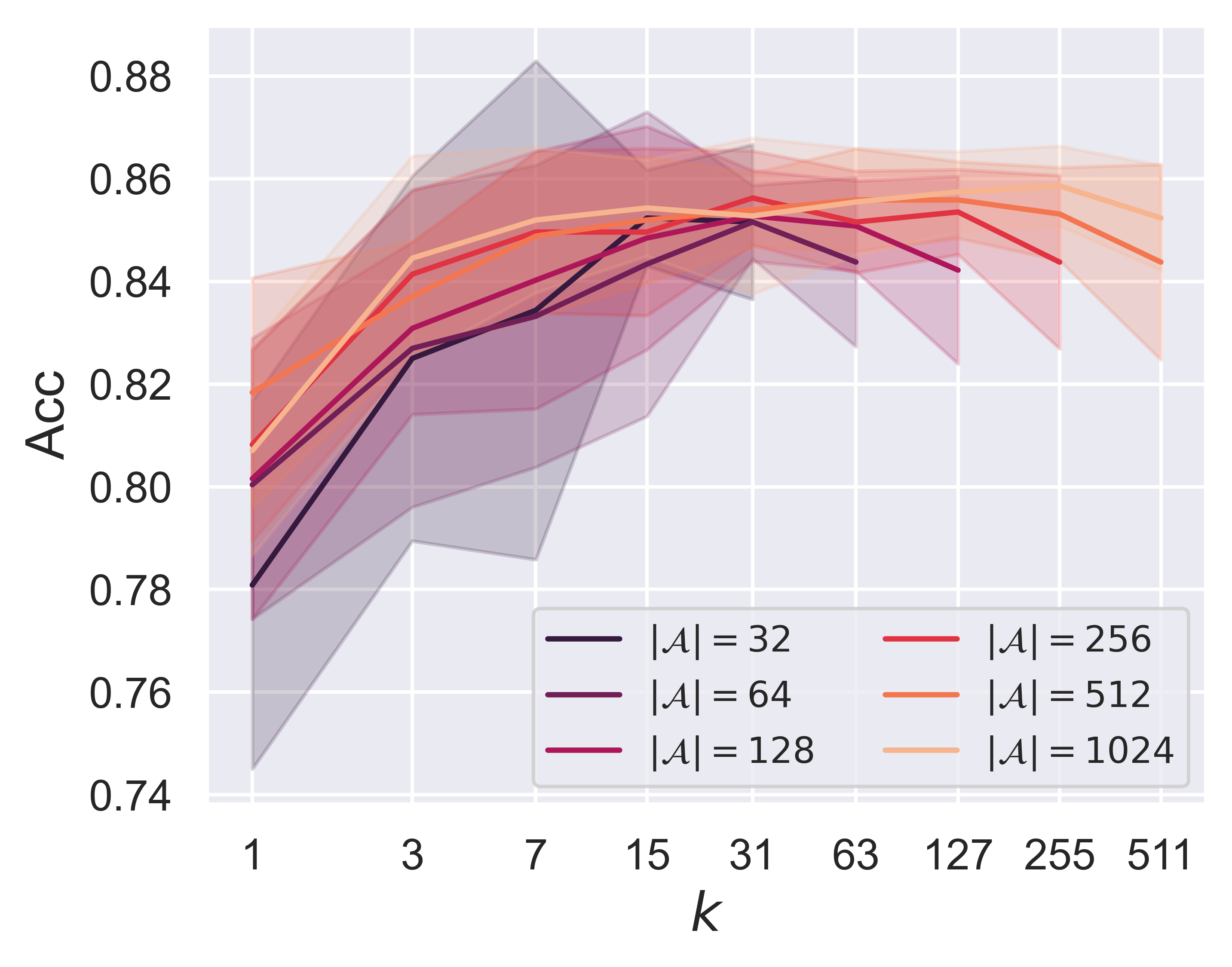}
  \label{fig:demoanchorsplitmpqa}
\end{subfigure}
\caption{Empirical choices of $k$. Conducted on SST2, SUBJ and MPQA respectively (left to right).}
\label{fig:empiricalchoiceofk}
\end{figure}

In Figure~\ref{fig:empiricalchoiceofk} we search for different choices of $k$ on MPQA, and found that it is generally a rather robust choice within the wide range $[3, |\mathcal{A}|/2 - 1]$.

\subsection{Robustness Under Imbalanced Scenario}\label{appendix:robustnessunderimbalance}
\begin{figure*}
    \begin{subfigure}[b]{0.42\columnwidth}
        \includegraphics[width=\textwidth]{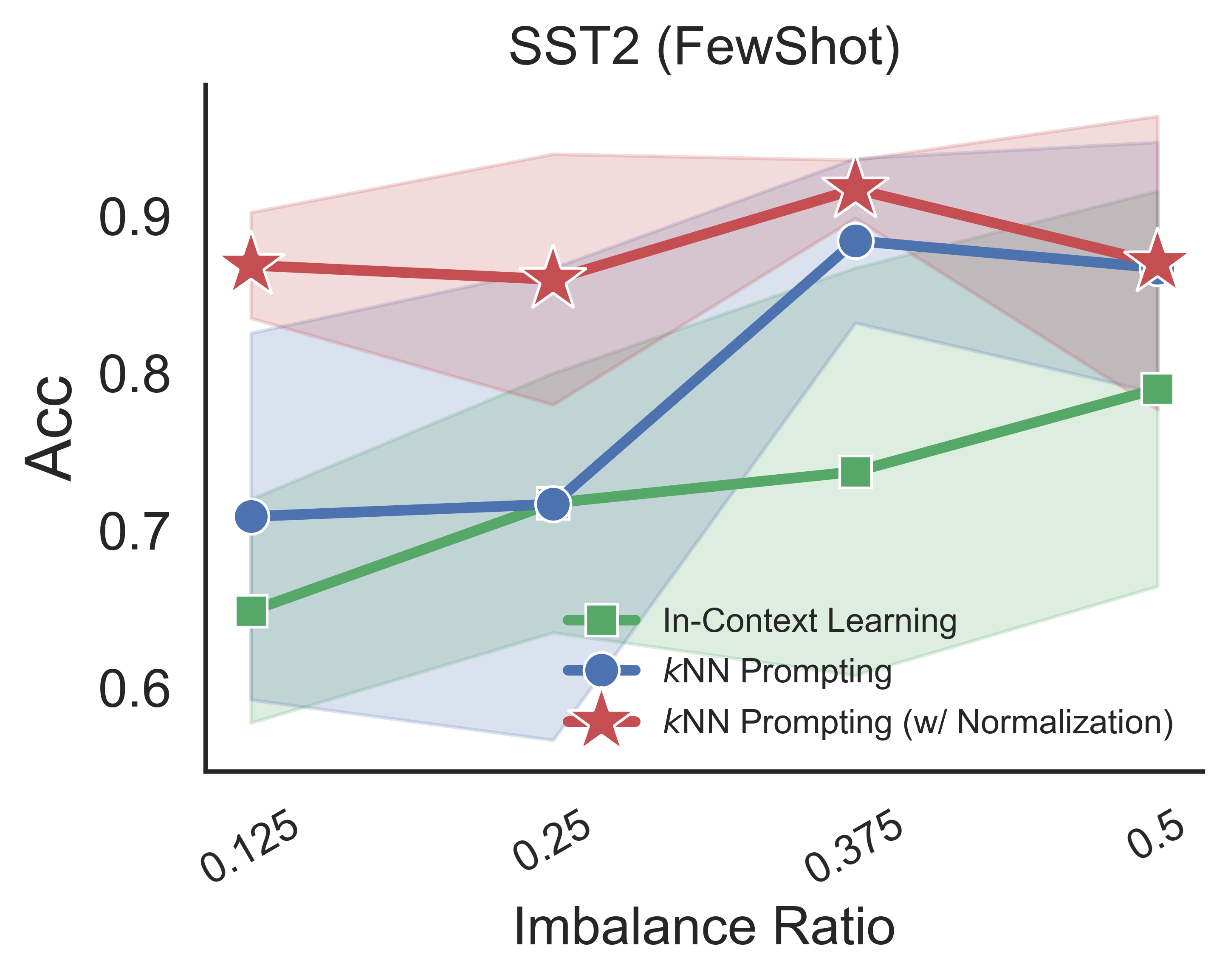}
        \label{fig:imbalancempqafewshot}
    \end{subfigure}
    \hfill
    \begin{subfigure}[b]{0.42\columnwidth}
        \includegraphics[width=\textwidth]{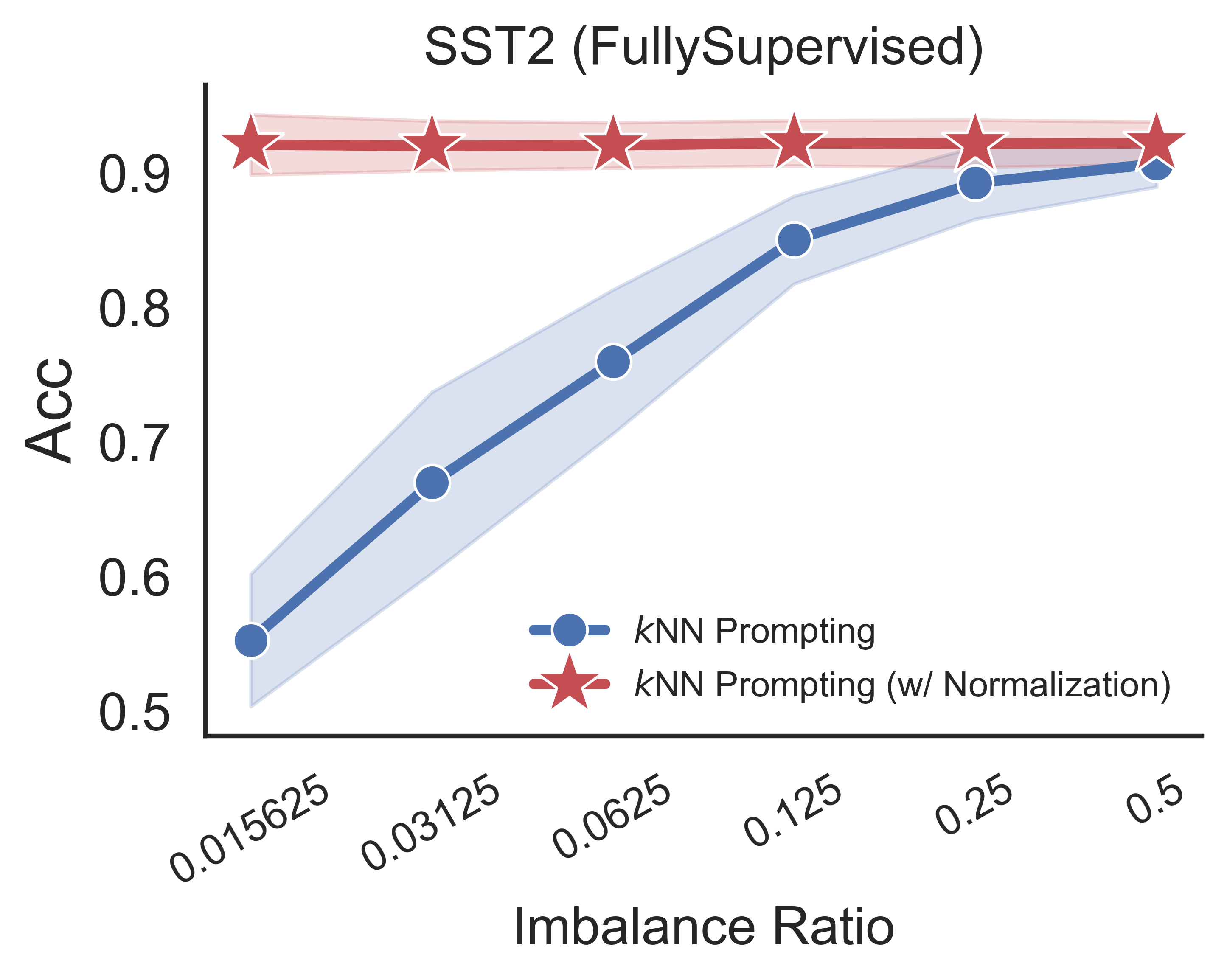}
        \label{fig:imbalancempqafullysupervised}
    \end{subfigure}

    \begin{subfigure}[b]{0.42\columnwidth}
        \includegraphics[width=\textwidth]{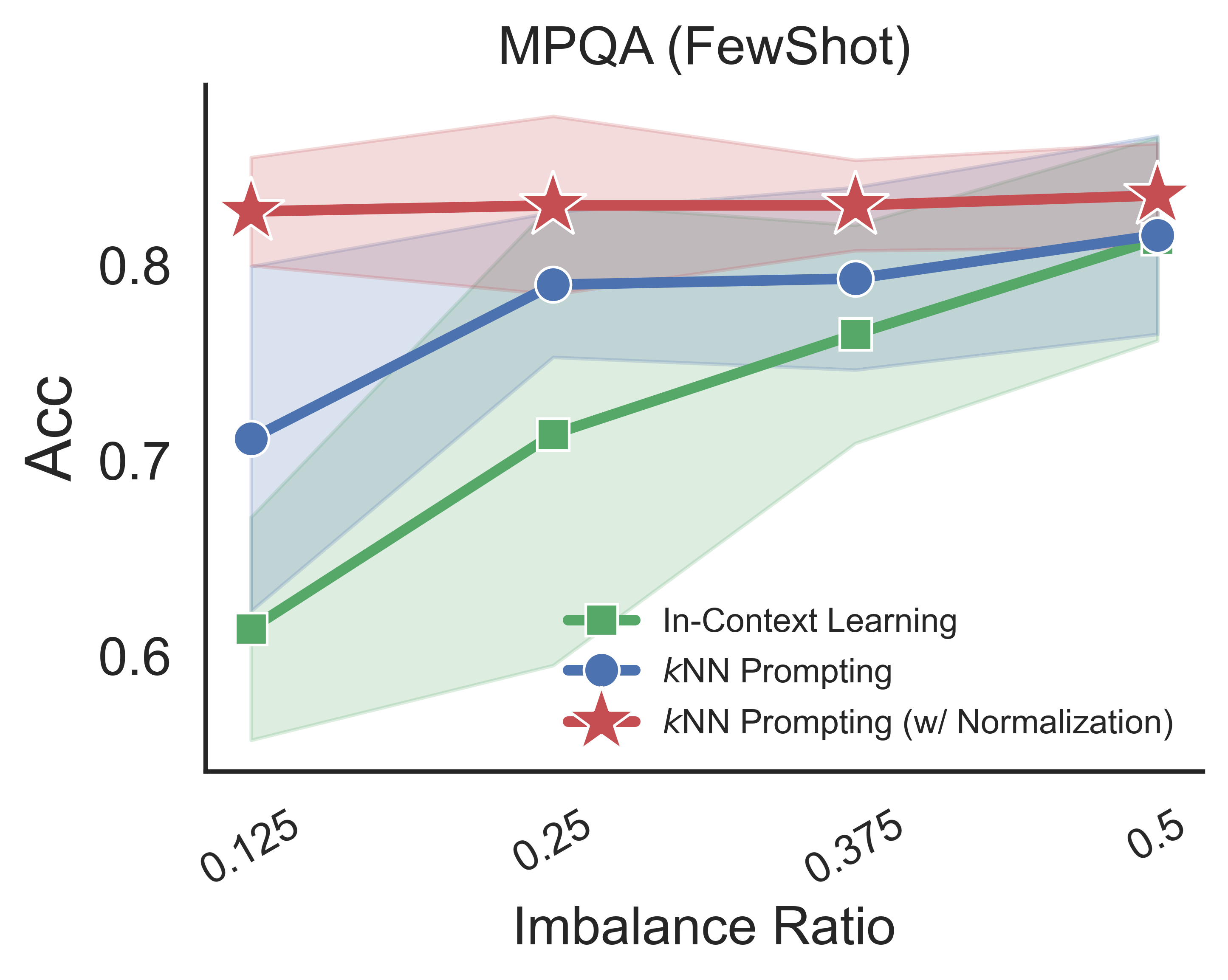}
        \label{fig:imbalancemrfewshot}
    \end{subfigure}
    \hfill
    \begin{subfigure}[b]{0.42\columnwidth}
        \includegraphics[width=\textwidth]{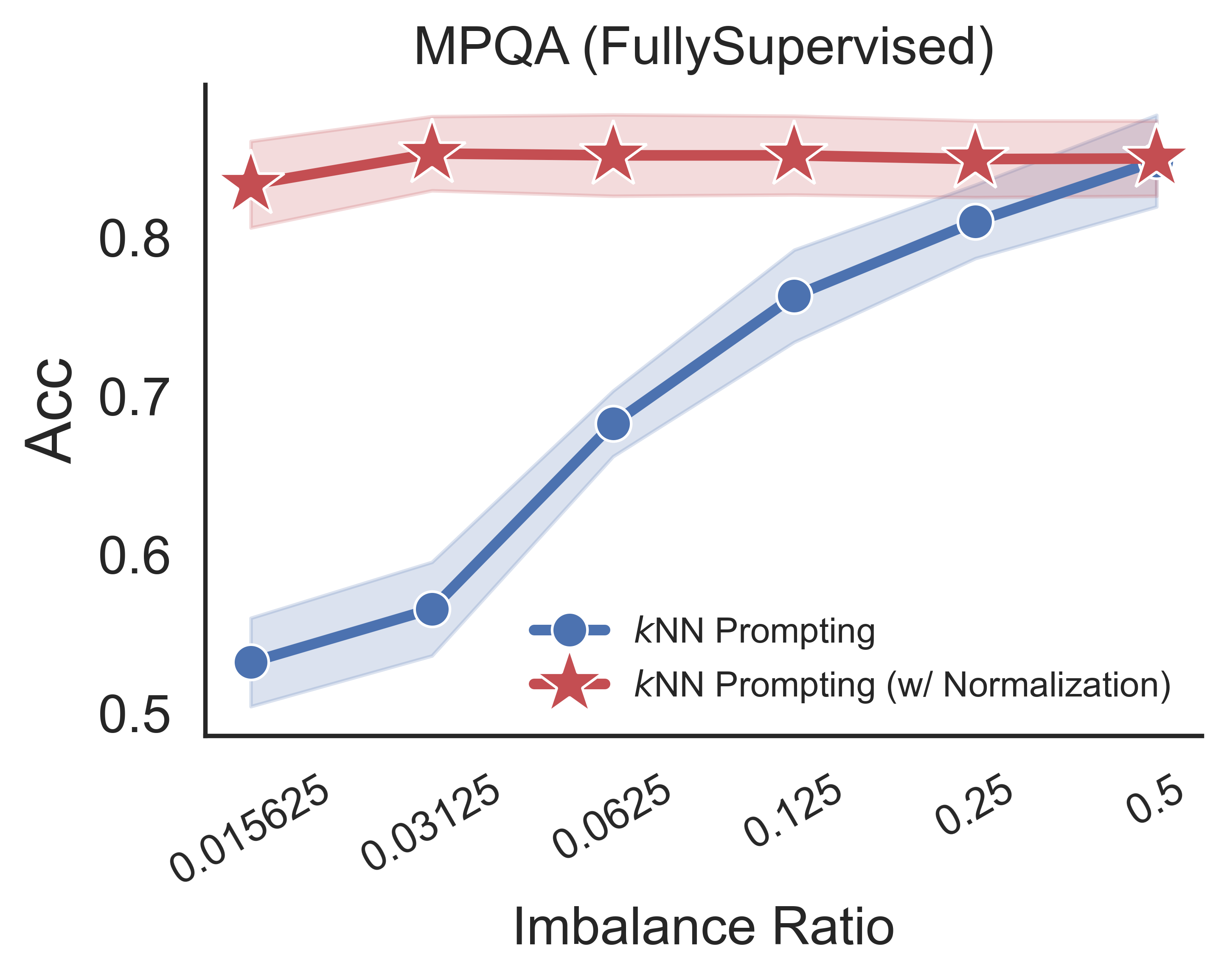}
        \label{fig:imbalancemrfullysupervised}
    \end{subfigure}
    \caption{Robustness under imbalanced scenario. Left: few-shot scenario (32). Right: fully-supervised scenario (1024).}
    \label{fig:moreimbalancedsetting}
\end{figure*}

We further test the robustness of $k$NN Prompting under imbalanced label scenario.
Take binary classification like SST2 as example, we simulate imbalance ratio by controlling one of the category proportionally to \{0.5, 0.25, 0.125, 0.0625, 0.03125, 0.015625\} of the entire training set, where 0.5 corresponds to the ordinary balanced scenario.
We keep the test set intact, which results in a challenging out-of-distribution (OOD) setting.
Results in Figure~\ref{fig:moreimbalancedsetting} reveal the vulnerability of the proposed method.
Under imbalanced setting, $k$NN Prompting is overwhelmed by the large quantity of anchors from the majority class where it is simply far more easier to find closer neighbors.

To address such performance degradation under challenging imbalanced scenario, we propose a simple normalization trick: we average the anchor representations to produce one centered anchor for each class, the resulting anchor is thus more representative and also avoids quantity distraction.
Such an adaptation works surprisingly well with no loss of performance even under ordinary balanced setting.

\section{Complete Results}
\subsection{Data Scaling Curve Under Few-shot Scenario}\label{appendix:fewshotscalingperdataset}
\begin{figure}[h!]
\begin{subfigure}{.33\textwidth}
  \centering
  \includegraphics[width=\linewidth]{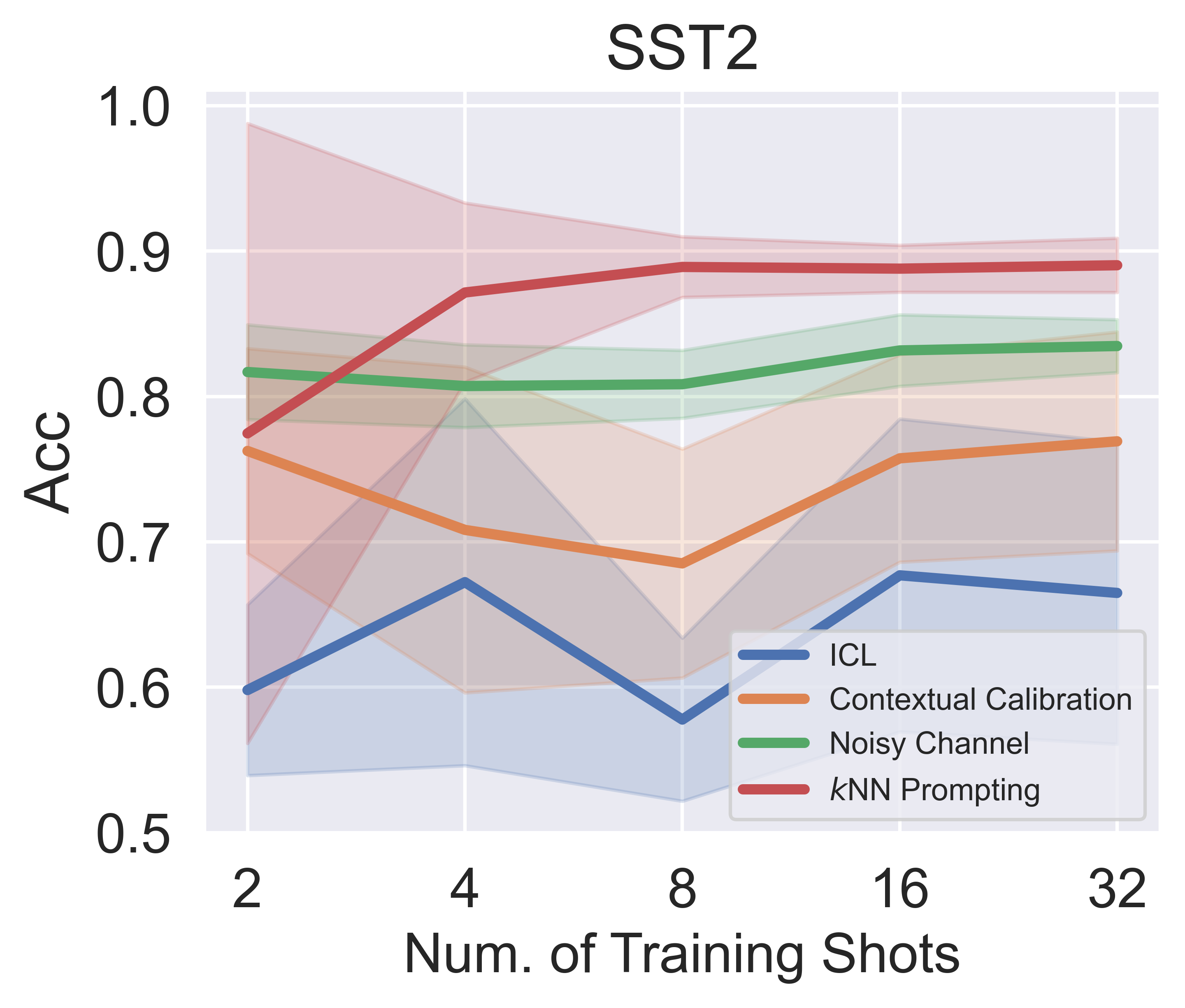}
  \label{fig:fewshotresults_sfig1}
\end{subfigure}%
\begin{subfigure}{.33\textwidth}
  \centering
  \includegraphics[width=\linewidth]{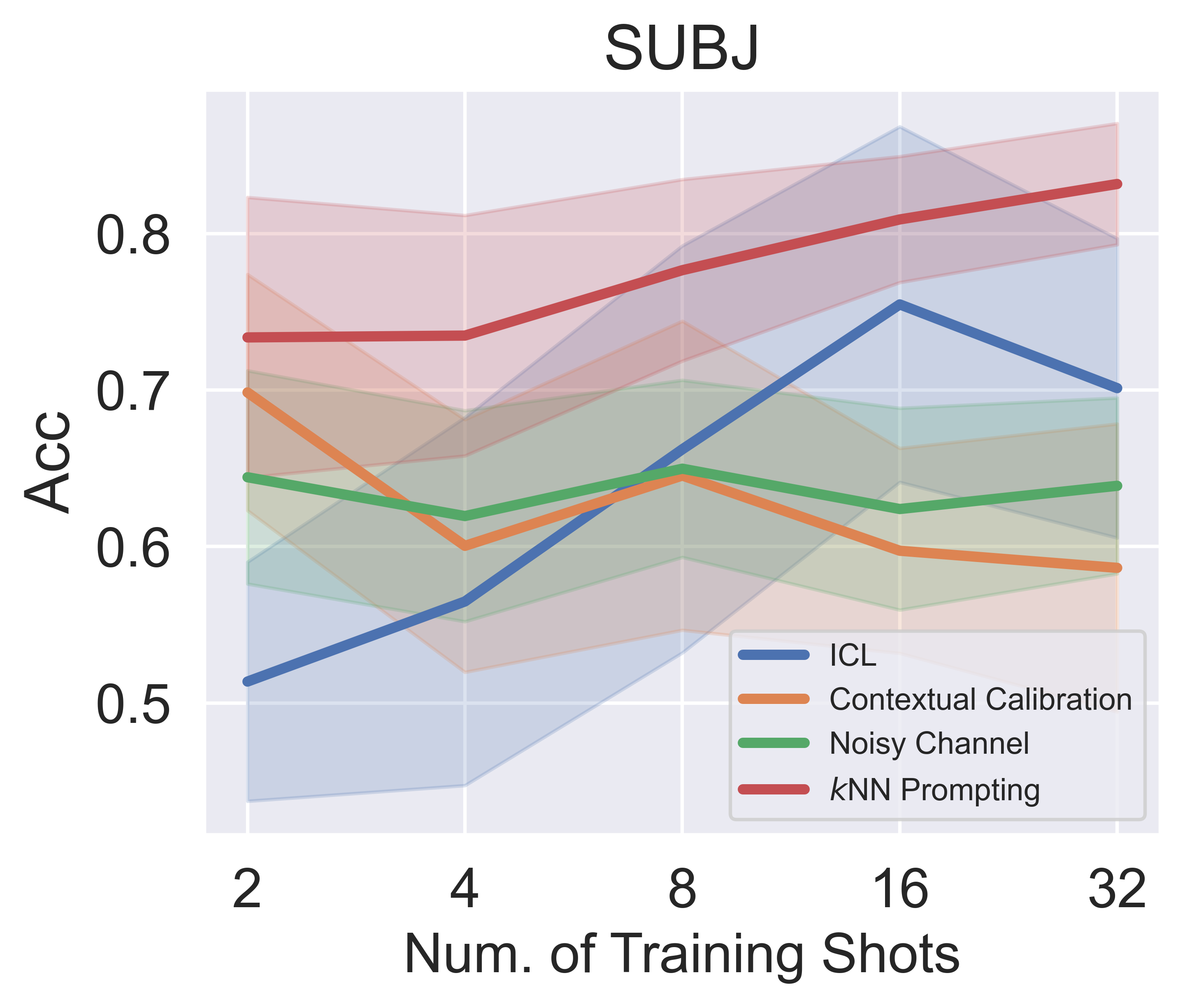}
  \label{fig:fewshotresults_sfig2}
\end{subfigure}
\begin{subfigure}{.33\textwidth}
  \centering
  \includegraphics[width=\linewidth]{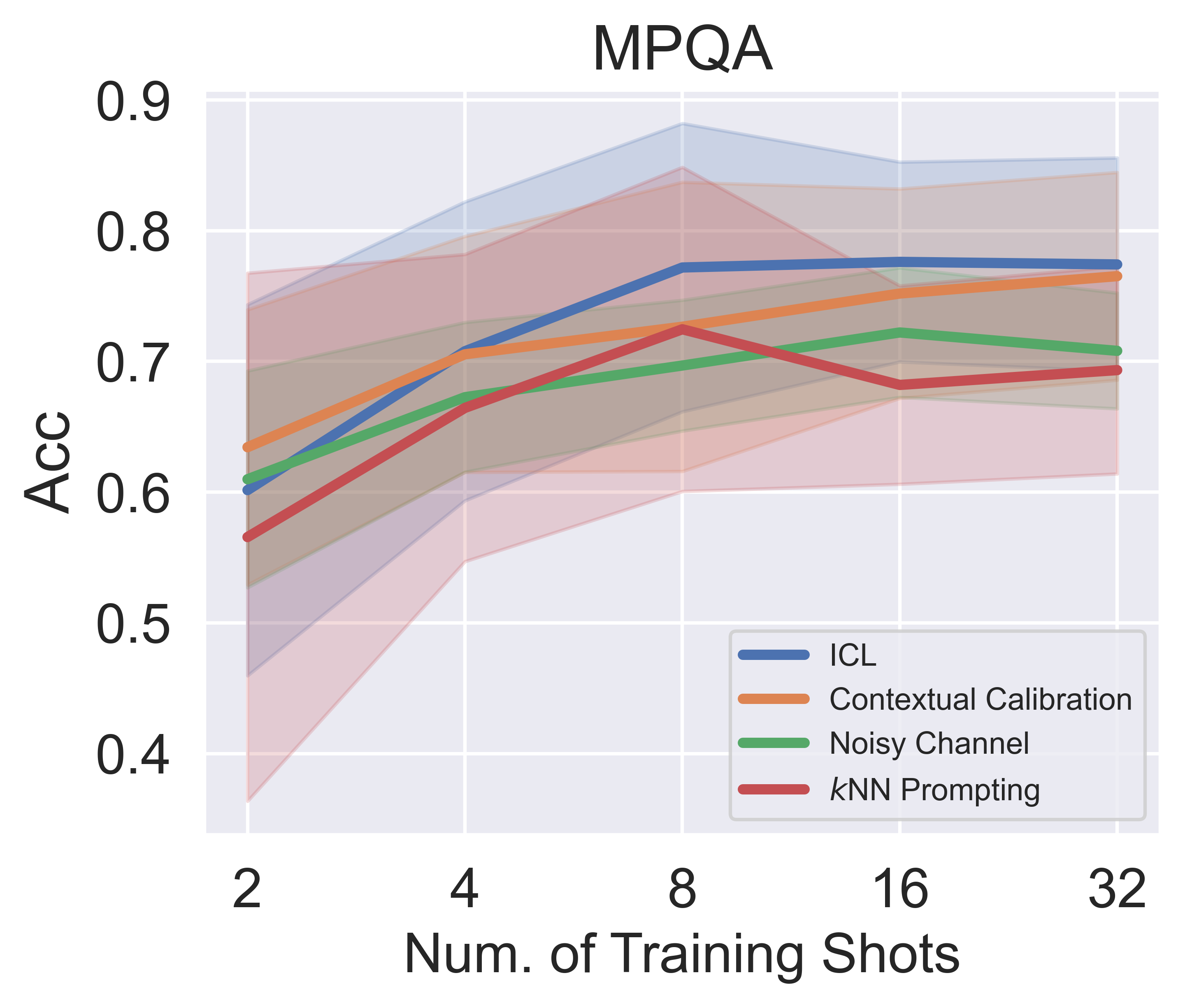}
  \label{fig:fewshotresults_sfig3}
\end{subfigure}
\begin{subfigure}{.33\textwidth}
  \centering
  \includegraphics[width=\linewidth]{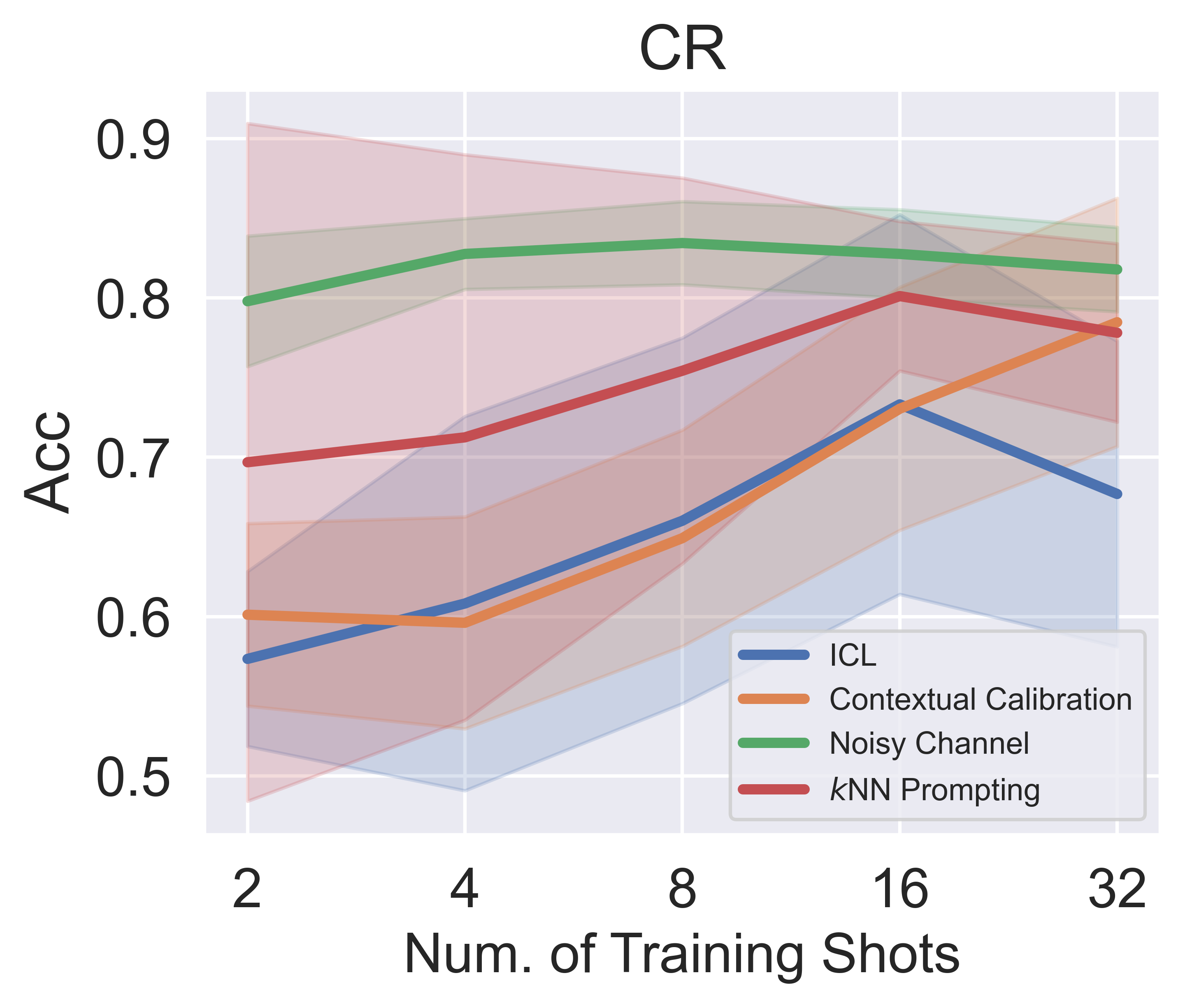}
  \label{fig:fewshotresults_sfig4}
\end{subfigure}
\begin{subfigure}{.33\textwidth}
  \centering
  \includegraphics[width=\linewidth]{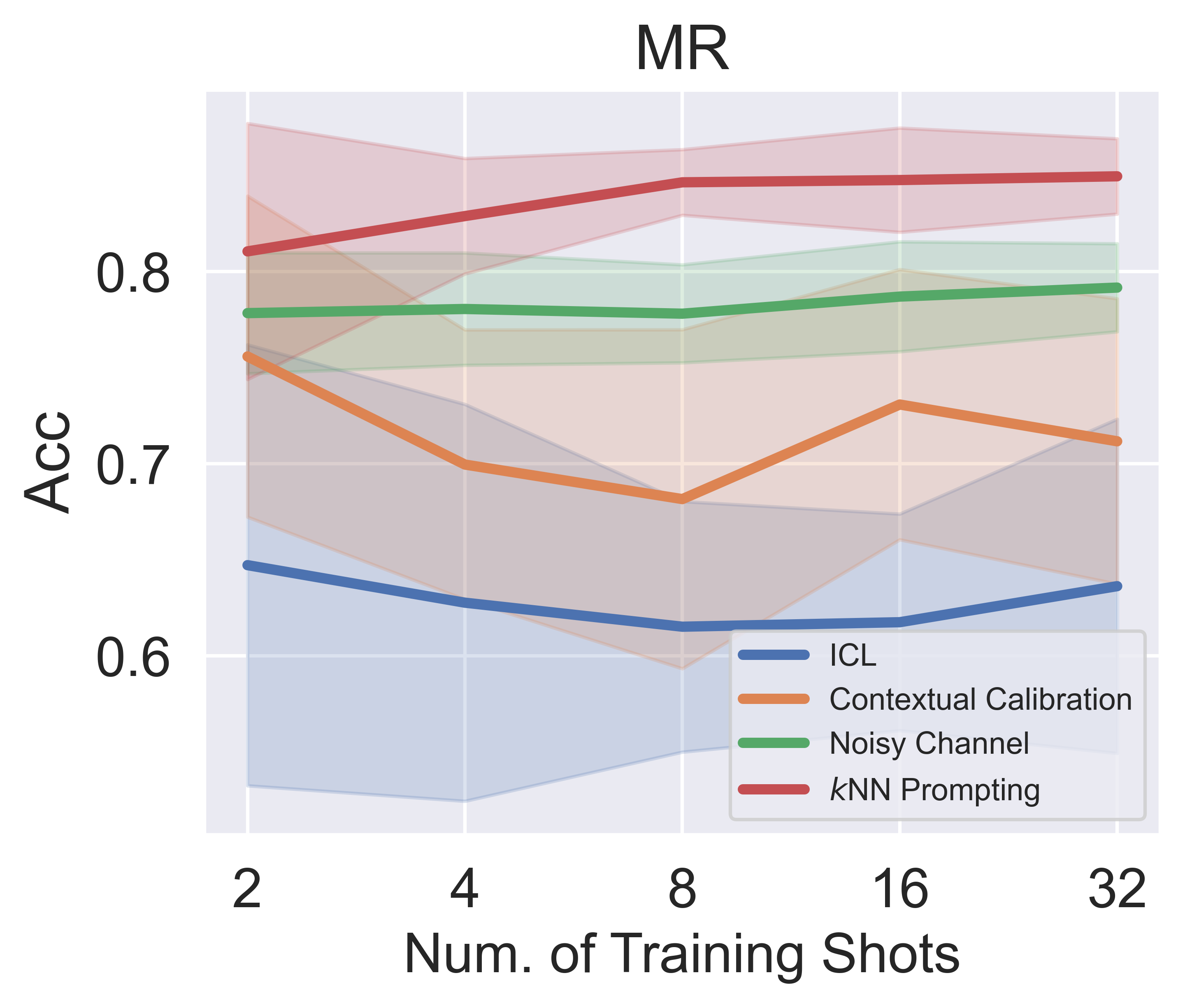}
  \label{fig:fewshotresults_sfig5}
\end{subfigure}
\begin{subfigure}{.33\textwidth}
  \centering
  \includegraphics[width=\linewidth]{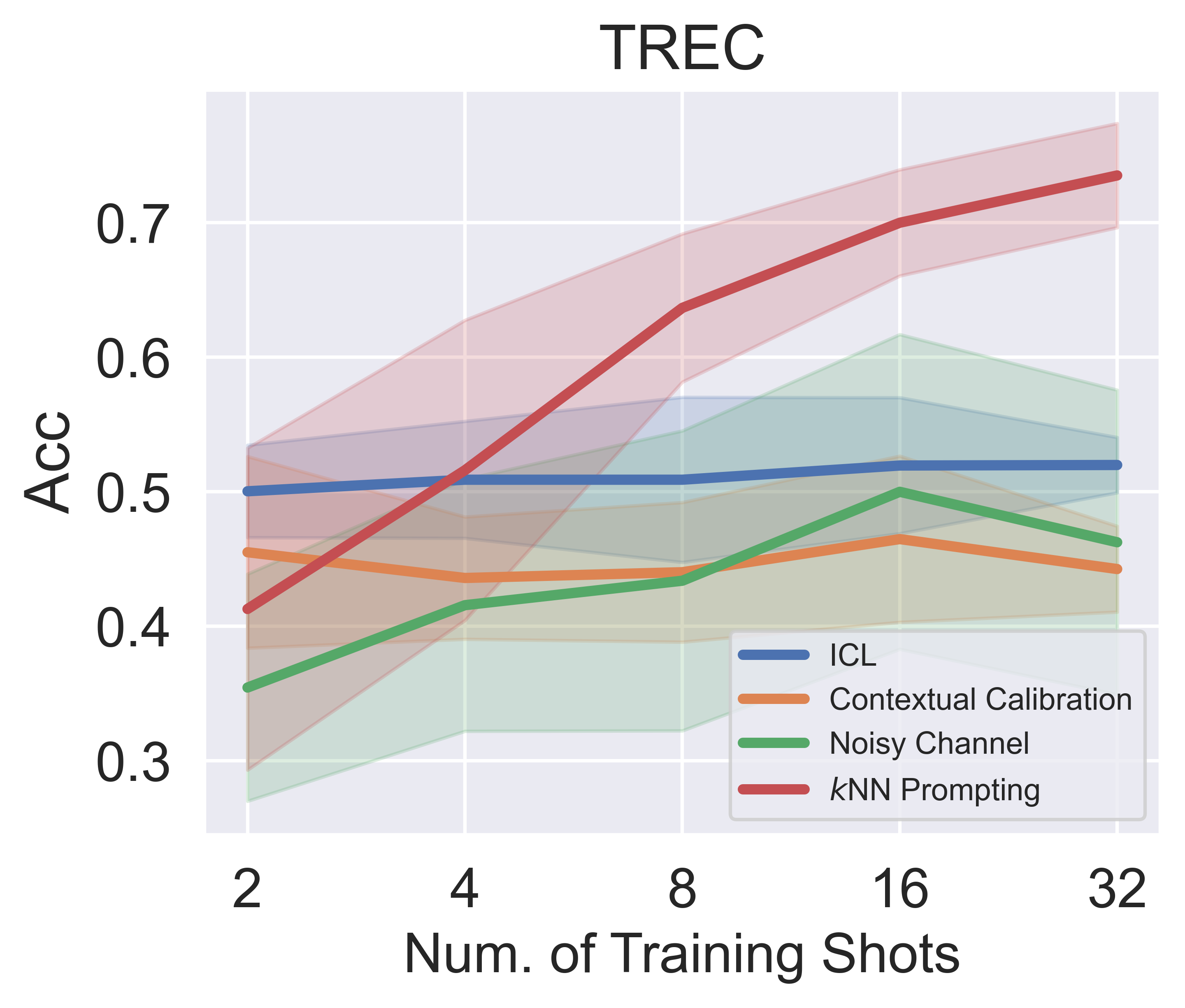}
  \label{fig:fewshotresults_sfig6}
\end{subfigure}
\caption{Scaling curve under few shot setting for each specific dataset. The baselines are strictly comparable under $m\leq 8$, some baselines might be truncated when $m\geq 16$.}
\label{fig:fewshotscalingperdataset}
\end{figure}

We provide scaling curve w.r.t. each specific dataset in Figure~\ref{fig:fewshotscalingperdataset}, corresponding to Table~\ref{table:mainresultsfewshot} and Figure~\ref{fig:fewshotscaling} in the main manuscript.

\subsection{Data Scaling Results Under Fully Supervised Scenario}\label{appendix:fullresults}

\begin{table*}[h!]
\centering
\resizebox{\columnwidth}{!}{
\begin{tabular}{{x{0.5cm}c|cccccccccc|c}}
\toprule
\multicolumn{2}{c|}{\textbf{Models \& Methods}}&\textbf{SST2}&\textbf{SUBJ}&\textbf{MPQA}&\textbf{AGNews}&\textbf{CB}&\textbf{CR}&\textbf{DBPedia}&\textbf{MR}&\textbf{RTE}&\textbf{TREC}&\textbf{AVG}\\
\midrule
\multirow{7}*{\textbf{0.8B}}&\small\textbf{In-Context Learning}&63.4$_{\pm\textrm{7.3}}$&58.9$_{\pm\textrm{8.7}}$&70.5$_{\pm\textrm{5.2}}$&61.7$_{\pm\textrm{15.4}}$&45.0$_{\pm\textrm{9.1}}$&83.3$_{\pm\textrm{13.7}}$&59.9$_{\pm\textrm{11.5}}$&77.0$_{\pm\textrm{15.7}}$&53.6$_{\pm\textrm{3.1}}$&54.4$_{\pm\textrm{1.7}}$&62.77\\
&\small\bm{$m=32$}&82.0$_{\pm\textrm{6.9}}$&82.8$_{\pm\textrm{1.2}}$&70.5$_{\pm\textrm{5.2}}$&82.2$_{\pm\textrm{0.9}}$&56.8$_{\pm\textrm{7.3}}$&87.5$_{\pm\textrm{3.0}}$&93.7$_{\pm\textrm{1.1}}$&83.0$_{\pm\textrm{2.5}}$&53.6$_{\pm\textrm{4.0}}$&71.5$_{\pm\textrm{6.2}}$&76.35\\
&\small\bm{$m=64$}&84.2$_{\pm\textrm{5.4}}$&84.6$_{\pm\textrm{0.7}}$&81.2$_{\pm\textrm{3.0}}$&83.3$_{\pm\textrm{1.9}}$&59.3$_{\pm\textrm{6.0}}$&90.3$_{\pm\textrm{1.4}}$&95.2$_{\pm\textrm{0.8}}$&82.2$_{\pm\textrm{2.9}}$&53.8$_{\pm\textrm{1.7}}$&72.4$_{\pm\textrm{4.1}}$&78.66\\
&\small\bm{$m=128$}&84.5$_{\pm\textrm{5.3}}$&85.8$_{\pm\textrm{1.6}}$&83.1$_{\pm\textrm{0.8}}$&84.5$_{\pm\textrm{1.3}}$&62.1$_{\pm\textrm{3.4}}$&89.7$_{\pm\textrm{0.6}}$&95.8$_{\pm\textrm{0.5}}$&84.0$_{\pm\textrm{1.8}}$&53.6$_{\pm\textrm{3.2}}$&74.2$_{\pm\textrm{4.4}}$&79.73\\
&\small\bm{$m=256$}&85.8$_{\pm\textrm{4.0}}$&86.0$_{\pm\textrm{2.5}}$&83.2$_{\pm\textrm{1.3}}$&86.5$_{\pm\textrm{1.3}}$&62.1$_{\pm\textrm{3.4}}$&89.5$_{\pm\textrm{0.4}}$&96.1$_{\pm\textrm{0.9}}$&82.8$_{\pm\textrm{0.9}}$&53.5$_{\pm\textrm{3.3}}$&81.0$_{\pm\textrm{2.5}}$&80.66\\
&\small\bm{$m=512$}&85.5$_{\pm\textrm{4.2}}$&86.7$_{\pm\textrm{1.5}}$&83.0$_{\pm\textrm{2.4}}$&86.3$_{\pm\textrm{0.7}}$&62.1$_{\pm\textrm{3.4}}$&89.1$_{\pm\textrm{1.0}}$&96.4$_{\pm\textrm{0.8}}$&83.5$_{\pm\textrm{4.1}}$&52.8$_{\pm\textrm{3.4}}$&82.4$_{\pm\textrm{2.9}}$&80.80\\
&\small\bm{$m=1024$}&85.5$_{\pm\textrm{4.5}}$&87.6$_{\pm\textrm{1.4}}$&84.8$_{\pm\textrm{1.0}}$&87.6$_{\pm\textrm{0.6}}$&62.1$_{\pm\textrm{3.4}}$&87.5$_{\pm\textrm{1.0}}$&96.7$_{\pm\textrm{0.7}}$&83.6$_{\pm\textrm{2.3}}$&54.1$_{\pm\textrm{2.6}}$&86.7$_{\pm\textrm{1.5}}$&81.61\\
\midrule
\multirow{7}*{\textbf{1.5B}}&\small\textbf{In-Context Learning}&81.3$_{\pm\textrm{5.4}}$&64.1$_{\pm\textrm{11.3}}$&75.2$_{\pm\textrm{8.8}}$&72.7$_{\pm\textrm{18.5}}$&60.7$_{\pm\textrm{2.8}}$&66.2$_{\pm\textrm{16.7}}$&83.5$_{\pm\textrm{3.8}}$&72.2$_{\pm\textrm{13.9}}$&53.0$_{\pm\textrm{1.7}}$&54.2$_{\pm\textrm{4.9}}$&68.31\\
&\small\bm{$m=32$}&87.7$_{\pm\textrm{3.5}}$&77.0$_{\pm\textrm{3.5}}$&75.2$_{\pm\textrm{8.8}}$&86.2$_{\pm\textrm{1.8}}$&58.9$_{\pm\textrm{2.2}}$&88.2$_{\pm\textrm{3.5}}$&94.1$_{\pm\textrm{2.3}}$&83.9$_{\pm\textrm{2.4}}$&53.6$_{\pm\textrm{3.0}}$&64.8$_{\pm\textrm{4.2}}$&76.97\\
&\small\bm{$m=64$}&87.2$_{\pm\textrm{2.8}}$&82.5$_{\pm\textrm{1.2}}$&83.0$_{\pm\textrm{4.0}}$&86.2$_{\pm\textrm{1.1}}$&62.9$_{\pm\textrm{6.1}}$&89.1$_{\pm\textrm{1.6}}$&95.5$_{\pm\textrm{1.4}}$&84.3$_{\pm\textrm{1.9}}$&49.1$_{\pm\textrm{3.2}}$&68.9$_{\pm\textrm{2.9}}$&78.86\\
&\small\bm{$m=128$}&86.3$_{\pm\textrm{2.9}}$&83.8$_{\pm\textrm{2.1}}$&82.3$_{\pm\textrm{1.5}}$&87.2$_{\pm\textrm{0.4}}$&64.6$_{\pm\textrm{7.9}}$&88.9$_{\pm\textrm{2.4}}$&96.5$_{\pm\textrm{0.7}}$&86.4$_{\pm\textrm{0.8}}$&51.1$_{\pm\textrm{1.7}}$&74.0$_{\pm\textrm{2.9}}$&80.12\\
&\small\bm{$m=256$}&89.1$_{\pm\textrm{1.6}}$&85.0$_{\pm\textrm{2.2}}$&83.4$_{\pm\textrm{1.5}}$&87.5$_{\pm\textrm{1.7}}$&64.6$_{\pm\textrm{7.9}}$&89.3$_{\pm\textrm{0.9}}$&97.3$_{\pm\textrm{0.3}}$&85.6$_{\pm\textrm{2.0}}$&52.7$_{\pm\textrm{3.3}}$&78.1$_{\pm\textrm{3.3}}$&81.27\\
&\small\bm{$m=512$}&89.0$_{\pm\textrm{1.5}}$&87.6$_{\pm\textrm{1.6}}$&83.7$_{\pm\textrm{1.2}}$&88.4$_{\pm\textrm{1.5}}$&64.6$_{\pm\textrm{7.9}}$&90.1$_{\pm\textrm{1.0}}$&97.6$_{\pm\textrm{0.5}}$&85.1$_{\pm\textrm{1.6}}$&52.9$_{\pm\textrm{1.8}}$&80.5$_{\pm\textrm{1.8}}$&81.94\\
&\small\bm{$m=1024$}&88.2$_{\pm\textrm{1.0}}$&88.4$_{\pm\textrm{1.8}}$&84.1$_{\pm\textrm{1.2}}$&89.1$_{\pm\textrm{1.2}}$&64.6$_{\pm\textrm{7.9}}$&86.7$_{\pm\textrm{1.3}}$&97.8$_{\pm\textrm{0.6}}$&84.4$_{\pm\textrm{1.5}}$&53.3$_{\pm\textrm{4.2}}$&83.0$_{\pm\textrm{1.4}}$&81.96\\
\midrule
\multirow{7}*{\textbf{2.7B}}&\small\textbf{In-Context Learning}&89.9$_{\pm\textrm{4.5}}$&77.7$_{\pm\textrm{5.8}}$&84.5$_{\pm\textrm{2.1}}$&78.8$_{\pm\textrm{4.0}}$&51.1$_{\pm\textrm{6.1}}$&92.6$_{\pm\textrm{0.6}}$&88.8$_{\pm\textrm{1.3}}$&92.5$_{\pm\textrm{1.0}}$&52.4$_{\pm\textrm{4.2}}$&64.6$_{\pm\textrm{3.0}}$&77.29\\
&\small\bm{$m=32$}&89.9$_{\pm\textrm{4.5}}$&82.3$_{\pm\textrm{4.4}}$&84.5$_{\pm\textrm{2.1}}$&84.4$_{\pm\textrm{1.2}}$&50.0$_{\pm\textrm{5.4}}$&89.9$_{\pm\textrm{3.2}}$&97.0$_{\pm\textrm{0.7}}$&91.2$_{\pm\textrm{2.6}}$&52.8$_{\pm\textrm{1.9}}$&75.1$_{\pm\textrm{5.5}}$&79.72\\
&\small\bm{$m=64$}&91.4$_{\pm\textrm{2.1}}$&84.9$_{\pm\textrm{3.7}}$&80.8$_{\pm\textrm{6.0}}$&86.2$_{\pm\textrm{1.7}}$&57.1$_{\pm\textrm{3.3}}$&91.0$_{\pm\textrm{1.7}}$&97.7$_{\pm\textrm{0.8}}$&91.5$_{\pm\textrm{1.9}}$&51.6$_{\pm\textrm{3.4}}$&79.1$_{\pm\textrm{2.8}}$&81.14\\
&\small\bm{$m=128$}&93.4$_{\pm\textrm{1.3}}$&87.5$_{\pm\textrm{2.0}}$&83.3$_{\pm\textrm{3.9}}$&86.7$_{\pm\textrm{2.4}}$&57.1$_{\pm\textrm{3.8}}$&90.2$_{\pm\textrm{2.0}}$&98.6$_{\pm\textrm{0.6}}$&91.4$_{\pm\textrm{1.5}}$&53.4$_{\pm\textrm{5.6}}$&80.8$_{\pm\textrm{2.0}}$&82.25\\
&\small\bm{$m=256$}&92.9$_{\pm\textrm{2.0}}$&87.7$_{\pm\textrm{1.7}}$&83.3$_{\pm\textrm{2.4}}$&86.4$_{\pm\textrm{1.3}}$&57.1$_{\pm\textrm{3.8}}$&91.6$_{\pm\textrm{1.5}}$&99.0$_{\pm\textrm{0.2}}$&90.5$_{\pm\textrm{1.7}}$&53.5$_{\pm\textrm{3.2}}$&81.2$_{\pm\textrm{2.5}}$&82.32\\
&\small\bm{$m=512$}&91.9$_{\pm\textrm{1.0}}$&89.4$_{\pm\textrm{1.0}}$&83.8$_{\pm\textrm{2.3}}$&88.1$_{\pm\textrm{2.1}}$&57.1$_{\pm\textrm{3.8}}$&91.3$_{\pm\textrm{1.2}}$&99.1$_{\pm\textrm{0.4}}$&91.2$_{\pm\textrm{1.4}}$&54.3$_{\pm\textrm{1.6}}$&83.1$_{\pm\textrm{4.0}}$&82.93\\
&\small\bm{$m=1024$}&91.7$_{\pm\textrm{2.3}}$&90.5$_{\pm\textrm{1.5}}$&83.7$_{\pm\textrm{2.2}}$&89.1$_{\pm\textrm{1.2}}$&57.1$_{\pm\textrm{3.8}}$&89.8$_{\pm\textrm{1.0}}$&99.1$_{\pm\textrm{0.3}}$&90.9$_{\pm\textrm{1.2}}$&57.7$_{\pm\textrm{3.0}}$&86.7$_{\pm\textrm{2.3}}$&83.64\\
\midrule
\multirow{7}*{\textbf{6B}}&\small\textbf{In-Context Learning}&92.7$_{\pm\textrm{1.8}}$&81.6$_{\pm\textrm{4.6}}$&86.2$_{\pm\textrm{1.9}}$&68.8$_{\pm\textrm{8.2}}$&52.5$_{\pm\textrm{9.7}}$&92.0$_{\pm\textrm{2.6}}$&89.8$_{\pm\textrm{0.9}}$&91.6$_{\pm\textrm{1.1}}$&55.0$_{\pm\textrm{1.6}}$&62.0$_{\pm\textrm{7.3}}$&77.19\\
&\small\bm{$m=32$}&92.7$_{\pm\textrm{1.8}}$&85.5$_{\pm\textrm{2.9}}$&86.2$_{\pm\textrm{1.9}}$&83.0$_{\pm\textrm{1.5}}$&65.0$_{\pm\textrm{1.6}}$&91.0$_{\pm\textrm{2.5}}$&97.4$_{\pm\textrm{0.6}}$&91.8$_{\pm\textrm{0.7}}$&55.2$_{\pm\textrm{3.2}}$&71.2$_{\pm\textrm{5.2}}$&81.90\\
&\small\bm{$m=64$}&92.7$_{\pm\textrm{0.8}}$&87.2$_{\pm\textrm{2.0}}$&85.2$_{\pm\textrm{3.9}}$&85.1$_{\pm\textrm{1.4}}$&66.4$_{\pm\textrm{6.6}}$&90.5$_{\pm\textrm{2.8}}$&98.1$_{\pm\textrm{0.5}}$&91.1$_{\pm\textrm{1.4}}$&51.6$_{\pm\textrm{4.4}}$&72.3$_{\pm\textrm{1.3}}$&82.02\\
&\small\bm{$m=128$}&92.2$_{\pm\textrm{1.2}}$&87.4$_{\pm\textrm{1.7}}$&85.2$_{\pm\textrm{2.5}}$&87.4$_{\pm\textrm{1.6}}$&63.6$_{\pm\textrm{4.1}}$&90.6$_{\pm\textrm{2.2}}$&98.5$_{\pm\textrm{0.4}}$&91.0$_{\pm\textrm{1.5}}$&55.6$_{\pm\textrm{2.0}}$&75.8$_{\pm\textrm{3.1}}$&82.73\\
&\small\bm{$m=256$}&92.9$_{\pm\textrm{0.8}}$&90.5$_{\pm\textrm{1.2}}$&84.9$_{\pm\textrm{0.9}}$&86.4$_{\pm\textrm{2.0}}$&63.6$_{\pm\textrm{4.1}}$&90.7$_{\pm\textrm{0.7}}$&99.0$_{\pm\textrm{0.2}}$&91.4$_{\pm\textrm{1.0}}$&56.2$_{\pm\textrm{2.4}}$&76.6$_{\pm\textrm{1.5}}$&83.22\\
&\small\bm{$m=512$}&93.5$_{\pm\textrm{1.3}}$&90.8$_{\pm\textrm{0.9}}$&85.5$_{\pm\textrm{1.7}}$&88.6$_{\pm\textrm{0.8}}$&63.6$_{\pm\textrm{4.1}}$&90.4$_{\pm\textrm{1.4}}$&99.0$_{\pm\textrm{0.2}}$&90.3$_{\pm\textrm{1.0}}$&59.4$_{\pm\textrm{2.0}}$&81.6$_{\pm\textrm{1.4}}$&84.27\\
&\small\bm{$m=1024$}&93.0$_{\pm\textrm{1.1}}$&90.6$_{\pm\textrm{1.6}}$&85.5$_{\pm\textrm{1.4}}$&89.5$_{\pm\textrm{1.7}}$&63.6$_{\pm\textrm{4.1}}$&88.3$_{\pm\textrm{1.6}}$&98.9$_{\pm\textrm{0.5}}$&90.4$_{\pm\textrm{1.5}}$&57.0$_{\pm\textrm{2.7}}$&81.5$_{\pm\textrm{0.9}}$&83.82\\
\midrule
\multirow{7}*{\textbf{13B}}&\small\textbf{In-Context Learning}&89.0$_{\pm\textrm{4.3}}$&91.3$_{\pm\textrm{2.4}}$&78.4$_{\pm\textrm{7.2}}$&78.1$_{\pm\textrm{5.6}}$&53.2$_{\pm\textrm{4.4}}$&93.4$_{\pm\textrm{1.1}}$&92.2$_{\pm\textrm{2.4}}$&89.9$_{\pm\textrm{2.2}}$&55.8$_{\pm\textrm{3.0}}$&55.5$_{\pm\textrm{6.1}}$&77.68\\
&\small\bm{$m=32$}&89.0$_{\pm\textrm{4.3}}$&91.1$_{\pm\textrm{1.1}}$&78.4$_{\pm\textrm{7.2}}$&84.5$_{\pm\textrm{2.2}}$&77.9$_{\pm\textrm{4.1}}$&89.5$_{\pm\textrm{4.6}}$&97.3$_{\pm\textrm{0.7}}$&91.1$_{\pm\textrm{1.4}}$&56.8$_{\pm\textrm{3.2}}$&71.4$_{\pm\textrm{4.1}}$&82.69\\
&\small\bm{$m=64$}&94.8$_{\pm\textrm{0.8}}$&90.8$_{\pm\textrm{2.1}}$&86.3$_{\pm\textrm{2.7}}$&86.0$_{\pm\textrm{1.1}}$&77.5$_{\pm\textrm{5.1}}$&91.8$_{\pm\textrm{1.1}}$&98.1$_{\pm\textrm{0.7}}$&91.1$_{\pm\textrm{2.1}}$&57.7$_{\pm\textrm{4.1}}$&70.5$_{\pm\textrm{2.7}}$&84.47\\
&\small\bm{$m=128$}&94.8$_{\pm\textrm{0.6}}$&90.1$_{\pm\textrm{1.5}}$&86.2$_{\pm\textrm{3.8}}$&87.4$_{\pm\textrm{1.7}}$&78.9$_{\pm\textrm{4.8}}$&89.6$_{\pm\textrm{1.1}}$&98.9$_{\pm\textrm{0.5}}$&92.0$_{\pm\textrm{0.5}}$&58.2$_{\pm\textrm{4.4}}$&73.1$_{\pm\textrm{2.9}}$&84.93\\
&\small\bm{$m=256$}&94.5$_{\pm\textrm{0.9}}$&90.5$_{\pm\textrm{1.6}}$&85.3$_{\pm\textrm{0.7}}$&87.2$_{\pm\textrm{1.8}}$&78.9$_{\pm\textrm{4.8}}$&91.2$_{\pm\textrm{1.9}}$&99.5$_{\pm\textrm{0.2}}$&92.0$_{\pm\textrm{1.4}}$&59.3$_{\pm\textrm{1.5}}$&74.8$_{\pm\textrm{4.1}}$&85.31\\
&\small\bm{$m=512$}&94.5$_{\pm\textrm{0.8}}$&91.2$_{\pm\textrm{2.1}}$&86.3$_{\pm\textrm{0.9}}$&86.9$_{\pm\textrm{1.5}}$&78.9$_{\pm\textrm{4.8}}$&90.5$_{\pm\textrm{1.3}}$&99.3$_{\pm\textrm{0.3}}$&92.7$_{\pm\textrm{0.3}}$&64.1$_{\pm\textrm{4.8}}$&75.9$_{\pm\textrm{4.1}}$&86.03\\
&\small\bm{$m=1024$}&94.5$_{\pm\textrm{0.4}}$&91.6$_{\pm\textrm{2.4}}$&87.0$_{\pm\textrm{2.1}}$&88.5$_{\pm\textrm{1.0}}$&78.9$_{\pm\textrm{4.8}}$&88.1$_{\pm\textrm{1.3}}$&99.3$_{\pm\textrm{0.2}}$&91.6$_{\pm\textrm{0.3}}$&63.0$_{\pm\textrm{3.6}}$&76.1$_{\pm\textrm{6.2}}$&85.87\\
\midrule
\multirow{7}*{\textbf{30B}}&\small\textbf{In-Context Learning}&90.8$_{\pm\textrm{4.1}}$&83.5$_{\pm\textrm{8.7}}$&80.7$_{\pm\textrm{1.9}}$&74.8$_{\pm\textrm{4.6}}$&64.6$_{\pm\textrm{8.3}}$&87.7$_{\pm\textrm{3.9}}$&93.3$_{\pm\textrm{0.4}}$&93.4$_{\pm\textrm{1.0}}$&61.6$_{\pm\textrm{2.7}}$&71.7$_{\pm\textrm{2.7}}$&80.21\\
&\small\bm{$m=32$}&90.8$_{\pm\textrm{4.1}}$&91.3$_{\pm\textrm{0.6}}$&80.7$_{\pm\textrm{1.9}}$&84.1$_{\pm\textrm{1.2}}$&70.0$_{\pm\textrm{9.9}}$&89.0$_{\pm\textrm{3.5}}$&98.3$_{\pm\textrm{0.2}}$&94.5$_{\pm\textrm{1.1}}$&59.5$_{\pm\textrm{3.6}}$&77.6$_{\pm\textrm{2.3}}$&83.56\\
&\small\bm{$m=64$}&93.6$_{\pm\textrm{1.6}}$&91.9$_{\pm\textrm{1.4}}$&85.5$_{\pm\textrm{1.8}}$&85.3$_{\pm\textrm{2.1}}$&69.3$_{\pm\textrm{7.6}}$&91.6$_{\pm\textrm{1.6}}$&98.3$_{\pm\textrm{0.2}}$&94.1$_{\pm\textrm{0.6}}$&60.2$_{\pm\textrm{6.0}}$&81.0$_{\pm\textrm{2.9}}$&85.08\\
&\small\bm{$m=128$}&94.3$_{\pm\textrm{0.9}}$&92.7$_{\pm\textrm{1.7}}$&84.5$_{\pm\textrm{0.9}}$&87.1$_{\pm\textrm{1.4}}$&70.7$_{\pm\textrm{8.2}}$&91.0$_{\pm\textrm{1.2}}$&98.8$_{\pm\textrm{0.5}}$&93.1$_{\pm\textrm{1.5}}$&61.7$_{\pm\textrm{4.8}}$&79.8$_{\pm\textrm{2.0}}$&85.38\\
&\small\bm{$m=256$}&94.4$_{\pm\textrm{0.4}}$&93.8$_{\pm\textrm{1.1}}$&84.1$_{\pm\textrm{1.1}}$&87.6$_{\pm\textrm{0.9}}$&70.7$_{\pm\textrm{8.2}}$&90.3$_{\pm\textrm{1.0}}$&99.1$_{\pm\textrm{0.3}}$&93.7$_{\pm\textrm{0.6}}$&62.1$_{\pm\textrm{2.5}}$&82.1$_{\pm\textrm{1.9}}$&85.80\\
&\small\bm{$m=512$}&94.1$_{\pm\textrm{0.3}}$&94.1$_{\pm\textrm{1.4}}$&85.6$_{\pm\textrm{1.5}}$&87.9$_{\pm\textrm{1.9}}$&70.7$_{\pm\textrm{8.2}}$&89.8$_{\pm\textrm{1.5}}$&99.1$_{\pm\textrm{0.3}}$&93.2$_{\pm\textrm{1.1}}$&61.8$_{\pm\textrm{2.6}}$&83.8$_{\pm\textrm{1.8}}$&86.01\\
&\small\bm{$m=1024$}&94.1$_{\pm\textrm{0.6}}$&93.9$_{\pm\textrm{0.8}}$&85.4$_{\pm\textrm{0.7}}$&87.9$_{\pm\textrm{2.3}}$&70.7$_{\pm\textrm{8.2}}$&88.4$_{\pm\textrm{1.1}}$&99.1$_{\pm\textrm{0.2}}$&93.4$_{\pm\textrm{1.6}}$&61.0$_{\pm\textrm{4.0}}$&86.2$_{\pm\textrm{1.5}}$&86.02\\
\bottomrule
\end{tabular}}
\caption{Full results for data scaling, corresponds to Figure~\ref{fig:fullysupervisedscaling}. Some ICL results are reused\protect\footnotemark.}
\label{table:fullysupervisedscaling}
\end{table*}
\footnotetext{In Table~\ref{table:fullysupervisedscaling}, there are few cases where $k$NN Prompting gets identical results with ICL baseline. This happens when $M_T$=32, which leaves $|\mathcal{A}|=0$ as we have invariably set $|\mathcal{D}|=M_T$. To avoid exploiting exceptional split strategies for such specific case, we simply re-use the Underperformed ICL baseline results. In general, this only occurs on MPQA (when $m=32$) and SST2 (when $m=32$ and LLM $>$ 2.7B, 2048 tokens context), and should have few impact on the overall results and conclusion. Similar situation happens in Table~\ref{table:comparisontoretrievetocompose}}

We provide comprehensive results of $k$NN Prompting across data scales and LLM scales in Table~\ref{table:fullysupervisedscaling}, corresponding to Figure~\ref{fig:fullysupervisedscaling} in the main manuscript.

\subsection{Full Results of Comparison to PromptCompose}\label{appendix:comparetopreretrieval}
We provide the full results of comparison between $k$NN Prompting and PromptCompose in Table~\ref{table:comparisontoretrievetocompose}, corresponding to Figure~\ref{fig:comparisontoretrievetocompose} in the main manuscript. The employed sentence encoder can be found at \url{https://huggingface.co/models}\footnote{\url{https://huggingface.co/sentence-transformers/all-MiniLM-L6-v2}}\footnote{\url{https://huggingface.co/cambridgeltl/trans-encoder-bi-simcse-bert-base}}\footnote{\url{https://huggingface.co/princeton-nlp/unsup-simcse-bert-base-uncased}}.

\begin{table*}[h]
\centering
\resizebox{\columnwidth}{!}{
\begin{tabular}{{ll|cccccccccc|c}}
\toprule
\multicolumn{2}{c|}{\textbf{Setting \& Methods}}&\textbf{SST2}&\textbf{SUBJ}&\textbf{MPQA}&\textbf{AGNews}&\textbf{CB}&\textbf{CR}&\textbf{DBPedia}&\textbf{MR}&\textbf{RTE}&\textbf{TREC}&\textbf{AVG}\\
\midrule
$m=M_T$&\small\textbf{In-Context Learning}&81.3$_{\pm\textrm{5.4}}$&64.1$_{\pm\textrm{11.3}}$&75.2$_{\pm\textrm{8.8}}$&72.7$_{\pm\textrm{18.5}}$&60.7$_{\pm\textrm{2.8}}$&66.2$_{\pm\textrm{16.7}}$&83.5$_{\pm\textrm{3.8}}$&72.2$_{\pm\textrm{13.9}}$&53.0$_{\pm\textrm{1.7}}$&54.2$_{\pm\textrm{4.9}}$&68.31\\
\midrule
\multirow{5}*{$m=32$}&\small\textbf{BM25}&68.4$_{\pm\textrm{3.7}}$&63.6$_{\pm\textrm{3.0}}$&75.2$_{\pm\textrm{8.8}}$&69.0$_{\pm\textrm{3.6}}$&65.0$_{\pm\textrm{6.5}}$&55.2$_{\pm\textrm{1.0}}$&80.7$_{\pm\textrm{1.3}}$&59.4$_{\pm\textrm{1.4}}$&53.0$_{\pm\textrm{2.3}}$&65.5$_{\pm\textrm{2.8}}$&65.50\\
&\small\textbf{SBERT}&71.0$_{\pm\textrm{4.9}}$&67.5$_{\pm\textrm{1.6}}$&75.2$_{\pm\textrm{8.8}}$&82.3$_{\pm\textrm{2.1}}$&62.9$_{\pm\textrm{2.9}}$&57.8$_{\pm\textrm{1.9}}$&83.8$_{\pm\textrm{1.2}}$&57.7$_{\pm\textrm{4.3}}$&51.2$_{\pm\textrm{3.1}}$&58.4$_{\pm\textrm{6.7}}$&66.78\\
&\small\textbf{SimCSE}&68.1$_{\pm\textrm{4.5}}$&69.8$_{\pm\textrm{3.1}}$&75.2$_{\pm\textrm{8.8}}$&80.7$_{\pm\textrm{2.9}}$&66.4$_{\pm\textrm{2.3}}$&55.9$_{\pm\textrm{2.8}}$&82.3$_{\pm\textrm{1.9}}$&57.3$_{\pm\textrm{2.4}}$&52.8$_{\pm\textrm{1.7}}$&54.5$_{\pm\textrm{1.8}}$&66.31\\
&\small\textbf{Trans-Encoder}&67.9$_{\pm\textrm{3.8}}$&70.9$_{\pm\textrm{4.1}}$&75.2$_{\pm\textrm{8.8}}$&77.7$_{\pm\textrm{2.5}}$&61.4$_{\pm\textrm{6.0}}$&56.7$_{\pm\textrm{1.7}}$&82.8$_{\pm\textrm{2.0}}$&57.3$_{\pm\textrm{4.3}}$&52.7$_{\pm\textrm{2.0}}$&59.3$_{\pm\textrm{3.9}}$&66.21\\
&\small\textbf{$k$NN Prompting}&87.7$_{\pm\textrm{3.5}}$&77.0$_{\pm\textrm{3.5}}$&75.2$_{\pm\textrm{8.8}}$&86.2$_{\pm\textrm{1.8}}$&58.9$_{\pm\textrm{2.2}}$&88.2$_{\pm\textrm{3.5}}$&94.1$_{\pm\textrm{2.3}}$&83.9$_{\pm\textrm{2.4}}$&53.6$_{\pm\textrm{3.0}}$&64.8$_{\pm\textrm{4.2}}$&\textbf{76.97}\\
\midrule
\multirow{5}*{$m=64$}&\small\textbf{BM25}&69.7$_{\pm\textrm{2.9}}$&67.7$_{\pm\textrm{2.5}}$&79.4$_{\pm\textrm{2.0}}$&71.7$_{\pm\textrm{1.9}}$&68.6$_{\pm\textrm{4.1}}$&54.7$_{\pm\textrm{1.1}}$&83.4$_{\pm\textrm{2.0}}$&58.7$_{\pm\textrm{1.5}}$&52.0$_{\pm\textrm{1.6}}$&65.9$_{\pm\textrm{5.6}}$&67.18\\
&\small\textbf{SBERT}&71.8$_{\pm\textrm{3.4}}$&71.6$_{\pm\textrm{3.8}}$&80.2$_{\pm\textrm{1.5}}$&84.6$_{\pm\textrm{2.2}}$&66.8$_{\pm\textrm{6.8}}$&59.8$_{\pm\textrm{1.8}}$&84.6$_{\pm\textrm{0.8}}$&57.6$_{\pm\textrm{1.6}}$&52.8$_{\pm\textrm{3.6}}$&63.7$_{\pm\textrm{4.7}}$&69.37\\
&\small\textbf{SimCSE}&68.4$_{\pm\textrm{4.7}}$&69.7$_{\pm\textrm{2.6}}$&81.6$_{\pm\textrm{0.6}}$&83.1$_{\pm\textrm{2.2}}$&71.4$_{\pm\textrm{4.6}}$&57.9$_{\pm\textrm{2.6}}$&84.8$_{\pm\textrm{2.1}}$&57.3$_{\pm\textrm{2.0}}$&52.3$_{\pm\textrm{3.6}}$&58.1$_{\pm\textrm{1.8}}$&68.46\\
&\small\textbf{Trans-Encoder}&69.3$_{\pm\textrm{4.3}}$&73.0$_{\pm\textrm{1.2}}$&82.0$_{\pm\textrm{2.2}}$&79.3$_{\pm\textrm{2.0}}$&69.3$_{\pm\textrm{3.4}}$&57.9$_{\pm\textrm{2.7}}$&85.6$_{\pm\textrm{1.2}}$&57.9$_{\pm\textrm{1.4}}$&52.1$_{\pm\textrm{4.5}}$&60.1$_{\pm\textrm{2.6}}$&68.65\\
&\small\textbf{$k$NN Prompting}&87.2$_{\pm\textrm{2.8}}$&82.5$_{\pm\textrm{1.2}}$&83.0$_{\pm\textrm{4.0}}$&86.2$_{\pm\textrm{1.1}}$&62.9$_{\pm\textrm{6.1}}$&89.1$_{\pm\textrm{1.6}}$&95.5$_{\pm\textrm{1.4}}$&84.3$_{\pm\textrm{1.9}}$&49.1$_{\pm\textrm{3.2}}$&68.9$_{\pm\textrm{2.9}}$&\textbf{78.86}\\
\midrule
\multirow{5}*{$m=128$}&\small\textbf{BM25}&69.1$_{\pm\textrm{0.5}}$&66.8$_{\pm\textrm{2.7}}$&75.2$_{\pm\textrm{6.2}}$&77.5$_{\pm\textrm{1.4}}$&71.4$_{\pm\textrm{0.0}}$&56.4$_{\pm\textrm{1.2}}$&85.5$_{\pm\textrm{1.7}}$&59.8$_{\pm\textrm{2.0}}$&54.5$_{\pm\textrm{1.3}}$&72.3$_{\pm\textrm{4.9}}$&68.87\\
&\small\textbf{SBERT}&71.7$_{\pm\textrm{1.9}}$&71.9$_{\pm\textrm{1.7}}$&79.6$_{\pm\textrm{3.6}}$&85.3$_{\pm\textrm{2.0}}$&69.6$_{\pm\textrm{0.0}}$&58.8$_{\pm\textrm{0.8}}$&87.2$_{\pm\textrm{0.9}}$&60.1$_{\pm\textrm{2.3}}$&53.3$_{\pm\textrm{2.3}}$&64.9$_{\pm\textrm{2.1}}$&70.24\\
&\small\textbf{SimCSE}&70.9$_{\pm\textrm{2.2}}$&71.6$_{\pm\textrm{3.2}}$&81.6$_{\pm\textrm{2.6}}$&84.5$_{\pm\textrm{1.7}}$&73.2$_{\pm\textrm{0.0}}$&58.5$_{\pm\textrm{2.2}}$&87.0$_{\pm\textrm{2.1}}$&58.7$_{\pm\textrm{1.4}}$&53.4$_{\pm\textrm{3.2}}$&59.6$_{\pm\textrm{3.3}}$&69.89\\
&\small\textbf{Trans-Encoder}&69.0$_{\pm\textrm{1.3}}$&75.5$_{\pm\textrm{2.2}}$&82.6$_{\pm\textrm{1.9}}$&82.9$_{\pm\textrm{0.6}}$&73.2$_{\pm\textrm{0.0}}$&56.1$_{\pm\textrm{0.8}}$&87.0$_{\pm\textrm{1.4}}$&57.1$_{\pm\textrm{1.3}}$&52.9$_{\pm\textrm{3.1}}$&63.9$_{\pm\textrm{2.3}}$&70.02\\
&\small\textbf{$k$NN Prompting}&86.3$_{\pm\textrm{2.9}}$&83.8$_{\pm\textrm{2.1}}$&82.3$_{\pm\textrm{1.5}}$&87.2$_{\pm\textrm{0.4}}$&64.6$_{\pm\textrm{7.9}}$&88.9$_{\pm\textrm{2.4}}$&96.5$_{\pm\textrm{0.7}}$&86.4$_{\pm\textrm{0.8}}$&51.1$_{\pm\textrm{1.7}}$&74.0$_{\pm\textrm{2.9}}$&\textbf{80.12}\\
\midrule
\multirow{5}*{$m=256$}&\small\textbf{BM25}&72.0$_{\pm\textrm{3.9}}$&72.3$_{\pm\textrm{1.4}}$&78.8$_{\pm\textrm{3.2}}$&77.3$_{\pm\textrm{3.0}}$&71.4$_{\pm\textrm{0.0}}$&57.1$_{\pm\textrm{0.8}}$&88.2$_{\pm\textrm{1.9}}$&58.4$_{\pm\textrm{1.6}}$&53.8$_{\pm\textrm{2.8}}$&76.6$_{\pm\textrm{3.1}}$&70.60\\
&\small\textbf{SBERT}&69.9$_{\pm\textrm{1.8}}$&72.3$_{\pm\textrm{0.8}}$&82.2$_{\pm\textrm{2.5}}$&86.3$_{\pm\textrm{1.2}}$&69.6$_{\pm\textrm{0.0}}$&58.8$_{\pm\textrm{1.5}}$&88.5$_{\pm\textrm{0.9}}$&59.8$_{\pm\textrm{2.6}}$&52.3$_{\pm\textrm{2.4}}$&69.1$_{\pm\textrm{0.8}}$&70.89\\
&\small\textbf{SimCSE}&71.4$_{\pm\textrm{3.7}}$&73.7$_{\pm\textrm{1.6}}$&82.9$_{\pm\textrm{0.8}}$&85.5$_{\pm\textrm{1.4}}$&73.2$_{\pm\textrm{0.0}}$&59.7$_{\pm\textrm{1.2}}$&89.1$_{\pm\textrm{1.7}}$&57.8$_{\pm\textrm{2.2}}$&51.1$_{\pm\textrm{2.6}}$&64.3$_{\pm\textrm{1.5}}$&70.87\\
&\small\textbf{Trans-Encoder}&70.0$_{\pm\textrm{1.0}}$&76.6$_{\pm\textrm{1.4}}$&82.1$_{\pm\textrm{2.0}}$&84.1$_{\pm\textrm{1.2}}$&73.2$_{\pm\textrm{0.0}}$&58.0$_{\pm\textrm{1.0}}$&89.9$_{\pm\textrm{1.3}}$&58.1$_{\pm\textrm{1.2}}$&52.0$_{\pm\textrm{2.4}}$&70.9$_{\pm\textrm{2.2}}$&71.49\\
&\small\textbf{$k$NN Prompting}&89.1$_{\pm\textrm{1.6}}$&85.0$_{\pm\textrm{2.2}}$&83.4$_{\pm\textrm{1.5}}$&87.5$_{\pm\textrm{1.7}}$&64.6$_{\pm\textrm{7.9}}$&89.3$_{\pm\textrm{0.9}}$&97.3$_{\pm\textrm{0.3}}$&85.6$_{\pm\textrm{2.0}}$&52.7$_{\pm\textrm{3.3}}$&78.1$_{\pm\textrm{3.3}}$&\textbf{81.27}\\
\midrule
\multirow{5}*{$m=512$}&\small\textbf{BM25}&74.6$_{\pm\textrm{2.0}}$&71.8$_{\pm\textrm{1.3}}$&79.0$_{\pm\textrm{2.3}}$&81.3$_{\pm\textrm{1.7}}$&71.4$_{\pm\textrm{0.0}}$&58.4$_{\pm\textrm{1.1}}$&88.7$_{\pm\textrm{1.9}}$&59.7$_{\pm\textrm{1.1}}$&53.7$_{\pm\textrm{2.3}}$&82.9$_{\pm\textrm{1.1}}$&72.14\\
&\small\textbf{SBERT}&71.6$_{\pm\textrm{2.8}}$&74.3$_{\pm\textrm{1.2}}$&83.1$_{\pm\textrm{2.2}}$&89.1$_{\pm\textrm{1.9}}$&69.6$_{\pm\textrm{0.0}}$&59.2$_{\pm\textrm{3.0}}$&88.6$_{\pm\textrm{1.5}}$&58.8$_{\pm\textrm{2.1}}$&51.8$_{\pm\textrm{3.0}}$&73.8$_{\pm\textrm{1.9}}$&72.00\\
&\small\textbf{SimCSE}&73.6$_{\pm\textrm{2.0}}$&75.5$_{\pm\textrm{3.1}}$&83.9$_{\pm\textrm{1.6}}$&86.6$_{\pm\textrm{1.6}}$&73.2$_{\pm\textrm{0.0}}$&59.1$_{\pm\textrm{1.0}}$&90.4$_{\pm\textrm{1.3}}$&59.2$_{\pm\textrm{2.4}}$&50.2$_{\pm\textrm{2.3}}$&69.1$_{\pm\textrm{2.1}}$&72.07\\
&\small\textbf{Trans-Encoder}&73.1$_{\pm\textrm{3.6}}$&78.9$_{\pm\textrm{1.7}}$&84.2$_{\pm\textrm{1.1}}$&87.1$_{\pm\textrm{1.7}}$&73.2$_{\pm\textrm{0.0}}$&57.7$_{\pm\textrm{1.1}}$&89.6$_{\pm\textrm{1.2}}$&58.4$_{\pm\textrm{2.5}}$&52.1$_{\pm\textrm{3.7}}$&74.1$_{\pm\textrm{2.3}}$&72.84\\
&\small\textbf{$k$NN Prompting}&89.0$_{\pm\textrm{1.5}}$&87.6$_{\pm\textrm{1.6}}$&83.7$_{\pm\textrm{1.2}}$&88.4$_{\pm\textrm{1.5}}$&64.6$_{\pm\textrm{7.9}}$&90.1$_{\pm\textrm{1.0}}$&97.6$_{\pm\textrm{0.5}}$&85.1$_{\pm\textrm{1.6}}$&52.9$_{\pm\textrm{1.8}}$&80.5$_{\pm\textrm{1.8}}$&\textbf{81.94}\\
\midrule
\multirow{5}*{$m=1024$}&\small\textbf{BM25}&76.9$_{\pm\textrm{1.8}}$&74.9$_{\pm\textrm{2.7}}$&79.3$_{\pm\textrm{2.2}}$&83.6$_{\pm\textrm{1.3}}$&71.4$_{\pm\textrm{0.0}}$&56.9$_{\pm\textrm{1.0}}$&90.2$_{\pm\textrm{1.6}}$&60.3$_{\pm\textrm{2.7}}$&52.7$_{\pm\textrm{1.0}}$&85.0$_{\pm\textrm{1.2}}$&73.12\\
&\small\textbf{SBERT}&73.9$_{\pm\textrm{1.6}}$&76.6$_{\pm\textrm{2.3}}$&83.7$_{\pm\textrm{1.2}}$&90.0$_{\pm\textrm{1.6}}$&69.6$_{\pm\textrm{0.0}}$&58.1$_{\pm\textrm{1.1}}$&91.1$_{\pm\textrm{1.2}}$&59.6$_{\pm\textrm{0.9}}$&52.8$_{\pm\textrm{1.8}}$&75.7$_{\pm\textrm{2.0}}$&73.12\\
&\small\textbf{SimCSE}&73.5$_{\pm\textrm{3.0}}$&74.5$_{\pm\textrm{1.8}}$&84.6$_{\pm\textrm{1.1}}$&88.7$_{\pm\textrm{1.1}}$&73.2$_{\pm\textrm{0.0}}$&57.5$_{\pm\textrm{1.0}}$&91.6$_{\pm\textrm{0.8}}$&59.8$_{\pm\textrm{1.3}}$&53.1$_{\pm\textrm{1.0}}$&69.4$_{\pm\textrm{1.3}}$&72.59\\
&\small\textbf{Trans-Encoder}&73.4$_{\pm\textrm{2.2}}$&78.4$_{\pm\textrm{1.6}}$&82.9$_{\pm\textrm{0.8}}$&87.8$_{\pm\textrm{2.3}}$&73.2$_{\pm\textrm{0.0}}$&55.9$_{\pm\textrm{0.8}}$&91.7$_{\pm\textrm{0.8}}$&59.5$_{\pm\textrm{3.2}}$&54.8$_{\pm\textrm{2.0}}$&76.6$_{\pm\textrm{0.7}}$&73.42\\
&\small\textbf{$k$NN Prompting}&88.2$_{\pm\textrm{1.0}}$&88.4$_{\pm\textrm{1.8}}$&84.1$_{\pm\textrm{1.2}}$&89.1$_{\pm\textrm{1.2}}$&64.6$_{\pm\textrm{7.9}}$&86.7$_{\pm\textrm{1.3}}$&97.8$_{\pm\textrm{0.6}}$&84.4$_{\pm\textrm{1.5}}$&53.3$_{\pm\textrm{4.2}}$&83.0$_{\pm\textrm{1.4}}$&\textbf{81.96}\\
\bottomrule
\end{tabular}}
\caption{Full results for comparison to PromptCompose, corresponding to Figure~\ref{fig:comparisontoretrievetocompose}. Some ICL results are reused (MPQA, 32 shot), see footnote in caption of Table~\ref{table:fullysupervisedscaling} for explanation.}
\label{table:comparisontoretrievetocompose}
\end{table*}

\subsection{Benefits of Whole LM Distribution Under Fully Supervised Scenario}\label{appendix:wholevslabelfullshot}
We provide comparison between whole and partial LM distribution also on fully supervised scenario in Table~\ref{table:wholevspartialfullysupervised}.
The conclusion remains invariant with few shot scenario as demonstrated in Table~\ref{table:mainresultsfewshot}.

\begin{table*}[h!]
\centering
\resizebox{\columnwidth}{!}{
\begin{tabular}{{c|cccccccccc|c}}
\toprule
\textbf{Methods}&\textbf{SST2}&\textbf{SUBJ}&\textbf{MPQA}&\textbf{AGNews}&\textbf{CB}&\textbf{CR}&\textbf{DBPedia}&\textbf{MR}&\textbf{RTE}&\textbf{TREC}&\textbf{AVG}\\
\midrule
\small\textbf{ICL}&81.3$_{\pm\textrm{5.4}}$&64.1$_{\pm\textrm{11.3}}$&75.2$_{\pm\textrm{8.8}}$&72.7$_{\pm\textrm{18.5}}$&60.7$_{\pm\textrm{2.8}}$&66.2$_{\pm\textrm{16.7}}$&83.5$_{\pm\textrm{3.8}}$&72.2$_{\pm\textrm{13.9}}$&53.0$_{\pm\textrm{1.7}}$&54.2$_{\pm\textrm{4.9}}$&68.31\\
\small\textbf{$k$NN Prompting}&88.2$_{\pm\textrm{1.0}}$&88.4$_{\pm\textrm{1.8}}$&84.1$_{\pm\textrm{1.2}}$&89.1$_{\pm\textrm{1.2}}$&64.6$_{\pm\textrm{7.9}}$&86.7$_{\pm\textrm{1.3}}$&97.8$_{\pm\textrm{0.6}}$&84.4$_{\pm\textrm{1.5}}$&53.3$_{\pm\textrm{4.2}}$&83.0$_{\pm\textrm{1.4}}$&\textbf{81.96}\\
\small\textbf{$k$NN Prompting (Partial)}&87.7$_{\pm\textrm{1.4}}$&69.3$_{\pm\textrm{8.9}}$&83.4$_{\pm\textrm{1.1}}$&84.5$_{\pm\textrm{1.2}}$&63.2$_{\pm\textrm{6.0}}$&84.1$_{\pm\textrm{2.7}}$&96.8$_{\pm\textrm{0.6}}$&84.7$_{\pm\textrm{1.8}}$&49.0$_{\pm\textrm{2.9}}$&65.7$_{\pm\textrm{3.5}}$&76.84\\
\bottomrule
\end{tabular}}
\caption{Comparison between whole and partial LM distribution on fully supervised scenario.}
\label{table:wholevspartialfullysupervised}
\end{table*}

\subsection{Split Strategy of Demonstration and Anchor Set}\label{appendix:splitofdemoandanchor}

We provide results on more datasets regarding the investigation of split strategy in Figure~\ref{fig:demoanchorsplit}, corresponding to Section~\ref{sec:splitstrategy}.

\begin{figure*}
    \begin{subfigure}[b]{0.42\columnwidth}
        \includegraphics[width=\textwidth]{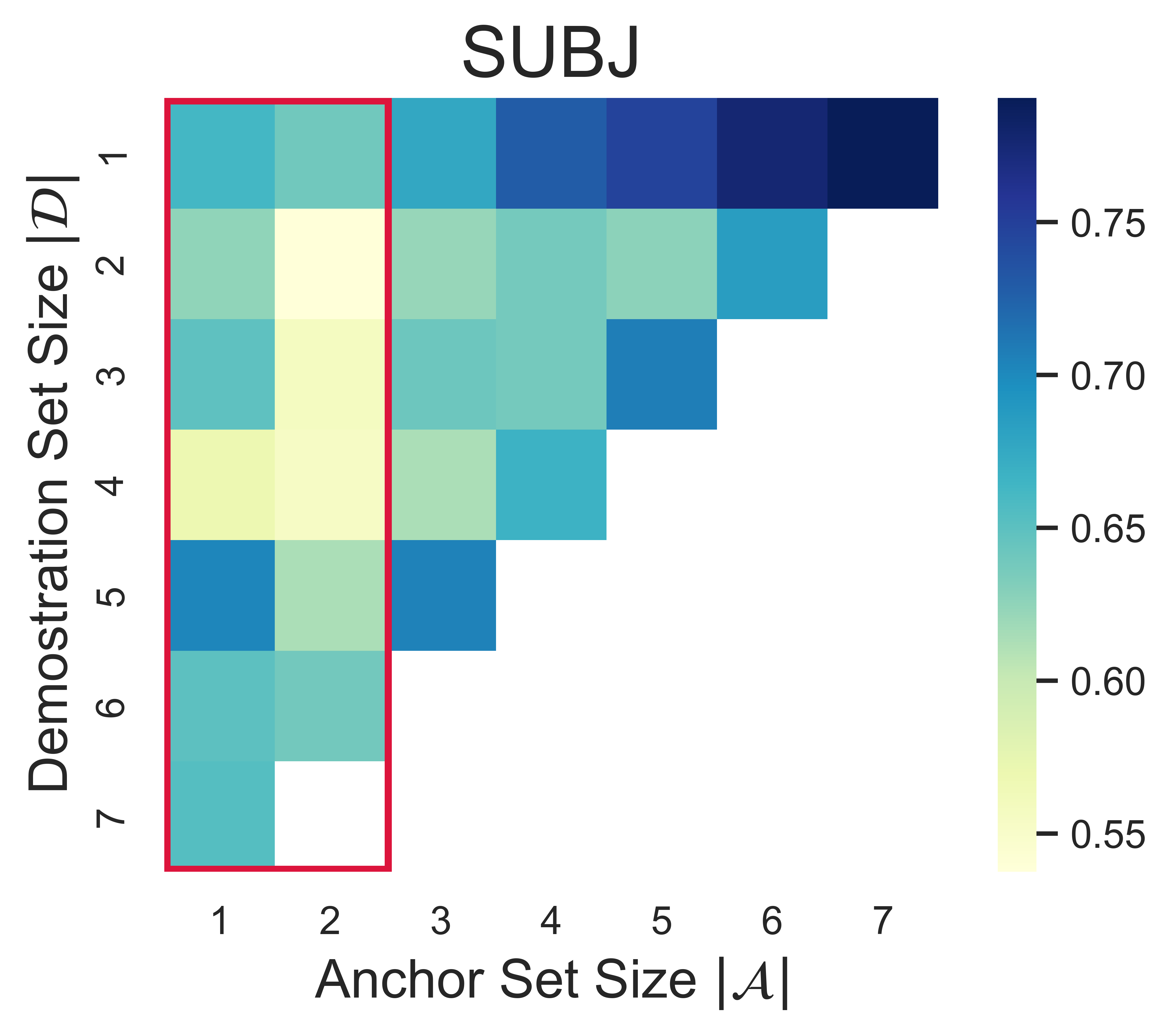}
        \label{fig:demoanchorsplitsubj}
    \end{subfigure}
    \hfill
    \begin{subfigure}[b]{0.42\columnwidth}
        \includegraphics[width=\textwidth]{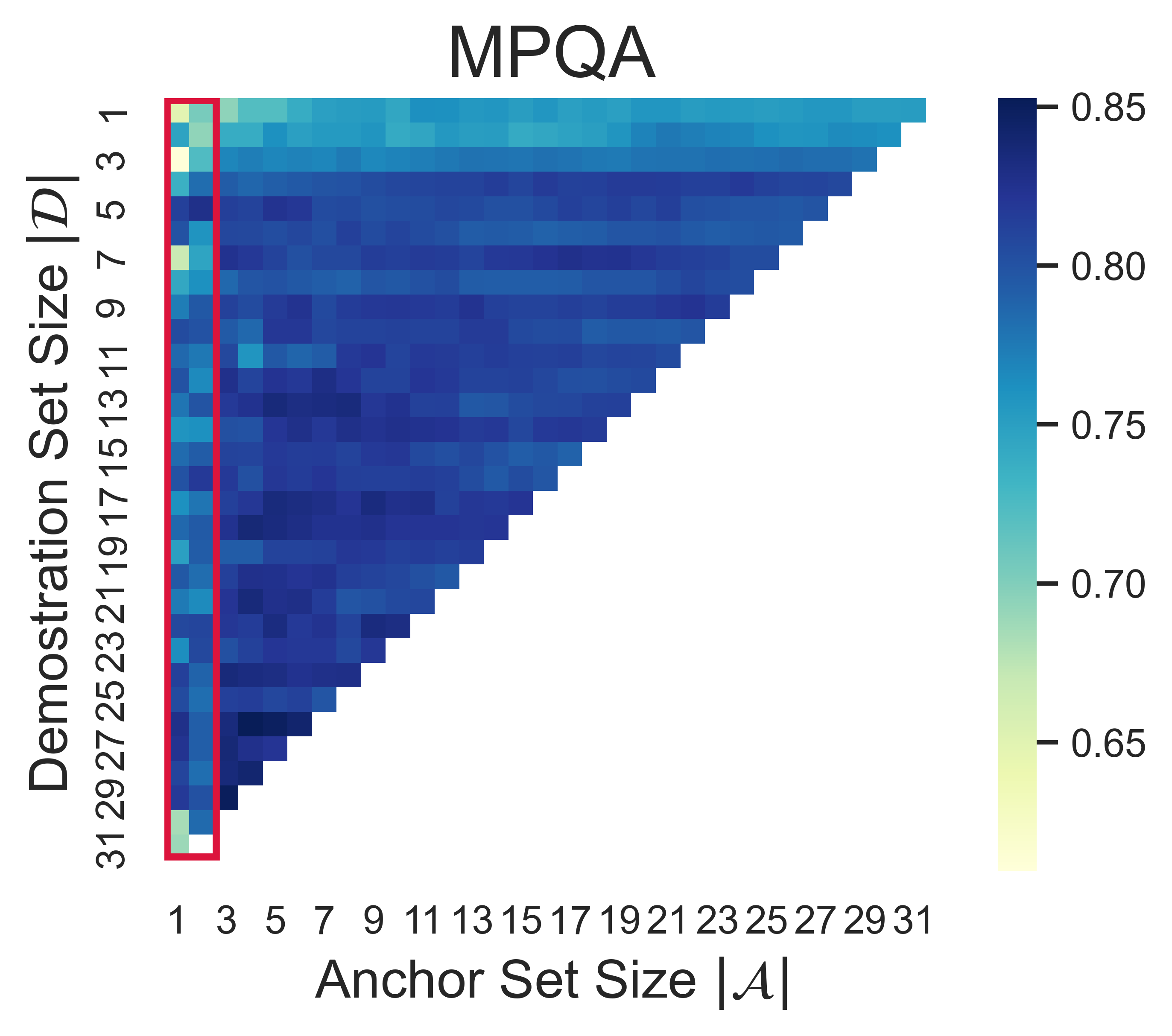}
        \label{fig:demoanchorsplitmmpqa}
    \end{subfigure}

    \begin{subfigure}[b]{0.42\columnwidth}
        \includegraphics[width=\textwidth]{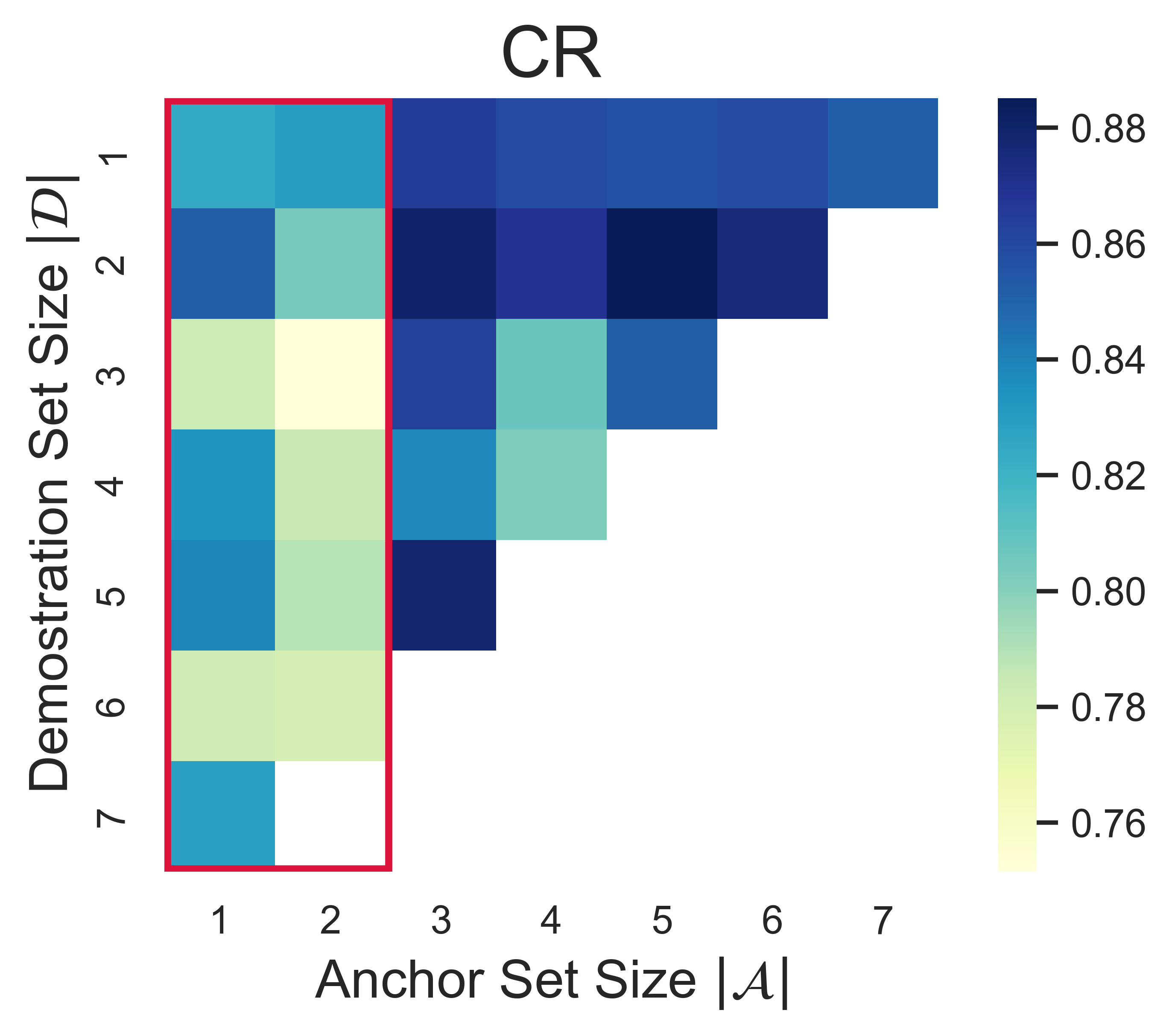}
        \label{fig:demoanchorsplitcr}
    \end{subfigure}
    \hfill
    \begin{subfigure}[b]{0.42\columnwidth}
        \includegraphics[width=\textwidth]{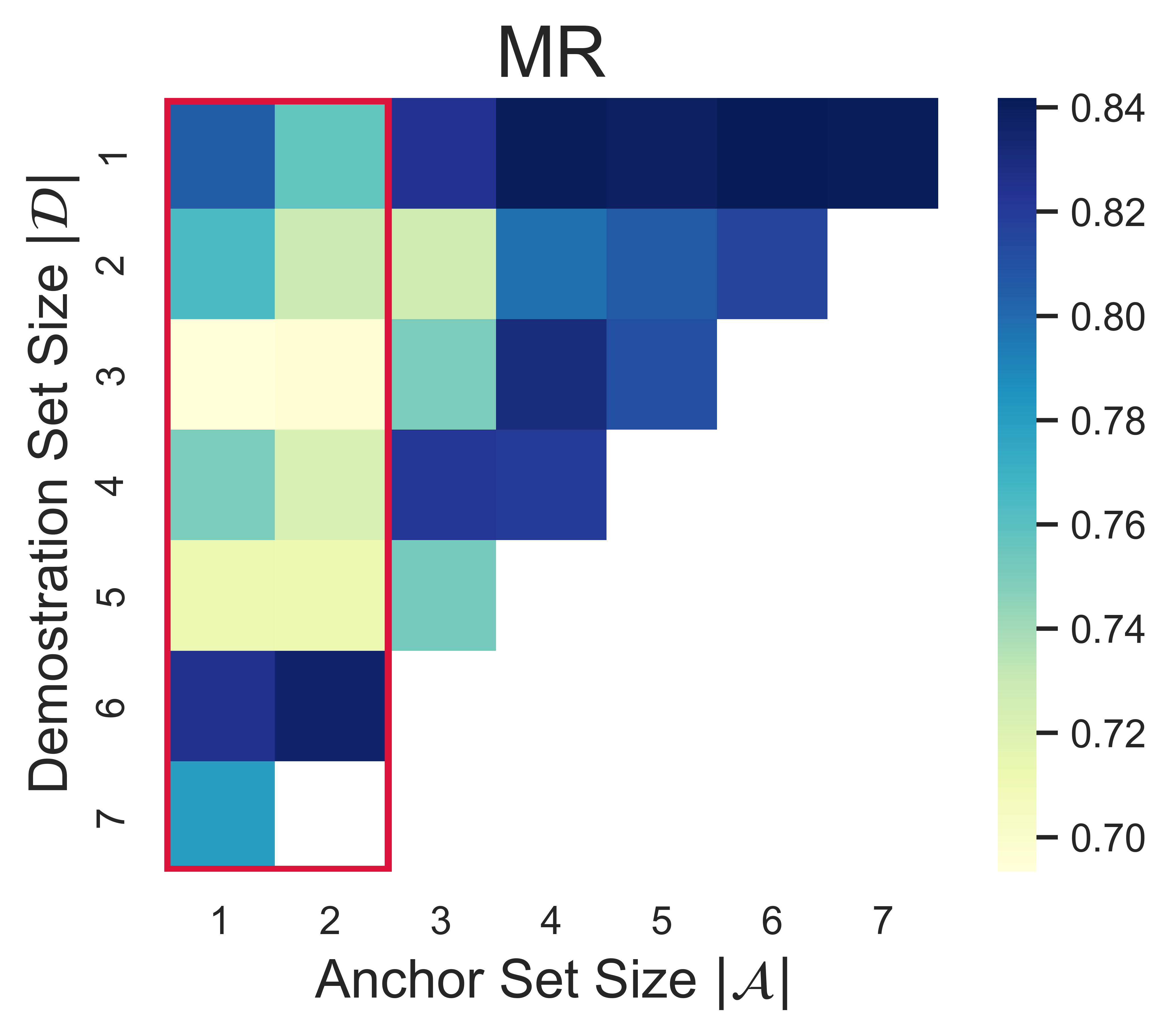}
        \label{fig:demoanchorsplitmr}
    \end{subfigure}
    \caption{Split of Demonstration and Anchor Set. $|\mathcal{A}|+|\mathcal{D}|\leq M_T$.}
    \label{fig:demoanchorsplit}
\end{figure*}

\section{Comparison to Finetuning Under Fully Supervised Scenario}\label{appendix:comparetoft}
\begin{figure*}[h!]
\centering
 \includegraphics[width=0.6\textwidth]{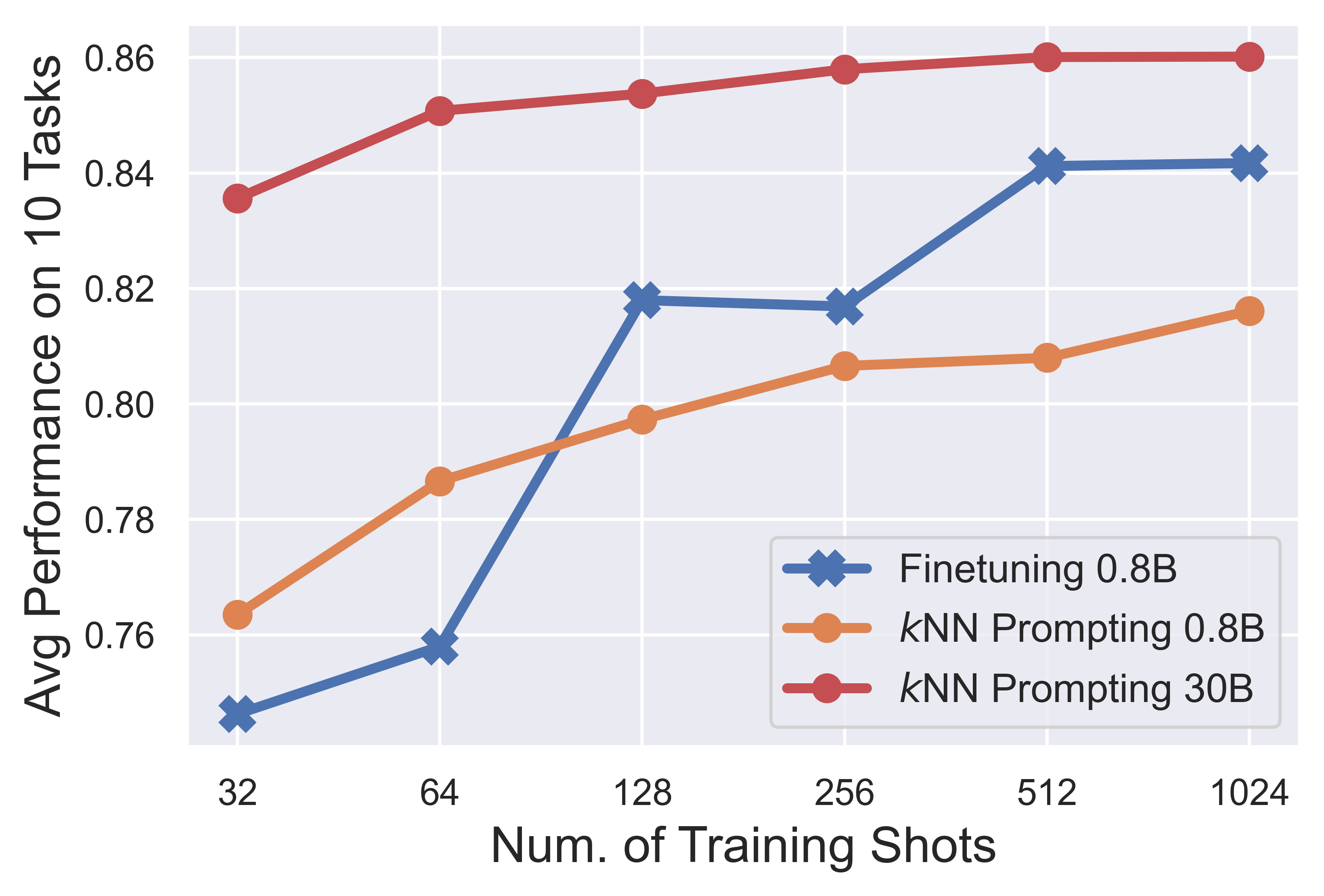}
\caption{Comparison to finetuning baseline.}
\label{fig:comparetoft}
\end{figure*}

We compare $k$NN Prompting to standard PLM finetuning in a more extensive data scale.
For finetuning baselines in both Table~\ref{table:lowresource} and Figure~\ref{fig:comparetoft}, we set hyper-parameters following previous works~\citep{schick-schutze-2021-exploiting}.
We set learning rate to 1e-5, batch size to 16, and training steps to 125, 250 or 500, respectively for $m\in \{32, 64\}, \{128, 256\}, \{512, 1024\}$.
For CB, AGNews and RTE, batch size is adjusted to 8, for DBPedia, batch size is adjusted to 4 to avoid OOM.
We observe that with the same model scale, $k$NN Prompting is superior than finetuning under the low resource setting, but inferior under fully supervised setting.
This indicates its data utility factor $\alpha$ is still smaller than finetuning.
However, the main advantage of $k$NN Prompting comes with LLM, which significantly outperforms the finetuning baseline without any gradient-based optimization.
We have also discussed this in Section~\ref{discussion}.

\section{More Case Study}\label{appendix:morecases}

\begin{figure*}[t!]
\centering
 \includegraphics[width=0.8\textwidth]{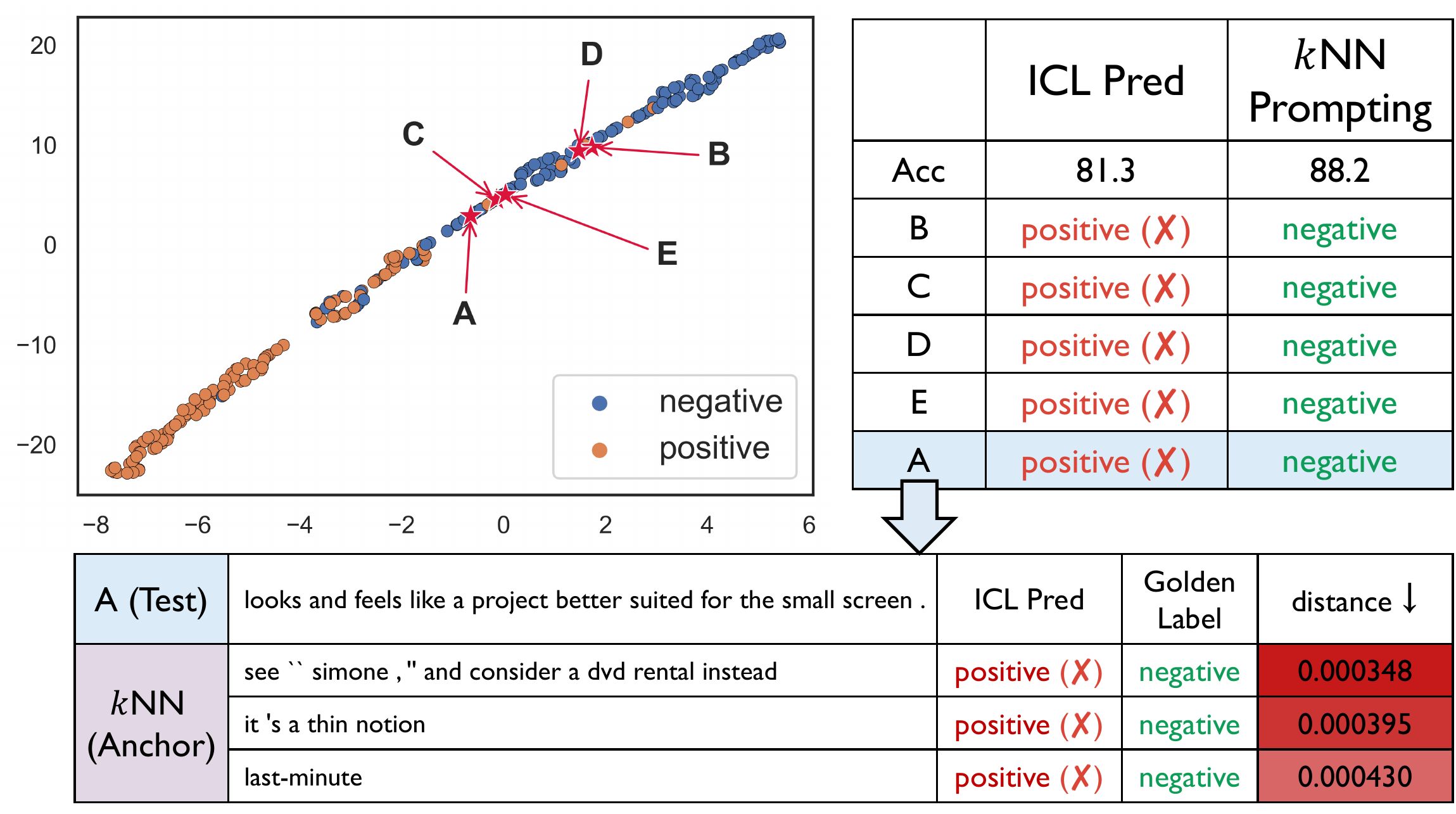}
\caption{Case study on SST2, all 5 test instances are randomly selected given that $k$NN Prompting outperforms ICL.}
\label{fig:casestudy_sst2}
\end{figure*}

\begin{figure*}[t!]
\centering
 \includegraphics[width=0.8\textwidth]{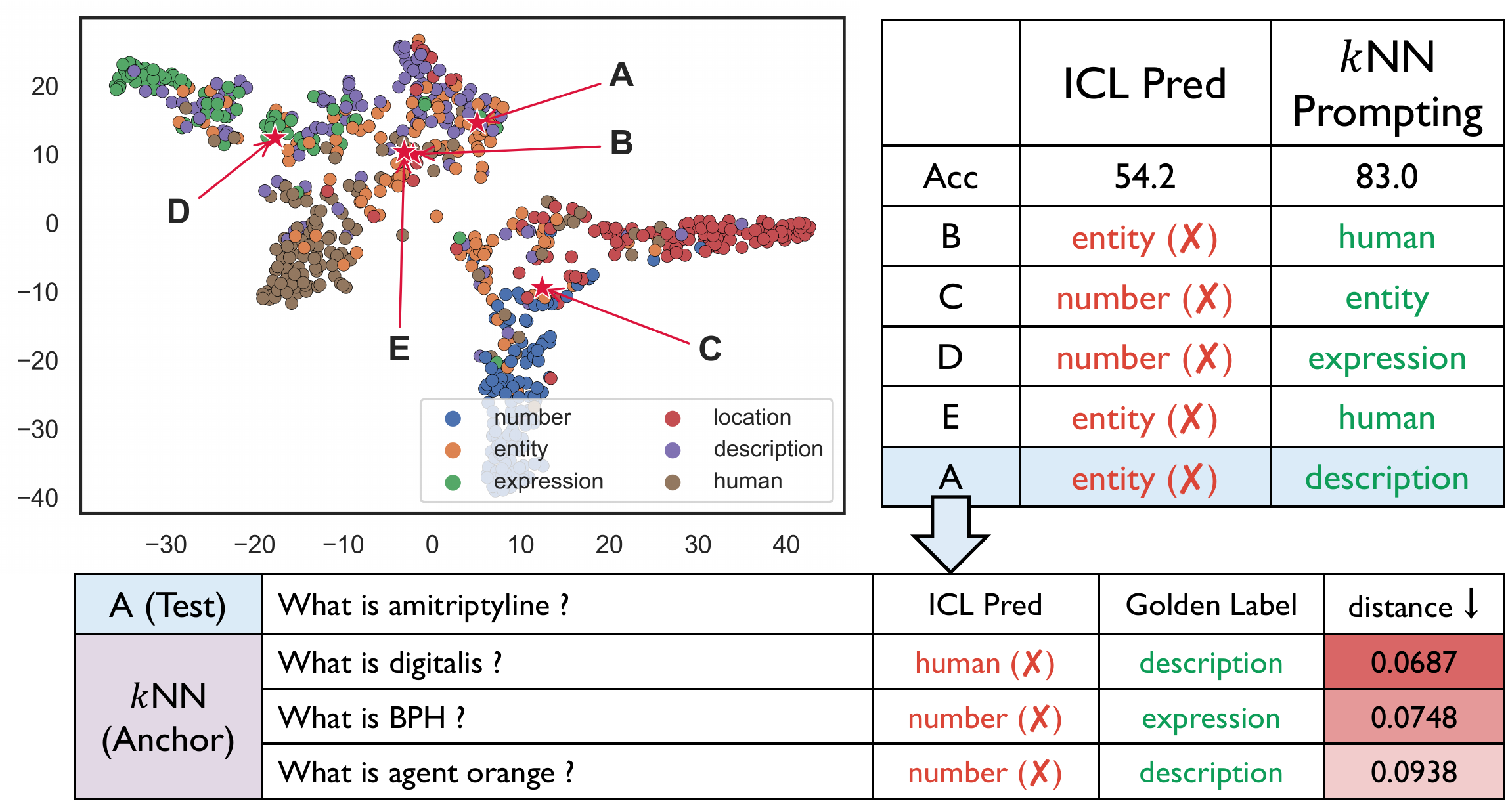}
\caption{Case study on TREC, all 5 test instances are randomly selected given that $k$NN Prompting outperforms ICL.}
\label{fig:casestudy_trec}
\end{figure*}

\begin{figure*}[t!]
\centering
 \includegraphics[width=0.9\textwidth]{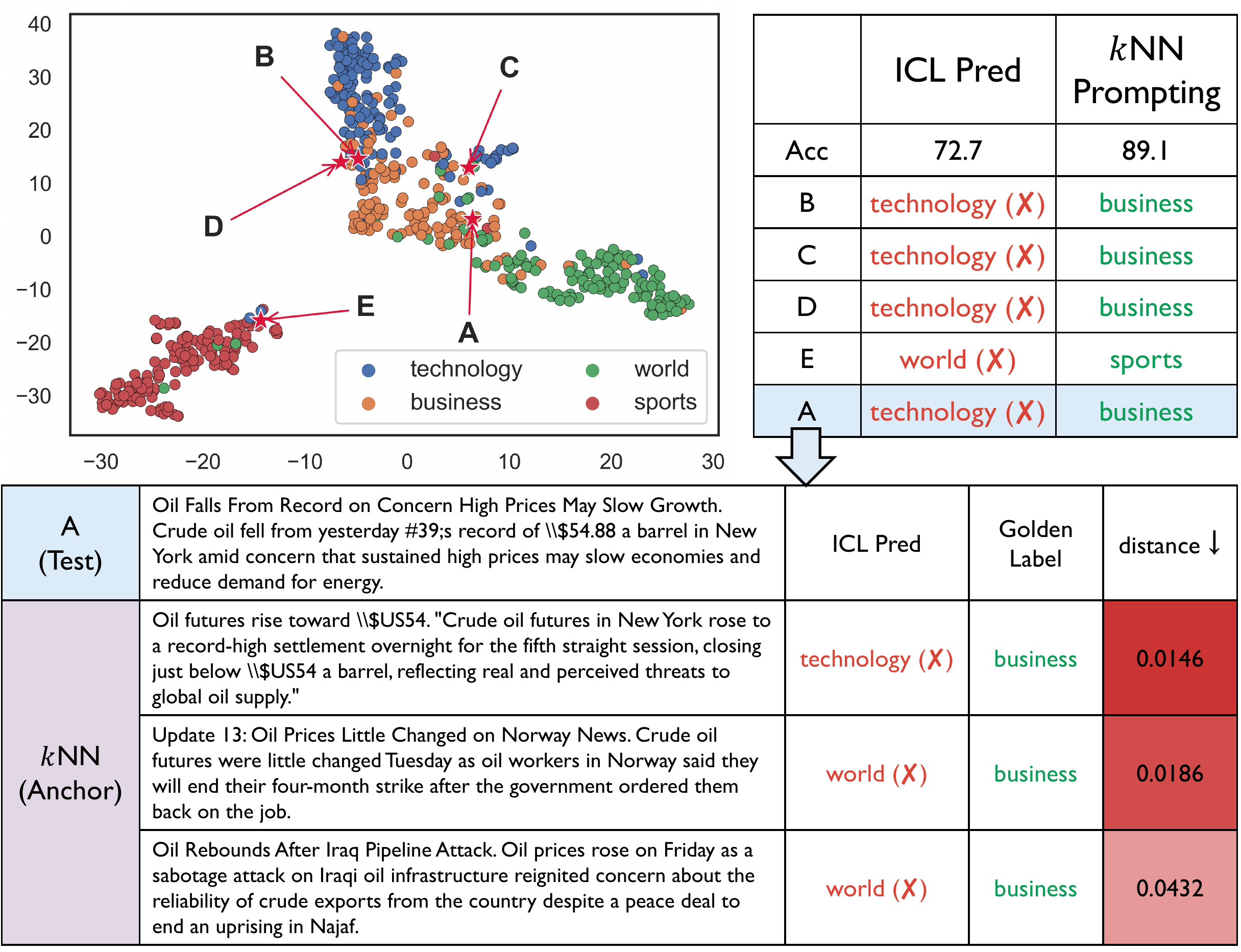}
\caption{Case study on AGNews, all 5 test instances are randomly selected given that $k$NN Prompting outperforms ICL.}
\label{fig:casestudy_agnews}
\end{figure*}

\begin{figure*}[t!]
\centering
 \includegraphics[width=0.9\textwidth]{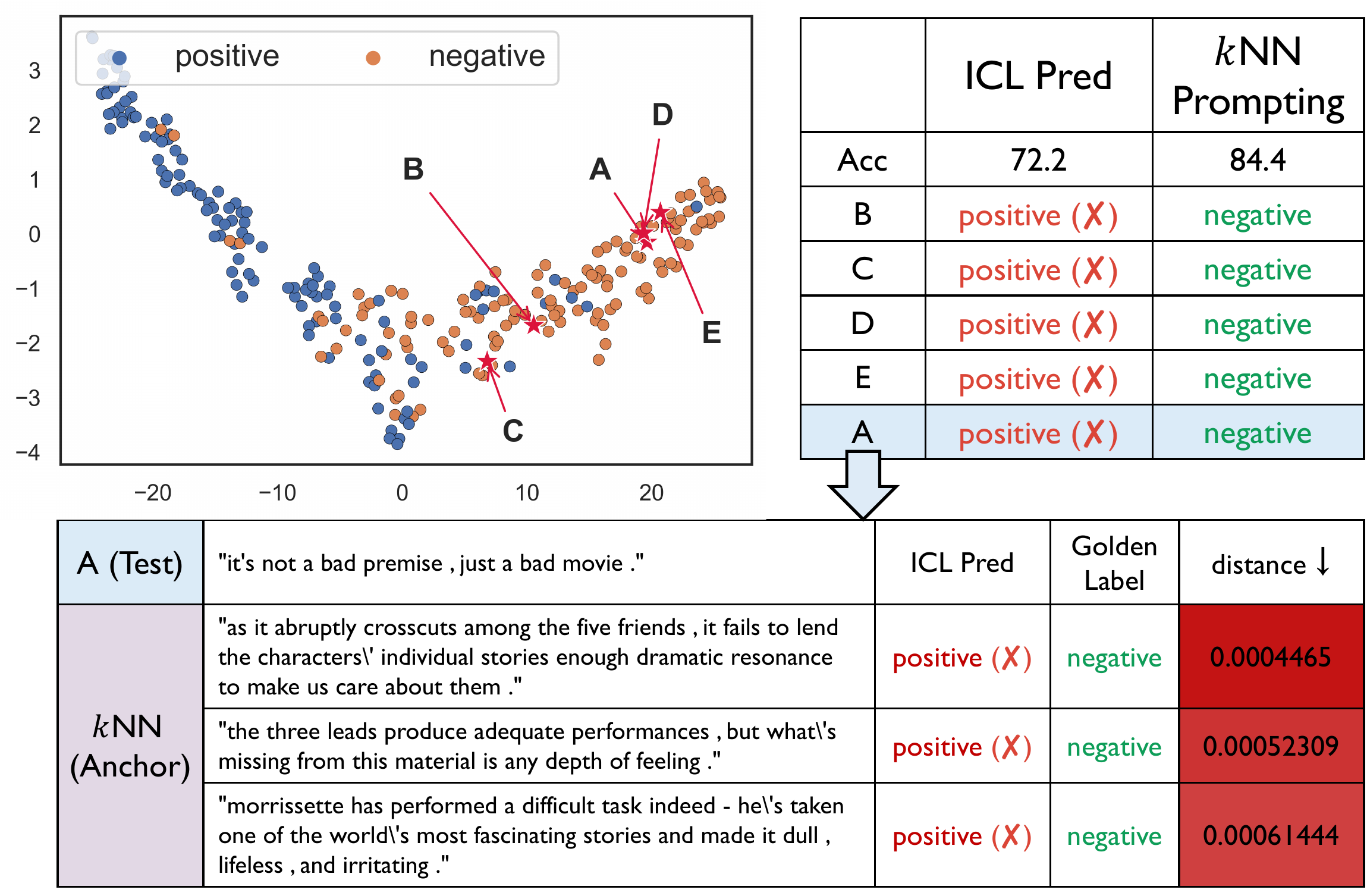}
\caption{Case study on MR, all 5 test instances are randomly selected given that $k$NN Prompting outperforms ICL.}
\label{fig:casestudy_mr}
\end{figure*}

We provide more case study, respectively on SST2, TREC, AGNews and MR (Figure~\ref{fig:casestudy_sst2},~\ref{fig:casestudy_agnews},~\ref{fig:casestudy_trec} and~\ref{fig:casestudy_mr}).

\section{Prompt Template}\label{appendix:template}

See Table~\ref{table:icltemplate} for the used templates (Adopted from~\citet{lu-etal-2022-fantastically}).
They are intuitively designed and the proposed method should be robust with choices of templates.

\newcolumntype{L}[1]{>{\raggedright\arraybackslash}p{#1}}
\newcolumntype{C}[1]{>{\centering\arraybackslash}p{#1}}
\newcolumntype{R}[1]{>{\raggedleft\arraybackslash}p{#1}}
\begin{table}[!h]
\centering
\scriptsize
\begin{tabularx}{\columnwidth}{l|X|L{2cm}}
\toprule
\textbf{Task}&\textbf{Template}&\textbf{Label Space}\\
\midrule
\multirow[t]{4}{*}{\textbf{SST2}}&Review: contains no wit , only labored gags&negative, positive\\
&Sentiment: negative&\\
&Review: the film is powerful , accessible and funny .&\\
&Sentiment:&\\
\midrule
\multirow[t]{4}{*}{\textbf{SUBJ}}&Input: the script isn't very good ; not even someone as gifted as hoffman ( the actor ) can make it work .&\multirow[t]{4}*{subjective, objective}\\
&Type: subjective&\\
&Input: he must do this in secret so that the parents and school personnel know nothing of his plan .&\\
&Type:&\\
\midrule
\multirow[t]{4}{*}{\textbf{MPQA}}&Review: would not find it at all strange&negative, positive\\
&Sentiment: negative&\\
&Review: as small ( yet acceptable ) as possible&\\
&Sentiment:&\\
\midrule
\multirow[t]{4}{*}{\textbf{AGNews}}&Input: Carlyle Looks Toward Commercial Aerospace (Reuters). "Reuters - Private investment firm Carlyle Group, which has a reputation for making well-timed and occasionally controversial plays in the defense industry, has quietly placed its bets on another part of the market.&world, sports, business, technology\\
&Type: technology&\\
&Input: Superstar Kewell remains centre of attention. Socceroo forward Harry Kewell loosens up by tossing around a ball at Bondi beach yesterday. Photo: Craig Golding. There were half a dozen Socceroos standing on a raised platform in Sydney \#39;s&\\
&Type:&\\
\midrule
\multirow[t]{6}{*}{\textbf{CB}}&Premise: It was a complex language. Not written down but handed down. One might say it was peeled down.&False, True, Neither\\
&Hypothesis: the language was peeled down&\\
&Prediction: False&\\
&Premise: A: so I don't know if I wasn't drug tested based on that or because the man who hired me didn't request the drug test, because I know that my company does drug testing on occasion. B: Right. Well, for instance, does the company you worked for before have the right or do they have the ability to say, hey, we've already drug tested her and she came up negative. A: Well, no, I don't think they can force another company to not drug test me just by saying that I didn't, I mean,&\\
&Hypothesis: they can force another company to not drug test her&\\
&Prediction:&\\
\midrule
\multirow[t]{4}{*}{\textbf{CR}}&Review: it 's not as stylized as a sony or samsung .&negative, positive\\
&Sentiment: negative&\\
&Review: i went out and got the canon today .&\\
&Sentiment:&\\
\midrule
\multirow[t]{4}{*}{\textbf{DBPedia}}&Input: Geoffrey D. Falksen (born July 31 1982) is an American steampunk writer.&company, school,\\
&Type: artist&artist, athlete,\\
&Input: Monster Night is a 2006 film directed by Leslie Allen and Lorenzo Doumani.&politics,\\
&Type:&transportation, building, nature, village, animal, plant, album, film, book\\
\midrule
\multirow[t]{4}{*}{\textbf{MR}}&Review: "you might say tykwer has done all that heaven allows , if you wanted to make as anti-kieslowski a pun as possible . suffice to say its total promise is left slightly unfulfilled ."&negative, positive\\
&Sentiment: negative&\\
&Review: an alternately raucous and sappy ethnic sitcom . . . you'd be wise to send your regrets .&\\
&Sentiment:&\\
\midrule
\multirow[t]{6}{*}{\textbf{RTE}}&Premise: A man is due in court later charged with the murder 26 years ago of a teenager whose case was the first to be featured on BBC One's Crimewatch. Colette Aram, 16, was walking to her boyfriend's house in Keyworth, Nottinghamshire, on 30 October 1983 when she disappeared. Her body was later found in a field close to her home. Paul Stewart Hutchinson, 50, has been charged with murder and is due before Nottingham magistrates later."&false, true\\
&Hypothesis: Paul Stewart Hutchinson is accused of having stabbed a girl.&\\
&Prediction: false&\\
&Premise: For women earning 22,000 a year, the total pay accumulated after six months maternity leave would be just 5,300 in the UK and 5,850 in Ireland. Entitlements in Germany would also be relatively low, at 5,900, along with those in France, Spain and the Netherlands, all at 6,750. At the other end of the scale, pay received after six months leave in Italy would be 9,150 while in Denmark and Norway it would be as much as 11,000.&\\
&Hypothesis: Maternity leave varies in Europe.&\\
&Prediction:&\\
\midrule
\multirow[t]{4}{*}{\textbf{TREC}}&Question: How did serfdom develop in and then leave Russia ?&description, entity, \\
&Type: description&expression, human, \\
&Question: What is Shakespeare 's nickname ?&location, number\\
&Type:&\\
\bottomrule
\end{tabularx}
\caption{Templates for ICL. These are minimum cases with only one demonstration example for illustration.}
\label{table:icltemplate}
\end{table}

\end{document}